\RequirePackage[svgnames,table]{xcolor}
\documentclass[11pt,letterpaper]{yalearxiv}

\usepackage[numbers,sort&compress]{natbib}
\usepackage{array}
\usepackage{multirow}
\usepackage{fontawesome5}
\usepackage{algorithm}
\usepackage[noend]{algpseudocode}
\algrenewcommand\algorithmicrequire{\textbf{Input:}}
\algrenewcommand\algorithmicensure{\textbf{Output:}}
\usepackage[capitalize,noabbrev]{cleveref}
\crefname{algorithm}{Algorithm}{Algorithms}
\Crefname{algorithm}{Algorithm}{Algorithms}

\definecolor{oursrow}{HTML}{EAF2FB}
\colorlet{pactrow}{oursrow}
\definecolor{bestred}{RGB}{178,24,43}
\newcolumntype{Y}{>{\centering\arraybackslash}X}
\newcommand{\task}[1]{\textsc{#1}}

\newcommand{\etal}{\textit{et al.}}
\newcommand{\pmstd}[1]{\,{\textcolor{black!55}{\scriptsize$\pm$#1}}}

\newtcolorbox{examplebox}{
  colback=oursrow!45, colframe=YaleLightBlue!50!black,
  boxrule=0.5pt, arc=2.5pt, left=8pt, right=8pt, top=5pt, bottom=5pt,
  before skip=6pt, after skip=6pt
}
\newtcolorbox{lemmabox}{
  colback=blue!3!white, colframe=YaleLightBlue!75!black,
  boxrule=0.55pt, arc=2.5pt, left=8pt, right=8pt, top=5pt, bottom=5pt,
  before skip=6pt, after skip=5pt
}
\newtcolorbox{corbox}{
  colback=orange!4!white, colframe=orange!65!black,
  boxrule=0.55pt, arc=2.5pt, left=8pt, right=8pt, top=5pt, bottom=5pt,
  before skip=6pt, after skip=5pt
}
\newtcolorbox{propbox}{
  colback=teal!4!white, colframe=teal!60!black,
  boxrule=0.55pt, arc=2.5pt, left=8pt, right=8pt, top=5pt, bottom=5pt,
  before skip=6pt, after skip=5pt
}

\newtheorem{lemma}{Lemma}
\newtheorem{proposition}{Proposition}
\newtheorem{corollary}{Corollary}


\input{tables/numbers}
\input{tables/numbers_ext}
\input{tables/numbers_r2}
\input{tables/numbers_stats}
\def\ptsameA{38}\def\ptsameB{36}

\newcommand{\Th}[1]{\csname Th@#1\endcsname}
\expandafter\def\csname Th@need\endcsname{0.71}
\expandafter\def\csname Th@plainclose\endcsname{0.42--0.51}
\expandafter\def\csname Th@anchorclose\endcsname{0.67--0.83}
\expandafter\def\csname Th@anchordrift\endcsname{2.2--3.7}
\expandafter\def\csname Th@plainmid\endcsname{$-$0.05}
\expandafter\def\csname Th@plainfar\endcsname{0.68}
\expandafter\def\csname Th@afterstallanchor\endcsname{1.3}
\expandafter\def\csname Th@afterstallplain\endcsname{+0.02}
\expandafter\def\csname Th@stallsplain\endcsname{673}
\expandafter\def\csname Th@stallsanchor\endcsname{199}
\expandafter\def\csname Th@recplain\endcsname{56}
\expandafter\def\csname Th@recanchor\endcsname{92}
\expandafter\def\csname Th@hfirstplain\endcsname{34}
\expandafter\def\csname Th@hlastplain\endcsname{28}
\expandafter\def\csname Th@hlastanchor\endcsname{9}
\expandafter\def\csname Th@pairgraph\endcsname{0.68}
\expandafter\def\csname Th@pairlatent\endcsname{0.54}
\expandafter\def\csname Th@pairdifflo\endcsname{0.13}
\expandafter\def\csname Th@pairdiffhi\endcsname{0.15}
\expandafter\def\csname Th@goals\endcsname{289}
\expandafter\def\csname Th@nearlat\endcsname{0.80}
\expandafter\def\csname Th@neargraph\endcsname{0.78}
\expandafter\def\csname Th@midlat\endcsname{0.50}
\expandafter\def\csname Th@midgraph\endcsname{0.55}
\expandafter\def\csname Th@farlat\endcsname{0.53}
\expandafter\def\csname Th@fargraph\endcsname{0.53}
\expandafter\def\csname Th@sepgraph\endcsname{0.84}
\expandafter\def\csname Th@seplatent\endcsname{0.64}
\expandafter\def\csname Th@r2block\endcsname{0.98}
\expandafter\def\csname Th@r2angle\endcsname{0.92}
\expandafter\def\csname Th@nnblock\endcsname{1.2}
\expandafter\def\csname Th@randblock\endcsname{120}
\expandafter\def\csname Th@rectemporal\endcsname{45}
\expandafter\def\csname Th@recmotion\endcsname{29}
\expandafter\def\csname Th@pmlinestart\endcsname{4.4}
\expandafter\def\csname Th@pmlineclosest\endcsname{2.1}
\expandafter\def\csname Th@pmlineone\endcsname{21}
\expandafter\def\csname Th@pmlinetwo\endcsname{49}
\expandafter\def\csname Th@pmhopclosest\endcsname{5}
\expandafter\def\csname Th@trlinestart\endcsname{3.9}
\expandafter\def\csname Th@trlineclosest\endcsname{0.7}
\expandafter\def\csname Th@trlineone\endcsname{60}
\expandafter\def\csname Th@trlinetwo\endcsname{78}
\expandafter\def\csname Th@trhopclosest\endcsname{5}
\expandafter\def\csname Th@ptlinestart\endcsname{3.6}
\expandafter\def\csname Th@ptlineclosest\endcsname{1.8}
\expandafter\def\csname Th@ptlineone\endcsname{24}
\expandafter\def\csname Th@ptlinetwo\endcsname{55}
\expandafter\def\csname Th@pthopclosest\endcsname{4}
\expandafter\def\csname Th@endnear\endcsname{23--26}
\expandafter\def\csname Th@endline\endcsname{60--65}
\expandafter\def\csname Th@withintwo\endcsname{8}
\expandafter\def\csname Th@stretchmidlo\endcsname{0.92}
\expandafter\def\csname Th@stretchmidhi\endcsname{1.08}
\expandafter\def\csname Th@stretchfarlo\endcsname{0.97}
\expandafter\def\csname Th@stretchfarhi\endcsname{1.03}
\expandafter\def\csname Th@trplain\endcsname{0.33--0.46}
\expandafter\def\csname Th@tranchor\endcsname{0.74}
\expandafter\def\csname Th@ptplain\endcsname{0.31--0.37}
\expandafter\def\csname Th@ptanchor\endcsname{0.53}

\input{tables/numbers_exp2x}
\providecommand{\Pf}[1]{\csname Pf@#1\endcsname}
\expandafter\def\csname Pf@pred-n\endcsname{20}
\expandafter\def\csname Pf@pred-ngain\endcsname{19}
\expandafter\def\csname Pf@pred-lewm\endcsname{$-$0.90}
\expandafter\def\csname Pf@pred-gain\endcsname{0.82}
\expandafter\def\csname Pf@bey-pusht-both\endcsname{53}
\expandafter\def\csname Pf@bey-pusht-lewm\endcsname{5}
\expandafter\def\csname Pf@bey-pusht-min\endcsname{9}
\expandafter\def\csname Pf@bey-pusht-ours\endcsname{30}
\expandafter\def\csname Pf@bey-pusht-opt\endcsname{53}
\expandafter\def\csname Pf@bey-reacher-lewm\endcsname{60}
\expandafter\def\csname Pf@bey-reacher-min\endcsname{77}
\expandafter\def\csname Pf@bey-reacher-ours\endcsname{68}
\expandafter\def\csname Pf@bey-reacher-opt\endcsname{80}
\expandafter\def\csname Pf@bey-cube-lewm\endcsname{19}
\expandafter\def\csname Pf@bey-cube-min\endcsname{24}
\expandafter\def\csname Pf@bey-cube-ours\endcsname{29}
\expandafter\def\csname Pf@bey-cube-opt\endcsname{31}
\expandafter\def\csname Pf@enc-pmm-lewm-1\endcsname{0.98}
\expandafter\def\csname Pf@enc-pmm-lewm-h\endcsname{9}
\expandafter\def\csname Pf@enc-pmm-pldm-1\endcsname{0.98}
\expandafter\def\csname Pf@enc-pmm-pldm-h\endcsname{8}
\expandafter\def\csname Pf@enc-pmm-dinov2-1\endcsname{0.93}
\expandafter\def\csname Pf@enc-pmm-dinov2-h\endcsname{6}
\expandafter\def\csname Pf@enc-pmm-hlo\endcsname{6}
\expandafter\def\csname Pf@enc-pmm-hhi\endcsname{9}
\expandafter\def\csname Pf@enc-tworoom-lewm-1\endcsname{0.97}
\expandafter\def\csname Pf@enc-tworoom-lewm-h\endcsname{5}
\expandafter\def\csname Pf@enc-tworoom-pldm-1\endcsname{0.95}
\expandafter\def\csname Pf@enc-tworoom-pldm-h\endcsname{6}
\expandafter\def\csname Pf@enc-tworoom-dinov2-1\endcsname{0.93}
\expandafter\def\csname Pf@enc-tworoom-dinov2-h\endcsname{8}
\expandafter\def\csname Pf@enc-tworoom-hlo\endcsname{5}
\expandafter\def\csname Pf@enc-tworoom-hhi\endcsname{8}
\expandafter\def\csname Pf@enc-r1lo\endcsname{0.93}
\expandafter\def\csname Pf@enc-r1hi\endcsname{0.98}
\expandafter\def\csname Pf@dr-pmm-released\endcsname{48}
\expandafter\def\csname Pf@dr-pmm-relmin\endcsname{47}
\expandafter\def\csname Pf@dr-pmm-ours\endcsname{72}
\expandafter\def\csname Pf@dr-tworoom-released\endcsname{33}
\expandafter\def\csname Pf@dr-tworoom-relmin\endcsname{46}
\expandafter\def\csname Pf@dr-tworoom-ours\endcsname{74}
\expandafter\def\csname Pf@dr-pusht-released\endcsname{31}
\expandafter\def\csname Pf@dr-pusht-relmin\endcsname{37}
\expandafter\def\csname Pf@dr-pusht-ours\endcsname{53}
\expandafter\def\csname Pf@pml-sd-400\endcsname{87}
\expandafter\def\csname Pf@pml-sd-x\endcsname{56}
\expandafter\def\csname Pf@dwmga-tworoom-75\endcsname{95}
\expandafter\def\csname Pf@dwmga-pmm-200\endcsname{20}
\expandafter\def\csname Pf@dwmc-pml-full-42\endcsname{95}
\expandafter\def\csname Pf@dwmc-pml-full-1\endcsname{100}
\expandafter\def\csname Pf@dwmc-pml-pix\endcsname{70}
\expandafter\def\csname Pf@dwmc-pml-prop\endcsname{95}
\expandafter\def\csname Pf@cube-lewm-25\endcsname{70}

\input{tables/numbers_contemp}
\providecommand{\Dg}[1]{\csname Dg@#1\endcsname}
\expandafter\def\csname Dg@pmm-r1\endcsname{0.99}
\expandafter\def\csname Dg@pmm-r6\endcsname{0.59}
\expandafter\def\csname Dg@pmm-hops\endcsname{34}
\expandafter\def\csname Dg@pmm-lewm\endcsname{13}
\expandafter\def\csname Dg@pmm-min\endcsname{12}
\expandafter\def\csname Dg@pmm-ours\endcsname{85}
\expandafter\def\csname Dg@pmm-ap\endcsname{80}
\expandafter\def\csname Dg@pmm-ngc\endcsname{100}
\expandafter\def\csname Dg@pmm-metro\endcsname{96}
\expandafter\def\csname Dg@pmm-gain\endcsname{71}
\expandafter\def\csname Dg@tworoom-r1\endcsname{0.99}
\expandafter\def\csname Dg@tworoom-r6\endcsname{0.50}
\expandafter\def\csname Dg@tworoom-hops\endcsname{11}
\expandafter\def\csname Dg@tworoom-lewm\endcsname{31}
\expandafter\def\csname Dg@tworoom-min\endcsname{62}
\expandafter\def\csname Dg@tworoom-ours\endcsname{93}
\expandafter\def\csname Dg@tworoom-ap\endcsname{97}
\expandafter\def\csname Dg@tworoom-ngc\endcsname{99}
\expandafter\def\csname Dg@tworoom-metro\endcsname{99}
\expandafter\def\csname Dg@tworoom-gain\endcsname{62}
\expandafter\def\csname Dg@reacher-r1\endcsname{1.00}
\expandafter\def\csname Dg@reacher-r6\endcsname{0.56}
\expandafter\def\csname Dg@reacher-hops\endcsname{10}
\expandafter\def\csname Dg@reacher-lewm\endcsname{60}
\expandafter\def\csname Dg@reacher-min\endcsname{77}
\expandafter\def\csname Dg@reacher-ours\endcsname{68}
\expandafter\def\csname Dg@reacher-ap\endcsname{34}
\expandafter\def\csname Dg@reacher-ngc\endcsname{75}
\expandafter\def\csname Dg@reacher-metro\endcsname{--}
\expandafter\def\csname Dg@reacher-gain\endcsname{8}
\expandafter\def\csname Dg@pusht-r1\endcsname{0.91}
\expandafter\def\csname Dg@pusht-r6\endcsname{0.74}
\expandafter\def\csname Dg@pusht-hops\endcsname{18}
\expandafter\def\csname Dg@pusht-lewm\endcsname{5}
\expandafter\def\csname Dg@pusht-min\endcsname{9}
\expandafter\def\csname Dg@pusht-ours\endcsname{30}
\expandafter\def\csname Dg@pusht-ap\endcsname{66}
\expandafter\def\csname Dg@pusht-ngc\endcsname{11}
\expandafter\def\csname Dg@pusht-metro\endcsname{33}
\expandafter\def\csname Dg@pusht-gain\endcsname{25}
\expandafter\def\csname Dg@cube-r1\endcsname{0.99}
\expandafter\def\csname Dg@cube-r6\endcsname{0.77}
\expandafter\def\csname Dg@cube-hops\endcsname{13}
\expandafter\def\csname Dg@cube-lewm\endcsname{19}
\expandafter\def\csname Dg@cube-min\endcsname{24}
\expandafter\def\csname Dg@cube-ours\endcsname{29}
\expandafter\def\csname Dg@cube-ap\endcsname{56}
\expandafter\def\csname Dg@cube-ngc\endcsname{58}
\expandafter\def\csname Dg@cube-metro\endcsname{--}
\expandafter\def\csname Dg@cube-gain\endcsname{10}
\expandafter\def\csname Dg@mw-xlo\endcsname{2.6}
\expandafter\def\csname Dg@mw-xhi\endcsname{14}
\expandafter\def\csname Dg@mw-ntask\endcsname{9}
\expandafter\def\csname Dg@mw-demo-solved\endcsname{65}
\expandafter\def\csname Dg@mw-demo-n\endcsname{72}
\expandafter\def\csname Dg@pt-obj-graph\endcsname{0.68}
\expandafter\def\csname Dg@pt-obj-lat\endcsname{0.54}
\expandafter\def\csname Dg@cube-fail\endcsname{40}
\expandafter\def\csname Dg@cube-never\endcsname{15}
\expandafter\def\csname Dg@cube-cm\endcsname{17}
\expandafter\def\csname Dg@ogb-ga\endcsname{39}
\expandafter\def\csname Dg@ogb-random\endcsname{0}
\expandafter\def\csname Dg@ogb-released\endcsname{0}
\expandafter\def\csname Dg@ogb-relmin\endcsname{0}
\expandafter\def\csname Dg@ogb-ga-tasks\endcsname{16/2/94/56/28}
\providecommand{\Fx}[1]{\csname Fx@#1\endcsname}
\expandafter\def\csname Fx@pusht-k6\endcsname{30}
\expandafter\def\csname Fx@pusht-k6-n\endcsname{3}
\expandafter\def\csname Fx@cube-k6\endcsname{26}
\expandafter\def\csname Fx@cube-k6-n\endcsname{3}
\expandafter\def\csname Fx@pusht-k3\endcsname{30}
\expandafter\def\csname Fx@pusht-k3-n\endcsname{3}
\expandafter\def\csname Fx@cube-k3\endcsname{38}
\expandafter\def\csname Fx@cube-k3-n\endcsname{1}
\expandafter\def\csname Fx@pusht-k3sw\endcsname{72}
\expandafter\def\csname Fx@pusht-k3sw-n\endcsname{3}
\expandafter\def\csname Fx@cube-k3sw\endcsname{26}
\expandafter\def\csname Fx@cube-k3sw-n\endcsname{1}
\expandafter\def\csname Fx@pusht-k2sw\endcsname{71}
\expandafter\def\csname Fx@pusht-k2sw-n\endcsname{3}
\expandafter\def\csname Fx@cube-k2sw\endcsname{39}
\expandafter\def\csname Fx@cube-k2sw-n\endcsname{3}
\expandafter\def\csname Fx@pusht-ap\endcsname{65}
\expandafter\def\csname Fx@pusht-ap-n\endcsname{3}
\expandafter\def\csname Fx@cube-ap\endcsname{53}
\expandafter\def\csname Fx@cube-ap-n\endcsname{3}
\expandafter\def\csname Fx@pusht-k2sw-seeds\endcsname{74/68/72}
\expandafter\def\csname Fx@cube-k2sw-seeds\endcsname{46/36/36}
\expandafter\def\csname Fx@pmm-k2sw\endcsname{94}
\expandafter\def\csname Fx@tworoom-k2sw\endcsname{100}
\expandafter\def\csname Fx@pmm-k2sw-n\endcsname{3}
\expandafter\def\csname Fx@tworoom-k2sw-n\endcsname{3}
\expandafter\def\csname Fx@pml-rt-nav-400\endcsname{76}
\expandafter\def\csname Fx@pml-rt-nav-x\endcsname{45}
\expandafter\def\csname Fx@pml-rt-lewm-400\endcsname{4}
\expandafter\def\csname Fx@pmmx-k2sw\endcsname{91}
\expandafter\def\csname Fx@reacher-k2sw\endcsname{54}
\expandafter\def\csname Fx@pmmx-k2sw-n\endcsname{3}
\expandafter\def\csname Fx@reacher-k2sw-n\endcsname{3}
\expandafter\def\csname Fx@pmm-k3\endcsname{92}
\expandafter\def\csname Fx@tworoom-k3\endcsname{100}
\expandafter\def\csname Dg@x23b-pmm-far-max\endcsname{19}
\expandafter\def\csname Dg@x23b-pml-max\endcsname{9}
\expandafter\def\csname Dg@x23b-ttgs-tworoom-75\endcsname{82}
\expandafter\def\csname Dg@x23b-gas-tworoom-75\endcsname{86}
\expandafter\def\csname Dg@pt-route-temporal\endcsname{45}
\expandafter\def\csname Dg@pt-route-blockmove\endcsname{29}
\expandafter\def\csname Dg@pt-route-jump-px\endcsname{22}
\expandafter\def\csname Dg@wg-pmm-lat-near\endcsname{0.97}
\expandafter\def\csname Dg@wg-pmm-lat-far\endcsname{0.17}
\expandafter\def\csname Dg@wg-pmm-graph-near\endcsname{0.75}
\expandafter\def\csname Dg@wg-pmm-graph-far\endcsname{0.69}
\expandafter\def\csname Dg@wg-pmm-nn-far\endcsname{0}
\expandafter\def\csname Dg@wg-tr-lat-near\endcsname{0.28}
\expandafter\def\csname Dg@wg-tr-lat-far\endcsname{0.05}
\expandafter\def\csname Dg@wg-tr-graph-near\endcsname{0.84}
\expandafter\def\csname Dg@wg-tr-graph-far\endcsname{0.50}
\expandafter\def\csname Dg@wg-tr-nn-far\endcsname{0}

\input{tables/numbers_e1}
\input{tables/numbers_e3b}
\input{tables/numbers_e2}
\makeatletter
\expandafter\def\csname Fx@pmm-k2sw-100\endcsname{97}
\expandafter\def\csname Fx@pmm-k2sw-100-seeds\endcsname{98/96/96}
\expandafter\def\csname Fx@pusht-k2sw-50\endcsname{85}
\expandafter\def\csname Fx@pusht-k2sw-50-seeds\endcsname{84/80/90}
\makeatother

\input{tables/numbers_pushtdist}

\title{Long-Horizon Planning with Latent World Models via Graph Anchors}
\keywords{Latent World Models, Long-Horizon Planning, Model-Predictive Control, Experience Graphs, Visual Control}

\author[1]{Syed Nazmus Sakib}
\author[1]{Abdul Monaf Chowdhury}
\author[3]{Peizhen Li}
\author[1]{Nafiul Haque}
\author[1]{Chandak Chakma}
\author[1]{Md Mehedi Hasan}
\author[3,*]{Zehong Cao}
\author[1,2,*]{Shifat E. Arman}

\affil[1]{Department of Robotics and Mechatronics Engineering, University of Dhaka}
\affil[2]{Department of Engineering Science, University of Oxford}
\affil[3]{School of CSIT, Adelaide University}

\hypersetup{
  colorlinks=true,
  linkcolor=YaleLightBlue,
  citecolor=YaleLightBlue,
  urlcolor=YaleLightBlue,
  pdftitle={Long-Horizon Planning with Latent World Models via Graph Anchors},
  pdfauthor={Syed Nazmus Sakib, Abdul Monaf Chowdhury, Peizhen Li, Nafiul Haque, Chandak Chakma, Md Mehedi Hasan, Zehong Cao, Shifat E. Arman}
}

\newcommand{\websiteurl}{https://syed-nazmus-sakib.github.io/graph-anchors/}
\newcommand{\codeurl}{https://github.com/syed-nazmus-sakib/graph-anchors}
\newcommand{\linksrow}{%
  \par\vspace{4pt}{\centering\fontsize{10}{12}\selectfont
    \href{\websiteurl}{\color{YaleLightBlue}\faGlobe\enspace Project Page}\hspace{2.4em}%
    \href{\codeurl}{\color{YaleLightBlue}\faGithub\enspace Code}\par}\vspace{2pt}}
\makeatletter
\let\orig@abscontent\abscontent
\renewcommand{\abscontent}{\linksrow\orig@abscontent}
\makeatother

\newcommand{\venue}{Preprint}
\fancypagestyle{firststyle}{%
  \fancyhf{}%
  \fancyhead[R]{\footerfont\itshape\venue}%
  \fancyfoot[L]{\footerfont\itshape $^{*}$Corresponding authors: \href{mailto:Jimmy.Cao@adelaide.edu.au}{\texttt{Jimmy.Cao@adelaide.edu.au}}, \href{mailto:shifat.arman@eng.ox.ac.uk}{\texttt{shifat.arman@eng.ox.ac.uk}}.}%
  \fancyfoot[R]{\footerfont\thepage}}
\begin{document}

\begin{abstract}
Latent world models plan in a compact latent space by predicting the outcomes of actions. Their planners judge a plan by the latent distance between its predicted end state and the goal image. We show that this distance fails in two distinct ways. First, it loses directional signal a few steps from the goal, causing distant goals to fail. Second, it weighs the robot arm over the manipulated object, causing manipulation to fail at contact. We measure both failures directly on recorded data by asking whether a lower latent distance corresponds to true progress toward the goal. Across five environments, six visual representations (including raw pixels and V-JEPA~2), and nine Meta-World tasks, this ranking test cleanly separates the two failures, and the graph distance to a goal predicts where closed-loop planning breaks down. Both failures share a single remedy: replacing the distant goal with a nearby target that lies on a recorded trajectory. We realize this principle with \emph{Graph Anchors}, a training-free planner for frozen latent world models. It builds a sparse graph once over the model's own training frames and aims each planning step at a node a few moves ahead along the shortest path to the goal, keeping paths on recorded transitions wherever contact matters. On goals up to 400 steps away, Graph Anchors raise LeWM's success from \R{pmm-released-200}\% to \R{pmm-ours-200}\% on PointMaze-Medium, from \R{pml-released-400}\% to \R{pml-ours-400}\% on PointMaze-Large, and from \R{pusht-released-150}\% to \Fx{pusht-k2sw}\% on Push-T, where stronger search toward the goal fails and where prior training-free planners break down under missing step horizons, subsampling shortcuts, or fragmented contact.
\end{abstract}

\maketitle

% ==============================================================================
\section{Introduction}
\label{sec:intro}

\begin{figure}[t]
\centering
\includegraphics[width=0.72\textwidth]{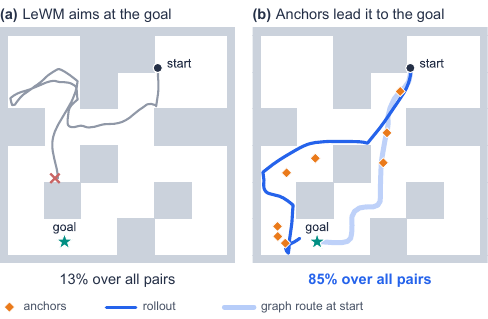}
\caption{\textbf{A distant goal misleads the planner, while near targets on a recorded route guide it.} \textbf{(a)}~LeWM aims every plan straight at a goal 200 steps away and loses its way in the maze. \textbf{(b)}~Graph Anchors aim each plan at a nearby node (orange) along the shortest route through the recorded data (light blue), and the exact same frozen planner reaches the goal.}
\label{fig:teaser}
\end{figure}

World models predict how actions change the world, allowing an agent to evaluate candidate plans in imagination before acting~\cite{ha2018world,hafner2019learning,hafner2019dream,hansen2024td}. Recent visual world models predict in the latent space of an image encoder~\cite{oquab2023dinov2,sobal2026learning,maes2026leworldmodel} and plan by sampling action sequences, rolling out their predicted futures, and keeping the actions whose final latent state lies closest to a goal image~\cite{chua2018deep,rubinstein1999cross}. This approach works reliably for nearby goals. On PointMaze-Medium, LeWorldModel (LeWM)~\cite{maes2026leworldmodel} solves \R{pmm-released-50}\% of goals 50 steps away, yet drops to just \R{pmm-released-200}\% at 200 steps, as illustrated in \cref{fig:teaser}. On Push-T it solves only \R{pusht-released-150}\% at 150 steps, often parking the pusher while leaving the T-block behind.

Why does a planner that easily solves short-range goals break down here? Rather than guessing, we test the latent goal cost directly. On recorded data, before running any planner, we measure how often a lower latent distance corresponds to a move that genuinely progresses toward the goal (\cref{sec:problem}). This simple test isolates two root failures. The first is \emph{range}: latent distance ranks local moves accurately right next to the goal, but loses directional signal within a few graph hops across every navigation task and across six visual encoders, including raw pixels, DINOv2, CLIP, SigLIP, V-JEPA~2, and LeWM (\cref{app:range}). The second is \emph{focus}: whole-image distance weighs the robot arm far more heavily than the manipulated object, preferring an accurate arm pose over completing the task across all nine Meta-World~\cite{yu2020meta} object tasks we measure (\cref{app:focus}).

Importantly, both failures have the exact same remedy. Whole-image distance works reliably whenever its target is nearby and lies along a route recorded in the data. On Meta-World, keeping the released cost untouched and simply aiming at a demonstration frame 20 steps ahead solves \Dg{mw-demo-solved} of \Dg{mw-demo-n} object episodes that the final goal image never solves. We introduce \emph{Graph Anchors} to provide such targets automatically without demonstrations or retraining. We construct a sparse graph once over the world model's own training frames, compute a shortest path to the goal, and aim each planning step at an anchor node a few moves ahead on that path. Where object contact matters, penalizing cross-episode jumps keeps the route on continuous recorded transitions. With the released search and world model untouched, replacing only the distant goal with our nearby anchor lifts PointMaze-Medium at 200 steps from \Q{fa-pmm-released}\% to \Q{fa-pmm-gah5}\%, whereas stronger search aimed at the distant goal stalls at \Q{fa-pmm-relmin}\%. On Push-T, preserving recorded transitions raises $d{=}150$ success from \Fx{pusht-k6}\% to \Fx{pusht-k2sw}\%.

Our main contributions are threefold:
\begin{itemize}[leftmargin=1.4em,itemsep=3pt,topsep=3pt]
\item \textbf{A direct diagnosis of the latent goal cost.} A local ranking test on recorded data separates two distinct failures, \emph{range} and \emph{focus}, across five environments, six image representations, and nine Meta-World tasks. Offline graph distance predicts where closed-loop planning fails, and a formal drift analysis (\cref{app:theory}) connects local ranking to linear versus quadratic reaching times.
\item \textbf{Graph Anchors.} A simple, training-free planner for frozen latent world models that cures both failures using nearby anchors on recorded routes, changing only the target and scoring rule of the planner while leaving the encoder, predictor, and CEM optimizer untouched.
\item \textbf{Long-horizon evaluation and failure analysis of alternative planners.} Across goals up to 400 steps away on navigation and manipulation, we benchmark against world-model planners, offline goal-conditioned policies, graph-guided policies (GAS~\cite{baek2025graph}, TTGS~\cite{opryshko2025test}), and training-free world-model planners (AP~\cite{liu2026aim}, NGC~\cite{you2026latent}, Metro-WM~\cite{lee2026metro}), showing how each missing diagnostic axis causes specific structural failures (\cref{sec:frozenplanners}, \cref{app:planners}).
\end{itemize}

% ==============================================================================
\section{Related Work}
\label{sec:related}

\paragraph{Latent world models and their goal costs.}
World models predict the consequences of actions in a learned latent space~\cite{ha2018world,hafner2019learning,hafner2019dream} and support planning via model-predictive control~\cite{chua2018deep,hansen2024td}. For visual goal reaching, DINO-WM~\cite{zhou2024dino} plans with CEM over predicted DINOv2~\cite{oquab2023dinov2} patch features, PLDM~\cite{sobal2026learning} learns latent dynamics from reward-free offline trajectories, and LeWM~\cite{maes2026leworldmodel} trains encoder and predictor jointly from pixels with a joint-embedding objective~\cite{assran2023self}. JEPA-WM~\cite{terver2025drives} studies the design choices behind such planners, and V-JEPA~2~\cite{assran2025v} extends them to robot actions. These planners minimize latent distance to a goal image and are typically evaluated around 25 steps ahead. Recent work improves the predictor~\cite{du2026beyond,gao2026fast} or questions the goal cost directly: Temporal-Distance JEPA~\cite{bai2026temporal} and TEMPO~\cite{moukheiber2026long} learn temporal distances, Li \etal~\cite{li2026beyond} learn reachability metrics, Singh~\cite{singh2026objective} shows that LeWM's latent distance misrepresents progress on Two-Room, and Alrasheed \etal~\cite{alrasheed2026planning} find that world models rank actions reliably only near their imagined horizon. We measure directional ranking directly against graph distance and against scene elements, and keep the world model and cost completely unchanged.

\paragraph{Subgoals for latent planners.}
Earlier visual planners reached goal images with pixel-space models~\cite{finn2017deep} or planned over intermediate subgoals proposed by learned generative models~\cite{nasiriany2019planning,pertsch2020long}. HWM~\cite{zhang2026hierarchical} trains world models at multiple temporal scales, FF-JEPA~\cite{masip2026ff} predicts intermediate subgoals, SAGE~\cite{cheng2026sage} trains a subgoal generator and action proposal model for LeWM, and PILOT~\cite{liu2026pilot} generates visual subgoals with a video model. All of these require training new components. Graph Anchors instead retrieve subgoals directly from the world model's existing training data, requiring no additional training whenever the data covers the route.

\paragraph{Graphs of experience.}
SPTM~\cite{savinov2018semi} links observations by temporal adjacency and learned reachability, SoRB~\cite{eysenbach2019search} plans over stored states with a goal-conditioned value function, HTM~\cite{liu2020hallucinative} and L3P~\cite{zhang2021world} search graphs of generated images and learned landmarks, and sparse graphical memory~\cite{emmons2020sparse} and landmark graphs~\cite{huang2019mapping} maintain compact state graphs for goal reaching. More recently, GAS~\cite{baek2025graph} builds a graph in a learned temporal-distance space for a low-level policy, and TTGS~\cite{opryshko2025test} searches an offline dataset over learned value distances to guide a frozen policy. Three other training-free planners retarget a frozen LeWM directly: Anchored Planning (AP)~\cite{liu2026aim} retrieves the single-episode segment best matching the current frame and goal given the step count $d$, NGC~\cite{you2026latent} replaces latent distance with shortest-path length over a subsampled undirected graph, and Metro-WM~\cite{lee2026metro} retrieves subgoals from a sparse graph of recorded frames. We compare all six paradigms directly under our diagnostic framework in \cref{sec:frozenplanners} and \cref{app:planners}.

% ==============================================================================
\section{Preliminaries}
\label{sec:prelim}

\paragraph{Problem setting.}
We study visual goal-conditioned planning from offline, reward-free trajectories~\cite{maes2026leworldmodel,sobal2026learning,park2025ogbench}. We are given a fixed dataset of image observations and actions, with no rewards and no extra environment interaction for training. Following LeWM~\cite{maes2026leworldmodel}, each action is a block of five consecutive environment steps. At test time, the agent observes $o_1$ and receives a goal image $o_g$, succeeding if it reaches the goal within its step budget.

\paragraph{Latent world model.}
A latent world model~\cite{maes2026leworldmodel} predicts future states inside a compact embedding space using an encoder $f$ and a dynamics predictor $p$:
\begin{equation}
z_t = f(o_t), \qquad \hat z_{t+1} = p(\hat z_t, a_t),
\label{eq:wm}
\end{equation}
where $z_t$ embeds observation $o_t$ and $\hat z_{t+1}$ predicts the embedding after executing action block $a_t$. Throughout our experiments we keep the released LeWM encoder and predictor completely frozen.

\paragraph{Latent planning.}
LeWM~\cite{maes2026leworldmodel} rolls out candidate sequences of $H$ action blocks from $\hat z_1 = z_1$ and scores each plan by the squared latent distance between its final predicted state and the goal embedding $z_g = f(o_g)$:
\begin{equation}
a^{*}_{1:H} = \arg\min_{a_{1:H}}\ \Vert\hat z_{H+1} - z_g\Vert_2^2, \qquad H=5.
\label{eq:plan}
\end{equation}
The planner minimizes \cref{eq:plan} using the cross-entropy method (CEM)~\cite{rubinstein1999cross}, iteratively sampling action sequences from a Gaussian, scoring them, and refitting the Gaussian to the lowest-cost elites. The agent executes the chosen plan and replans from its new observation. Consequently, closed-loop planning moves toward the goal only if lower latent distance actually reflects true progress.

% ==============================================================================
\section{Graph Anchors}
\label{sec:method}

We first diagnose why latent planning breaks down at long range and during object contact (\cref{sec:problem}). We then present three lightweight components that address both failures without modifying or retraining the world model (\cref{fig:method}): a sparse graph of recorded frames supplying nearby anchors on a shortest path (\cref{sec:graph,sec:anchor}), an episode-switch cost that keeps routes on continuous recorded transitions (\cref{sec:switch}), and closest-approach scoring that guides rollouts smoothly through each anchor (\cref{sec:search}).

\subsection{Where Latent Planning Fails}
\label{sec:problem}\label{sec:diagnosis}

To check whether lower latent distance truly indicates progress, we test the goal cost directly on recorded data using the graph of \cref{sec:graph}. For a goal node $g$, a state $v$ located $h$ hops from $g$, and two neighbours $u, u'$ of $v$ where $u$ is genuinely closer to $g$ than $u'$, we define the directional ranking accuracy:
\begin{equation}
\rho(h) = \Pr\big[\Vert z_u - z_g\Vert_2^2 < \Vert z_{u'} - z_g\Vert_2^2\big].
\label{eq:rho}
\end{equation}
A score of $\rho{=}1$ means the cost always points downhill toward the goal, whereas $\rho{=}\tfrac12$ carries no directional information and behaves like a coin flip. Two concrete cases illustrate what happens in closed loop:

\begin{examplebox}
\textbf{Example 1 (A far goal).} On PointMaze-Medium, a goal 200 steps ahead lies \Th{hfirstplain} graph hops away. Six hops from a goal, squared latent distance orders local moves correctly only \Dg{pmm-r6} of the time. In closed loop, just \Pf{dr-pmm-released}\% of LeWM's decisions end closer to the goal. After 15 decisions the agent is still \Th{hlastplain} hops away, and only \R{pmm-released-200}\% of episodes succeed.
\end{examplebox}

\begin{examplebox}
\textbf{Example 2 (Pushing a block).} On Push-T, the T-block must reach a goal pose 150 steps ahead. Overall local ranking still looks healthy at \Dg{pusht-r1} one hop away and \Dg{pusht-r6} six hops away. Yet latent distance orders the block's true object progress correctly only \Dg{pt-obj-lat} of the time, compared to \Dg{pt-obj-graph} for graph distance along recorded routes. LeWM solves just \R{pusht-released-150}\% of episodes.
\end{examplebox}

\begin{figure}[t]
\centering
\includegraphics[width=0.76\textwidth]{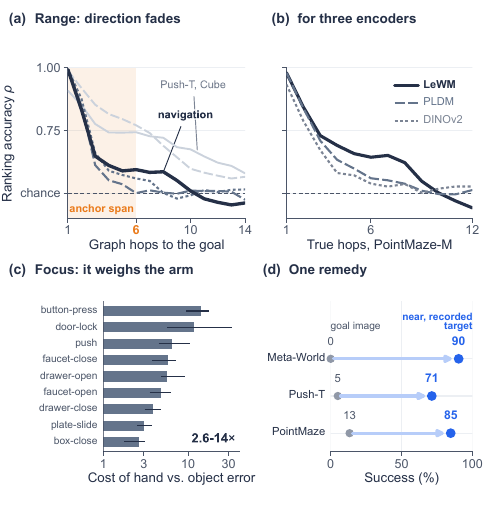}
\caption{\textbf{The diagnosis in four steps.} \textbf{(a)}~Directional ranking accuracy $\rho(h)$ of the LeWM goal cost against graph distance to the goal across 300 goals per environment, with navigation tasks in bold/dashed/dotted and manipulation tasks faded; shaded band marks our anchor span. \textbf{(b)}~Ranking accuracy on PointMaze-Medium for three encoders measured on a true-position reference graph over 200 goals (extended to six encoders including raw pixels and V-JEPA~2 in \cref{fig:e1range}). \textbf{(c)}~JEPA-WM on Meta-World: cost of moving only the hand by 5\,cm divided by leaving only the object off target by its success threshold (median and IQR across 30 placements per task; extended to five encoders in \cref{fig:exchange}). \textbf{(d)}~Closed-loop success aiming at the final goal versus aiming at a nearby target on a recorded route with the exact same cost.}
\label{fig:diagnosis}
\end{figure}

The planner of \cref{sec:prelim} fails in these settings for three coupled reasons.

First, whole-image distance has a short \emph{range}. Across every navigation benchmark, $\rho(h)$ starts near 1 right next to the goal and drops to \Dg{tworoom-r6}--\Dg{pmm-r6} within six hops (\cref{fig:diagnosis}a). \Cref{app:checks} verifies the exact same collapse directly along recorded time steps without building any graph. Measuring hops on a true-position reference graph, LeWM, PLDM, and frozen DINOv2 all start at \Pf{enc-r1lo}--\Pf{enc-r1hi} and fall below 0.55 within \Pf{enc-pmm-hlo}--\Pf{enc-pmm-hhi} hops on PointMaze-Medium (\cref{fig:diagnosis}b). In \cref{app:range} we further test raw pixels, CLIP, SigLIP, and video-trained V-JEPA~2 alongside 10-epoch training checkpoints (\cref{fig:e1range,fig:e2epochs}). Because even raw pixels decay at the exact same rate as V-JEPA~2, short range is an inherent property of comparing two whole images across corridors and walls rather than an undertrained encoder. Far from the goal, heavier search cannot rescue the planner because candidate rollouts are ranked by a directionless score: doubling the horizon leaves Example~1 at \R{pmm-relmin-200}\%. \Cref{sec:anchor} cures this by replacing the distant goal with a nearby target.

Second, whole-image distance has the wrong \emph{focus}. In Example~2, overall $\rho$ decays slowly, so range alone does not explain why Push-T fails. Because whole-image distance averages over all pixels, a large robot arm or agent dominates a small manipulated object. Testing the released JEPA-WM planner~\cite{terver2025drives} on Meta-World~\cite{yu2020meta} by placing arm and object independently with inverse kinematics shows that shifting only the hand by 5\,cm costs \Dg{mw-xlo} to \Dg{mw-xhi}$\times$ more than leaving only the object off target across all nine object tasks (\cref{fig:diagnosis}c), a bias that persists across raw pixels, DINOv2, CLIP, SigLIP, and V-JEPA~2 (\cref{fig:exchange}), even though a linear probe on the exact same frozen features locates the object cleanly (\cref{app:focus}). Under exact simulator dynamics, the released cost solves 0 of 72 object episodes. Merely picking a closer target is not enough either: on Push-T, an anchor three hops ahead without preserving recorded transitions stays at \Fx{pusht-k3}\%. However, aiming the unchanged released cost at a nearby frame \emph{on a recorded route}, such as the frame 20 steps ahead on the episode's own demonstration, immediately solves \Dg{mw-demo-solved} of \Dg{mw-demo-n} Meta-World episodes (\cref{fig:diagnosis}d). Matching a nearby recorded arm pose naturally moves the object as recorded. \Cref{sec:switch} preserves recorded transitions automatically inside our graph.

Third, scoring only the final state of a plan rewards stopping at the target rather than moving past it. A nearby anchor is a stepping stone to pass through, not a resting point. \Cref{sec:search} replaces terminal-step scoring with closest-approach scoring.

\subsection{A Graph of Recorded Experience}
\label{sec:graph}

\begin{figure}[t]
\centering
\includegraphics[width=\textwidth]{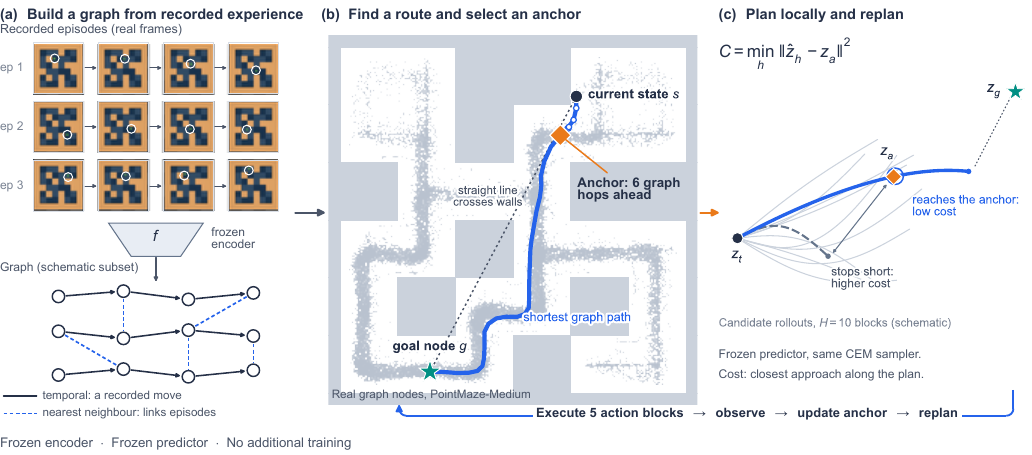}
\caption{\textbf{Graph Anchors turn one distant target into a sequence of nearby ones.} \textbf{(a)}~Encoded training frames are linked by directed temporal edges and local nearest-neighbour edges. \textbf{(b)}~At each replanning step, the active anchor lies a few moves ahead along the shortest graph path to the goal. \textbf{(c)}~CEM scores rollouts by their closest approach to the anchor, executes, and re-anchors from the new observation.}
\label{fig:method}
\end{figure}

We construct a sparse graph $G=(V,E)$ once from the world model's own training trajectories. Each node $i$ represents a recorded frame sampled every five steps (one action block), storing its episode index $e_i$, timestamp $t_i$, and frozen latent embedding $z_i = f(o_i)$. Directed \emph{temporal edges} connect consecutive frames within an episode, while bidirectional \emph{neighbour edges} connect close embeddings across episodes:
\begin{equation}
\begin{aligned}
E_T &= \{(i,j): e_i=e_j,\ t_j=t_i+5\},\\
E_N &= \{(i,j),(j,i): j\in\mathrm{NN}_{10}(i),\ \Vert z_i-z_j\Vert_2^2<\tau\}.
\end{aligned}
\label{eq:edges}
\end{equation}
Temporal edges record real physical transitions that occurred in the dataset. Neighbour edges rely on Euclidean latent distance strictly at one-hop scale, where $\rho$ is near 1, allowing routes to stitch across episodes without tuning any environment-specific distance threshold: $\tau$ is set automatically to the empirical median squared distance of a single recorded action block. Test episodes are always removed before evaluation so no anchor can replay a test trajectory. Building the full graph takes \Q{time-build-s}\,s once on PointMaze-Medium (\cref{app:walkthrough}).

\begin{figure}[t]
\centering
\includegraphics[width=0.74\textwidth]{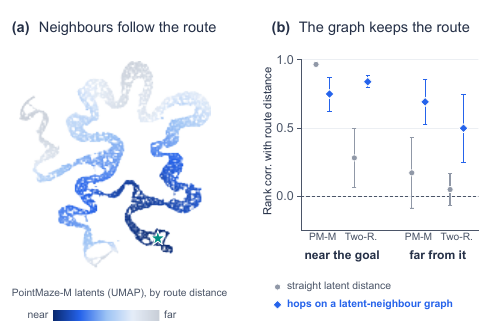}
\caption{\textbf{Why a graph over latents works.} \textbf{(a)}~UMAP~\cite{mcinnes2018umap} visualization of 30k PointMaze-Medium latents coloured by true route distance to one goal (star), showing smooth local corridor structure. \textbf{(b)}~Rank correlation with true walking distance for direct latent distance versus shortest-path graph hops, near the goal (under 15\% of maximum route distance) and far from it (over 50\%), averaged over 20 random goals.}
\label{fig:whygraph}
\end{figure}

Why does chaining local latent neighbours cure long-range failure? Frames that are close in latent space are physically close in the environment: on PointMaze-Medium and Two-Room, virtually none (\Dg{wg-pmm-nn-far}\%) of a frame's ten nearest latent neighbours sits more than one maze cell away by walking distance (\cref{fig:whygraph}a), and checking all 994{,}834 unique neighbour edges on PointMaze-Medium yields zero wall crossings (\cref{app:impl}). Over long distances, direct latent distance breaks down: rank correlation with true walking distance on PointMaze-Medium drops from \Dg{wg-pmm-lat-near} near the goal to \Dg{wg-pmm-lat-far} far from it (\cref{fig:whygraph}b). Shortest-path hops over the local neighbour graph hold correlation at \Dg{wg-pmm-graph-far} far from the goal, and raise Two-Room from \Dg{wg-tr-lat-far} to \Dg{wg-tr-graph-far}. The graph therefore handles global routing while the visual cost only needs to rank the last few hops to a nearby anchor.

\subsection{Choosing the Anchor}
\label{sec:anchor}

At test time, we map the current embedding $z_t$ and goal embedding $z_g$ to their nearest graph nodes $s$ and $g$, find a shortest path $\pi(s,g)$, and select the node $K$ moves ahead as the active target:
\begin{equation}
z_a = z_{\pi_K(s,g)} \quad \text{if } K < d_G(s,g) < \infty, \qquad \text{else } z_g,
\label{eq:anchor}
\end{equation}
where $d_G$ denotes shortest-path length. When the goal is already within $K$ moves or unreachable in the graph, the planner aims directly at $z_g$. Running one reversed Dijkstra search from $g$ caches the shortest path from every node in the graph, so each replanning step reduces to a single fast lookup (\cref{alg:sga} in \cref{app:walkthrough}).

Moving closer to this anchor transfers progress directly to the distant goal. Because $a$ lies on a shortest path from $x$ to $g$, we have $d_G(x,g) = d_G(x,a) + d_G(a,g)$. For any reached state $y$ closer to $a$ than $x$, the directed triangle inequality immediately gives
\begin{equation}
d_G(y,g) \le d_G(y,a) + d_G(a,g) < d_G(x,g).
\label{eq:transfer}
\end{equation}
\Cref{app:theory} formalizes this transfer under metric distortion (\cref{lem:transfer,cor:true}) and proves that sustained local drift converts quadratic random-walk hitting times into linear $O(D)$ progress (\cref{prop:drift}). The span $K$ balances directional reliability against step progress: six hops from a goal, $\rho$ is \Dg{pmm-r6} on PointMaze-Medium versus \diagA{pmm}{10} at ten hops. We use $K{=}6$ as our default navigation span (\cref{sec:design}).

\subsection{Keeping the Route on Recorded Transitions}
\label{sec:switch}

In contact manipulation (Example~2), a nearby anchor alone is insufficient. On Push-T, unweighted shortest paths follow a recorded temporal transition on only \Dg{pt-route-temporal}\% of hops, and only \Dg{pt-route-blockmove}\% carry recorded block motion; typical cross-episode jumps shift the T-block by \Dg{pt-route-jump-px}\,px with no recorded push connecting them. To keep paths on continuous recorded pushes, we charge a higher edge weight $\lambda \ge 1$ on cross-episode neighbour edges:
\begin{equation}
w(i,j) = \begin{cases} 1, & (i,j)\in E_T,\\ \lambda, & (i,j)\in E_N\setminus E_T, \end{cases}
\label{eq:switch}
\end{equation}
and select the anchor after $K$ \emph{temporal} moves along this weighted shortest path so cross-episode jumps do not count as progress. Along temporal edges, arm and object move together as recorded, reproducing the Meta-World demonstration effect (\cref{fig:diagnosis}d) purely from offline training data. Because \cref{eq:transfer} holds for any directed weighted graph with $w \ge 1$, progress transfer remains intact while cross-episode stitching still occurs whenever a jump saves more than $\lambda{-}1$ moves. We use two task-level settings: \emph{navigation} ($K{=}6, \lambda{=}1$) and \emph{contact} ($K{=}2, \lambda{=}3$).

\subsection{Planning Through the Anchor}
\label{sec:search}

When an anchor lies six hops ahead on the way to a much farther goal, the planner should move smoothly \emph{through} the anchor rather than braking at it. CEM therefore scores each rolled-out trajectory $\hat z_{2:H+1}$ by its closest predicted approach to $z_a$:
\begin{equation}
C(\hat z_{2:H+1}, z_a) = \min_{2 \le h \le H+1} \Vert\hat z_h - z_a\Vert_2^2.
\label{eq:cost}
\end{equation}
After executing, the agent observes the new frame, updates its anchor, and replans. Terminal-step scoring wastes action blocks after passing a waypoint and penalizes rollouts that pass cleanly through it; taking the minimum over the plan removes both effects. In navigation we use $H{=}10$ blocks and execute 5; in the contact setting we use $H{=}2$ blocks and execute up to the block of closest approach. CEM preserves LeWM's exact hyperparameters (300 samples, 30 elites, 30 iterations).

\subsection{Overall Objective}
\label{sec:objective}

Combining \cref{eq:anchor,eq:switch,eq:cost}, Graph Anchors replace \cref{eq:plan} with
\begin{equation}
a^{*}_{1:H} = \arg\min_{a_{1:H}}\ \min_{2 \le h \le H+1} \Vert\hat z_h - z_a\Vert_2^2.
\label{eq:overall}
\end{equation}
Only two elements change relative to LeWM: the target shifts from the distant goal to a nearby anchor on a recorded route, and rollout cost shifts from terminal distance to closest approach. The encoder, predictor, and CEM optimizer remain frozen. \Cref{alg:ga} summarizes the loop.

\begin{algorithm}[t]
\caption{Graph Anchors}
\label{alg:ga}
\small
\begin{algorithmic}[1]
\Require Frozen world model ($f$, $p$), training frames, goal image $o_g$, span $K$, switch cost $\lambda$, horizon $H$
\State Build sparse graph $G$ from training frames (\cref{eq:edges}) with edge weights $w$ (\cref{eq:switch})
\State $z_g \gets f(o_g)$; $g \gets$ nearest node; compute reversed Dijkstra tree from $g$ on $G$
\For{each replanning step}
  \State $z_t \gets f(o_t)$; $s \gets$ nearest node
  \State $z_a \gets$ node $K$ temporal moves along $\pi(s,g)$, or $z_g$ (\cref{eq:anchor})
  \State Run unchanged CEM over $a_{1:H}$ with closest-approach cost $\min_h \Vert\hat z_h - z_a\Vert_2^2$ (\cref{eq:overall})
  \State Execute plan (navigation: 5 blocks; contact: up to closest approach)
\EndFor
\end{algorithmic}
\end{algorithm}

% ==============================================================================
\section{Experiments}
\label{sec:experiments}

We evaluate across six visual control benchmarks: Push-T, Two-Room, OGBench-Cube~\cite{park2025ogbench}, and Reacher~\cite{tassa2018deepmind} using the released LeWM checkpoints and datasets~\cite{maes2026leworldmodel}, PointMaze-Medium using D4RL~\cite{fu2020d4rl} expert data and the public LeWM checkpoint of~\cite{jang2026latent}, and PointMaze-Large using OGBench navigation trajectories with LeWM trained via its default script (\cref{app:impl}). Start and goal frames come from the same held-out episode $d$ steps apart with step budget $2d$; \emph{stitched} goals draw start and goal from different held-out episodes (budget 400 on mazes, 200 on Two-Room; both episodes removed from the graph). Each seed fixes 50 identical start and goal pairs across all methods, evaluated over three seeds unless noted, with paired McNemar significance tests reported in \cref{app:stats}. \emph{LeWM} denotes the released planner and \emph{LeWM + min-over-plan} applies our closest-approach search directly to the distant goal, isolating the target change alone. We also compare against PLDM~\cite{sobal2026learning}, DINO-WM~\cite{zhou2024dino}, offline policies GCBC, GCIVL, GCIQL~\cite{park2025ogbench}, and HILP~\cite{park2024foundation} (trained on identical data under a matched short budget detailed in \cref{app:gasttgs}), graph-guided policies GAS~\cite{baek2025graph} and TTGS~\cite{opryshko2025test}, and training-free world-model planners AP~\cite{liu2026aim}, NGC~\cite{you2026latent}, and Metro-WM~\cite{lee2026metro}.

\begin{table}[t]
\centering
\caption{\textbf{Success (\%) on distant goals.} All methods evaluate identical start and goal pairs under budget $2d$ (400 for stitched maze goals). Entries report mean $\pm$ std over 50-pair seeds (usually three seeds), with best in \textcolor{bestred}{\textbf{red}} and second \underline{underlined}; $^\dagger$marks one evaluation. Graph Anchors use the navigation setting ($K{=}6, \lambda{=}1$) on the first row and the contact setting ($K{=}2, \lambda{=}3$) on the second row (PointMaze-Large contact ran on the retrained checkpoint of \cref{tab:retrain}). Full distance sweeps appear in \cref{tab:compare}.}
\label{tab:main}
\small
\setlength{\tabcolsep}{4pt}
\renewcommand{\arraystretch}{1.05}
\begin{tabularx}{\textwidth}{@{}lYYYYYYY@{}}
\toprule
Method & Push-T 150 & Two-Room 75 & PM-M 200 & PM-M stitch. & PM-L 400 & PM-L stitch. & Reacher 150 \\
\midrule
LeWM~\cite{maes2026leworldmodel} & 5\,{\scriptsize$\pm$4} & 31\,{\scriptsize$\pm$3} & 13\,{\scriptsize$\pm$2} & 39\,{\scriptsize$\pm$10} & 8\,{\scriptsize$\pm$0} & 5\,{\scriptsize$\pm$1} & 60\,{\scriptsize$\pm$6} \\
LeWM + min-over-plan & 9\,{\scriptsize$\pm$1} & 62\,{\scriptsize$\pm$10} & 12\,{\scriptsize$\pm$5} & 41\,{\scriptsize$\pm$4} & \underline{26\,{\scriptsize$\pm$6}} & \underline{17\,{\scriptsize$\pm$1}} & \textcolor{bestred}{\textbf{77\,{\scriptsize$\pm$5}}} \\
PLDM~\cite{sobal2026learning} & 0\,{\scriptsize$\pm$0} & 50\,{\scriptsize$\pm$6} & 3\,{\scriptsize$\pm$3} & 28\,{\scriptsize$\pm$4} & 1\,{\scriptsize$\pm$1} & 2\,{\scriptsize$\pm$0} & 15\,{\scriptsize$\pm$1} \\
\midrule
GCIVL~\cite{park2025ogbench} & 0\,{\scriptsize$\pm$0} & 81\,{\scriptsize$\pm$2} & 0$^\dagger$ & 16\,{\scriptsize$\pm$0} & 1\,{\scriptsize$\pm$1} & 3\,{\scriptsize$\pm$1} & 0\,{\scriptsize$\pm$1} \\
GCIQL~\cite{park2025ogbench} & 0\,{\scriptsize$\pm$0} & 37\,{\scriptsize$\pm$4} & 1\,{\scriptsize$\pm$1} & 12\,{\scriptsize$\pm$3} & 3\,{\scriptsize$\pm$1} & 3\,{\scriptsize$\pm$1} & 0\,{\scriptsize$\pm$0} \\
HILP~\cite{park2024foundation} & 0\,{\scriptsize$\pm$0} & 87\,{\scriptsize$\pm$3} & 1\,{\scriptsize$\pm$1} & 18\,{\scriptsize$\pm$0} & 9\,{\scriptsize$\pm$1} & 10\,{\scriptsize$\pm$6} & 0\,{\scriptsize$\pm$0} \\
\midrule
\rowcolor{oursrow}\textbf{Graph Anchors (ours)} & \underline{30\,{\scriptsize$\pm$3}} & \underline{93\,{\scriptsize$\pm$1}} & \underline{85\,{\scriptsize$\pm$6}} & \textcolor{bestred}{\textbf{95\,{\scriptsize$\pm$4}}} & \textcolor{bestred}{\textbf{87\,{\scriptsize$\pm$1}}} & \textcolor{bestred}{\textbf{66\,{\scriptsize$\pm$6}}} & \underline{68\,{\scriptsize$\pm$4}} \\
\rowcolor{oursrow}\quad contact setting ($K{=}2$, $\lambda{=}3$) & \textcolor{bestred}{\textbf{71\,{\scriptsize$\pm$3}}} & \textcolor{bestred}{\textbf{100\,{\scriptsize$\pm$0}}} & \textcolor{bestred}{\textbf{94\,{\scriptsize$\pm$5}}} & \underline{91\,{\scriptsize$\pm$4}} & 13\,{\scriptsize$\pm$6} & 9\,{\scriptsize$\pm$5} & 54\,{\scriptsize$\pm$11} \\
 
\bottomrule
\end{tabularx}
\end{table}

\begin{figure}[t]
\centering
\includegraphics[width=\textwidth]{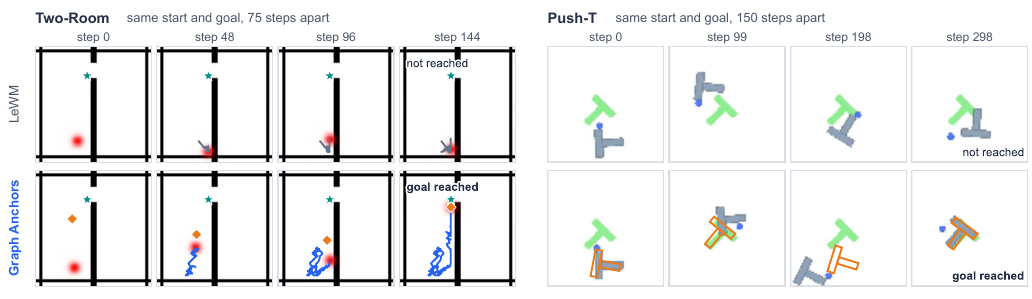}
\caption{\textbf{What the planner sees.} Camera frames on identical start and goal pairs for LeWM (top) and Graph Anchors in the navigation setting (bottom, seed 0), showing the longest solved path missed by LeWM. Two-Room overlays the path so far, goal (star), and active anchor (diamond); Push-T overlays the active anchor block outline and green goal pose. Full videos appear on our project page.}
\label{fig:frames}
\end{figure}

\begin{figure}[t]
\centering
\includegraphics[width=\textwidth]{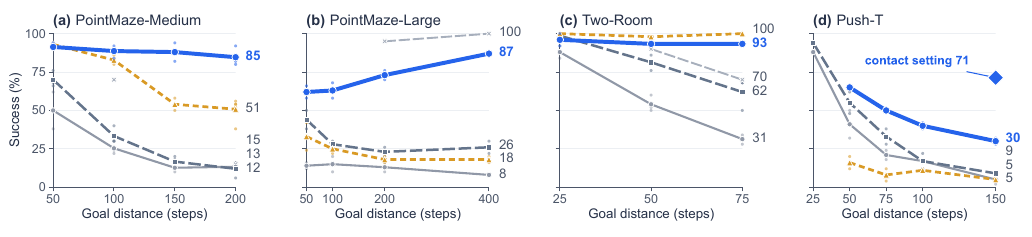}
\caption{\textbf{Near targets eliminate the collapse with goal distance.} Blue: Graph Anchors (navigation setting; $\blacklozenge$ contact setting on Push-T over three seeds). Amber dashed: GCBC, the strongest offline policy. Dark gray dashed: LeWM + min-over-plan. Gray: LeWM. Light gray dashed: DINO-WM (20 pairs, seed 0; its default cost also reads proprioceptive coordinates, ablated in \cref{tab:dwmcost}). Lines show seed means and dots show single 50-pair runs. PointMaze-Large rises with $d$ as the $2d$ budget expands.}
\label{fig:results}
\end{figure}

\subsection{Does Graph Anchors Reach Distant Goals?}
\label{sec:p1}\label{sec:compare}

\Cref{tab:main} and \cref{fig:results} show both the range collapse and its removal. As goals move farther away, LeWM loses \R{dropmin} to \R{dropmax} percentage points across Push-T, Two-Room, and PointMaze-Medium, and LeWM + min-over-plan collapses similarly on both mazes. Graph Anchors hold \R{pmm-ours-200}--\R{pmm-ours-50}\% across $d{=}50$ to $200$ on PointMaze-Medium and \R{tworoom-ours-50}--\R{tworoom-ours-25}\% on Two-Room, while lifting PointMaze-Large at $d{=}400$ from \R{pml-released-400}\% to \R{pml-ours-400}\%. In \cref{fig:frames}, LeWM stalls against the dividing wall on Two-Room and abandons the block on Push-T, whereas anchors guide the agent cleanly through the doorway and push the block toward its target. Across every seed at $d{\ge}50$, Graph Anchors outperform both LeWM baselines ($p<10^{-2}$ on Two-Room and both mazes in \cref{app:stats}) and stitch cross-episode goals cleanly, solving \R{pmm-ours-x}\% of stitched PointMaze-Medium goals versus \R{pmm-released-x}\% for LeWM.

Offline policies excel on compact environments such as Two-Room, where GCBC reaches \R{tworoom-gcbc-75}\% and HILP reaches \R{tworoom-hilp-75}\% at $d{=}75$, yet collapse on larger mazes: GCBC falls to \R{pmm-gcbc-200}\% on PointMaze-Medium $d{=}200$ and \R{pml-gcbc-400}\% on PointMaze-Large $d{=}400$. Simply adding a graph does not rescue them. Even when fed waypoints from our verified latent graph or paired with GAS~\cite{baek2025graph} and TTGS~\cite{opryshko2025test} search, offline policies reach at most \R{x23-pmm-far-max}\% on PointMaze-Medium $d{=}200$ and \R{x23-pml-max}\% on PointMaze-Large $d{=}400$ (\cref{app:gasttgs} diagnoses both value saturation and policy execution failure in detail). DINO-WM's high default PointMaze-Large score (\R{pml-dwm-400}\%) relies on ground-truth agent coordinates inside its goal cost; using images only drops it to \Pf{dwmc-pml-pix}\% versus \R{pml-ours-400}\% for image-only Graph Anchors (\cref{tab:dwmcost}).

Graph Anchors also transfer to other world models whenever their local dynamics are accurate: on Two-Room, adding anchors lifts PLDM from \R{pldm-tworoom-released-75}\% to \R{pldm-tworoom-ours-75}\% and DINO-WM (over frozen DINOv2 latents) from \R{tworoom-dwm-75}\% to \Pf{dwmga-tworoom-75}\%, whereas on PointMaze-Medium PLDM (\R{pldm-pmm-ours-200}\%) and DINO-WM (\Pf{dwmga-pmm-200}\%) cannot reliably reach targets six hops away. Compute overhead is minimal: building the 199k-node PointMaze-Medium graph takes \Q{time-build-s}\,s once, one cached CPU Dijkstra search takes 84--93\,ms, and exact per-decision GPU overhead is just 64.6\,ms (\cref{tab:comp_mech}b).

\subsection{Does It Help at Contact?}
\label{sec:contact}

\begin{figure}[t]
\centering
\includegraphics[width=\textwidth]{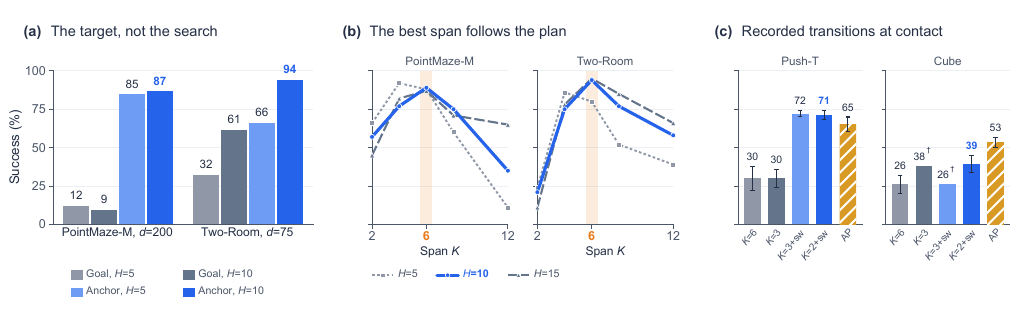}
\caption{\textbf{Why each component is needed.} \textbf{(a)}~2$\times$2 factorial crossing target choice and search rule across three seeds. \textbf{(b)}~Anchor span $K$ crossed with planning horizon $H$ (mean of seeds 0 and 42; shaded: $K{=}6$). \textbf{(c)}~Contact tasks at $d{=}150$ varying anchor span and switch cost ($\lambda{=}3$) against AP (hatched) on identical machine and pairs across three seeds ($^\dagger$seed 0 only).}
\label{fig:ablations}\label{fig:contact}
\end{figure}

Push-T exposes focus failure most sharply, and the contact setting ($K{=}2, \lambda{=}3$) makes a dramatic difference: at $d{=}150$ it reaches \Fx{pusht-k2sw}\%, compared to \Fx{pusht-k6}\% for the navigation setting and \Fx{pusht-ap}\% for AP~\cite{liu2026aim} on identical pairs (\cref{fig:ablations}c). On Cube, gains are more modest (\Fx{cube-k6}\% to \Fx{cube-k2sw}\%, remaining below AP's \Fx{cube-ap}\%) because \Dg{cube-never} of \Dg{cube-fail} navigation failures never initiate contact with the cube; a recorded trajectory alone cannot synthesize a delicate grasp. On medium mazes, the contact setting remains strong at \Fx{pmm-k2sw}\% on PointMaze-Medium and \Fx{tworoom-k2sw}\% on Two-Room, trading off slight efficiency on stitched routes (\Fx{pmmx-k2sw}\% versus \R{pmm-ours-x}\%) and short-range Reacher (\Fx{reacher-k2sw}\% versus \Dg{reacher-ours}\%).

\subsection{Why Is Each Component Needed?}
\label{sec:ablation}

Ablating our three components one by one confirms their distinct roles (\cref{fig:ablations}). On PointMaze-Medium, changing the target alone accounts for nearly the entire gain (\cref{fig:ablations}a): aiming LeWM's unchanged search at our anchors lifts success from \Q{fa-pmm-released}\% to \Q{fa-pmm-gah5}\%, while aiming our closest-approach search at the distant goal stays at \Q{fa-pmm-relmin}\%. Heavier search toward the goal cannot replace anchors either: doubling the horizon and more than tripling CEM samples leaves goal-aimed planning at \R{x22-pmm-best}\% on PointMaze-Medium and \R{x22-tworoom-best}\% on Two-Room, far below default anchors in the same runs (\R{x22-pmm-ours}\% and \R{x22-tworoom-ours}\%, \cref{tab:search}b). On Two-Room, both target and closest-approach search contribute (\Q{fa-tworoom-gah5}\% to \Q{fa-tworoom-ours}\%). Finally, on Push-T, the switch cost $\lambda{=}3$ is the decisive contact ingredient: shrinking the anchor to $K{=}3$ without $\lambda{=}3$ stays at \Fx{pusht-k3}\%, whereas adding $\lambda{=}3$ jumps immediately to \Fx{pusht-k3sw}\% (\cref{fig:ablations}c).

\subsection{Which Design Choices Matter?}
\label{sec:design}

Anchor span $K$ balances directional signal against progress per decision (\cref{fig:ablations}b). Placing anchors too far ($K{=}12$) reintroduces range failure, dropping PointMaze-Medium from \R{span-pmm-6}\% at $K{=}6$ to \R{span-pmm-12}\%, whereas $K{=}2$ stalls under fixed 5-block navigation execution at \R{span-pmm-2}\% (resolved by short-horizon closest-approach execution in \cref{tab:exec}b). Optimal span increases with horizon $H$, from $K{=}\R{x21-pmm-best-h5}$ at $H{=}5$ to $K{=}\R{x21-pmm-best-h10}$ at $H{=}10$, with $K{=}6$ staying within \R{x21-gap6} points of optimal across the entire grid (\cref{tab:spanh}a). Cross-episode neighbour edges are indispensable: removing them collapses PointMaze-Medium from \R{abl-full}\% to \R{abl-knn0}\%, while varying $k \in \{5,10,20\}$ barely changes results (\cref{fig:ablate}b).

\begin{figure}[t]
\centering
\begin{minipage}[t]{0.475\textwidth}
\centering
\includegraphics[width=\linewidth]{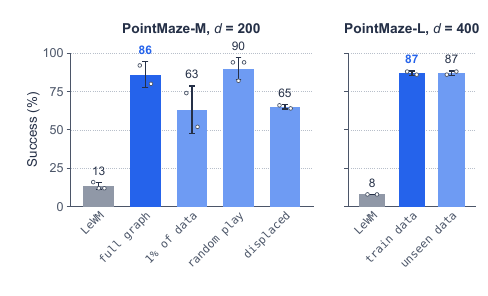}
\caption{\textbf{Little and imperfect data suffice.} Navigation setting on PointMaze-Medium ($d{=}200$: 100\% vs.\ 1\% data, random-play graph, mid-episode kidnapping) and PointMaze-Large ($d{=}400$: training vs.\ unseen dataset). Bars show means with standard deviations and single-seed dots.}
\label{fig:robustbars}
\end{minipage}\hfill
\begin{minipage}[t]{0.475\textwidth}
\centering
\includegraphics[width=\linewidth]{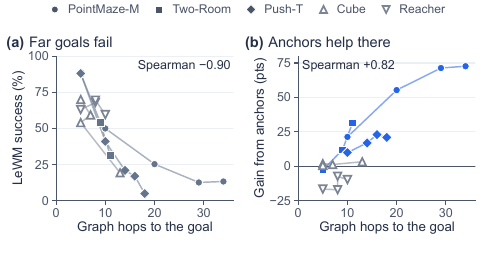}
\caption{\textbf{Graph distance predicts where anchors help.} One point per environment and distance across all settings. \textbf{(a)}~LeWM success drops sharply as goal graph distance grows. \textbf{(b)}~Gain of Graph Anchors over LeWM + min-over-plan tracks goal graph distance.}
\label{fig:predict}
\end{minipage}
\end{figure}

Graph Anchors are also remarkably robust to scarce or shifted data (\cref{fig:robustbars} and \cref{tab:stress}). On PointMaze-Medium $d{=}200$, keeping just 1\% of training episodes still reaches \Q{frac-0.01-ours}\%, and building the graph purely from random play reaches \R{rand-ours-rand-200}\%. On PointMaze-Large $d{=}400$, indexing a separate unseen dataset never used during world-model training still achieves \Pf{pml-sd-400}\%. Finally, mid-episode displacement (kidnapping) requires no special recovery logic: replanning re-anchors from the new observation automatically, shifting success only from \Q{st-kidref-pmm-ours}\% to \Q{st-kid-pmm-ours}\%.

\subsection{Diagnosing Training-Free World-Model Planners (AP, NGC, Metro-WM)}
\label{sec:frozenplanners}

Our diagnosis identifies four practical requirements for training-free long-horizon visual planning: (i)~a reliable short-range controller, (ii)~a sparse graph scaling to 100\% of frames without subsampling shortcuts, (iii)~preserved recorded transitions where contact matters, and (iv)~target-search decoupling without oracle step counts or fragile per-task horizon coupling. \Cref{tab:taxonomy} maps prior graph-guided policies (GAS~\cite{baek2025graph}, TTGS~\cite{opryshko2025test}) and training-free world-model planners (AP~\cite{liu2026aim}, NGC~\cite{you2026latent}, Metro-WM~\cite{lee2026metro}) onto these exact axes. We re-implemented AP, NGC, and Metro-WM inside our shared evaluation framework on identical frozen checkpoints, 5-step training nodes, seeds, and $2d$ budgets; \cref{tab:fidelity} verifies that all three match their published results within 9 points despite our tighter $2d$ budget.

\begin{table}[t]
\centering
\caption{\textbf{Diagnostic taxonomy of experience-guided planners.} Each prior method drops at least one axis identified by our diagnosis. GAS and TTGS rely on learned value or policy heads that saturate or collapse on visual mazes. AP omits a graph and requires an oracle step count $d$. NGC discards temporal transitions and stores an $O(N^2)$ distance table. Metro-WM prices cross-episode bridges below recorded transitions and scores only the final step of fixed-$h$ plans. Graph Anchors satisfies all six axes.}
\label{tab:taxonomy}
\footnotesize
\renewcommand{\arraystretch}{1.15}
\setlength{\tabcolsep}{5pt}
\resizebox{\textwidth}{!}{%
\begin{tabular}{@{}lcccccc@{}}
\toprule
Method & Training-Free & Local Controller & Full-Data Sparse Graph & No Oracle $d$ & Contact-Safe & Pass-Through \\
\midrule
GAS~\cite{baek2025graph} & $\times$ & Policy (HILP) & $\times$ & $\checkmark$ & $\times$ & $\times$ \\
TTGS~\cite{opryshko2025test} & $\times$ & Policy (GCIVL) & $\times$ & $\checkmark$ & $\times$ & $\times$ \\
AP~\cite{liu2026aim} & $\checkmark$ & Frozen LeWM & $\times$ & $\times$ & $\checkmark$ & $\times$ \\
NGC~\cite{you2026latent} & $\checkmark$ & Frozen LeWM & $\times$ & $\checkmark$ & $\times$ & $\times$ \\
Metro-WM~\cite{lee2026metro} & $\checkmark$ & Frozen LeWM & $\checkmark$ & $\checkmark$ & $\times$ & $\times$ \\
\rowcolor{oursrow}\textbf{Graph Anchors (Ours)} & $\checkmark$ & Frozen LeWM & $\checkmark$ & $\checkmark$ & $\checkmark$ & $\checkmark$ \\
\bottomrule
\end{tabular}}
\end{table}

\begin{table}[t]
\centering
\caption{\textbf{Head-to-head comparison of training-free planners on identical frozen LeWM checkpoints, pairs, and $2d$ budgets} (success, \%). AP receives the oracle step count $d$; Metro-WM tunes $(h,H)$ per task on the evaluation pairs. PointMaze-Large runs use the shared \texttt{lewm\_pml} checkpoint (seed 0; original checkpoint in brackets). Mean (far) averages PM-M $d{=}200$, Two-Room $d{=}75$, and Push-T $d{=}150$. A single unchanged Graph Anchors contact setting ($K{=}2, \lambda{=}3$) leads Two-Room, Push-T, and the three-task mean; navigation leads PointMaze-Large by +52 to +70 points.}
\label{tab:contemp}
\footnotesize
\renewcommand{\arraystretch}{1.10}
\setlength{\tabcolsep}{4pt}
\resizebox{\textwidth}{!}{%
\begin{tabular}{@{}lcccccccc@{}}
\toprule
 & \multicolumn{2}{c}{PointMaze-M} & Two-Room & \multicolumn{2}{c}{Push-T} & Mean & \multicolumn{2}{c}{PointMaze-L} \\
\cmidrule(lr){2-3}\cmidrule(lr){4-4}\cmidrule(lr){5-6}\cmidrule(lr){7-7}\cmidrule(l){8-9}
Method & 100 & 200 & 75 & 50 & 150 & (far) & 400 & Stitched \\
\midrule
LeWM~\cite{maes2026leworldmodel} & \R{pmm-released-100} & \R{pmm-released-200} & \R{tworoom-released-75} & \R{pusht-released-50} & \R{pusht-released-150} & 16.3 & 10 & 6 \\
AP~\cite{liu2026aim} (uses true $d$) & \Cc{ap-pmm-100} & \Cc{ap-pmm-200} & \Cc{ap-tworoom-75} & \Cc{ap-pusht-50}$^\dagger$ & \Fx{pusht-ap} & 80.7 & 18 & 9 \\
NGC~\cite{you2026latent} & \textbf{\Cc{ngc-pmm-100}} & \textbf{\Cc{ngc-pmm-200}} & \Cc{ngc-tworoom-75} & \Cc{ngc-pusht-50} & \Cc{ngc-pusht-150} & 70.0 & 24 & 20 \\
Metro-WM-style~\cite{lee2026metro} (tuned) & \Cc{metro-pmm-100} & \Cc{metro-pmm-200} & \Cc{metro-tworoom-75} & \Cc{metro-pusht-50} & \Cc{metro-pusht-150} & 76.0 & 6 & 6 \\
\rowcolor{oursrow}Graph Anchors (nav.) & \R{pmm-ours-100} & \R{pmm-ours-200} & \R{tworoom-ours-75} & \R{pusht-ours-50} & \R{pusht-ours-150} & 69.3 & \textcolor{bestred}{\textbf{76}} [87] & \textcolor{bestred}{\textbf{45}} [66] \\
\rowcolor{oursrow}\textbf{Graph Anchors (contact)} & \Fx{pmm-k2sw-100} & \Fx{pmm-k2sw} & \textcolor{bestred}{\textbf{\Fx{tworoom-k2sw}}} & \textbf{\Fx{pusht-k2sw-50}} & \textcolor{bestred}{\textbf{\Fx{pusht-k2sw}}} & \textcolor{bestred}{\textbf{88.3}} & 13 & 9 \\
\bottomrule
\end{tabular}}
\end{table}

\begin{figure}[t]
\centering
\includegraphics[width=\textwidth]{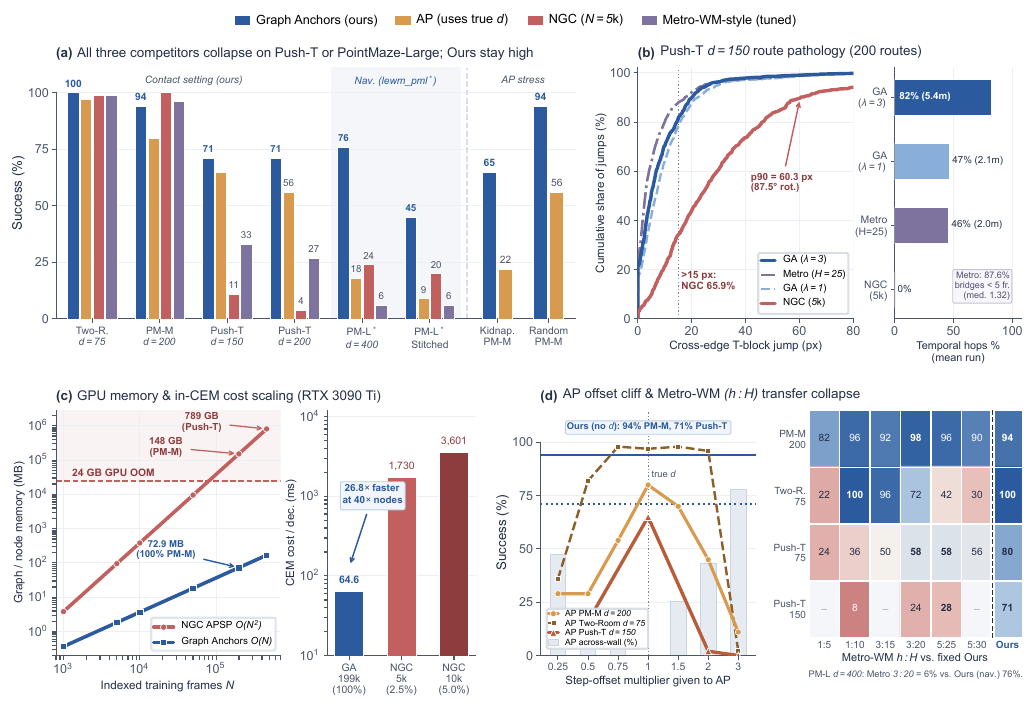}
\caption{\textbf{Quantitative failure profile, scaling wall, and fragility of AP, NGC, and Metro-WM compared to Graph Anchors.} \textbf{(a)}~Four-way closed-loop success across six long-range benchmarks (contact setting on Two-Room, PointMaze-Medium, and Push-T; navigation on PointMaze-Large \texttt{lewm\_pml} and AP stress tests). \textbf{(b)}~Push-T $d{=}150$ route continuity over 200 held-out routes: cross-edge block-jump CDF (left) and recorded temporal share with unbroken push length (right). \textbf{(c)}~RTX 3090 Ti GPU memory and per-decision CEM cost latency: Graph Anchors indexes $40\times$ more nodes at $26.8\times$ lower latency than NGC. \textbf{(d)}~AP's sharp drop under mis-specified step distance $d$ tracking across-wall retrieval (left), and Metro-WM's full $(h{:}H)$ sweep heatmap (right) where every column has collapse cells versus Graph Anchors' single unchanged contact column.}
\label{fig:compquant}
\end{figure}

\begin{figure}[t]
\centering
\includegraphics[width=\textwidth]{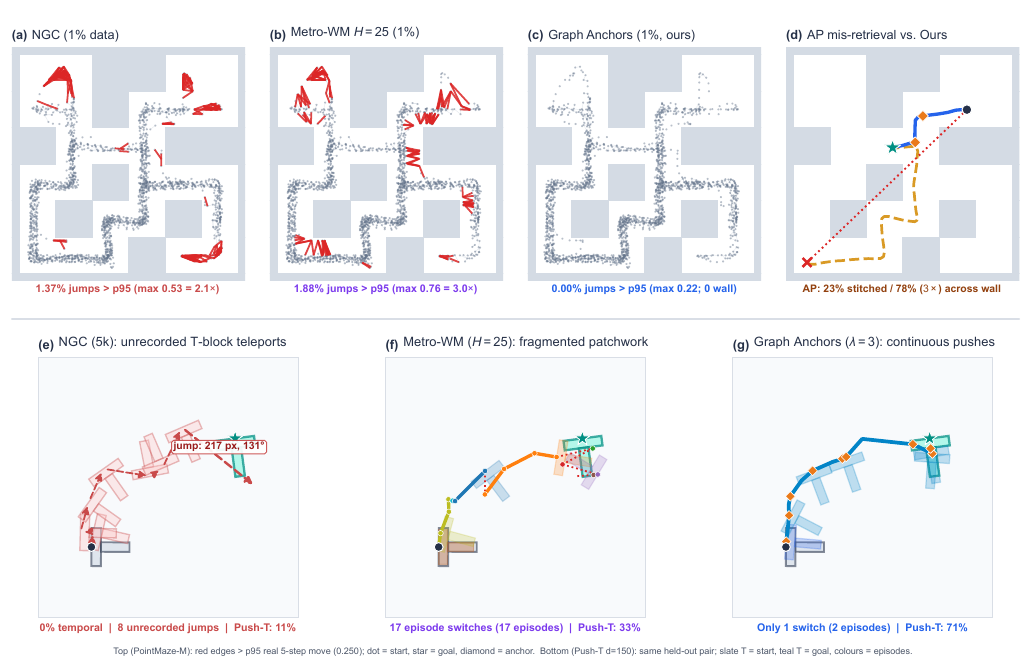}
\caption{\textbf{Visual breakdown of planner failures on real encoded datasets.} \textbf{Top (PointMaze-Medium, exact $8{\times}8$ walls):} On 1\% sparse data, NGC \textbf{(a)} and Metro-WM ($H{=}25$) \textbf{(b)} admit red long-jump shortcuts exceeding a real 5-step move (1.37\% up to $2.1\times$ and 1.88\% up to $3.0\times$), whereas Graph Anchors' 1-block $\tau$ \textbf{(c)} admits 0.00\% shortcuts. \textbf{(d)}~On a stitched pair, AP retrieves a start 6.07 units away across solid walls (red dotted), while Graph Anchors stitches cleanly around the wall (blue path, orange diamonds). \textbf{Bottom (Push-T $d{=}150$, identical held-out pair with true T-block poses):} NGC \textbf{(e)} teleports the block by up to 217\,px and 131$^\circ$; Metro-WM \textbf{(f)} breaks the path into 17 episode switches (colours); Graph Anchors ($\lambda{=}3$) \textbf{(g)} follows continuous pushes with just 1 switch, lifting Push-T $d{=}150$ from 11\% and 33\% up to 71\%.}
\label{fig:companatomy}
\end{figure}

As shown in \cref{tab:contemp} and \cref{fig:compquant}a, all four methods resolve basic range failure on medium mazes, yet each alternative breaks down precisely where it drops a diagnostic axis of \cref{tab:taxonomy}:

\paragraph{AP: trajectory retrieval without a graph.}
Staying inside one recorded episode preserves contact continuity, explaining AP's strong Push-T and Cube scores when given the exact step distance $d$. However, omitting a graph forces AP to know $d$ in advance and limits it to routes covered by a single training episode. When $d$ is off by $2$--$3\times$ (\cref{fig:compquant}d, \cref{tab:comp_mech}c), the eligible episode bank shrinks drastically and 21--78\% of retrieved starts fall across solid maze walls, dropping PointMaze-Medium from \Cc{ap-pmm-200}\% to \Cc{apx3-pmm-200}\%, Two-Room from \Cc{ap-tworoom-75}\% to \Cc{apx3-tworoom-75}\%, and Push-T from \Fx{pusht-ap}\% to 16\% ($0.5\times$) and 0\% ($3\times$). AP breaks similarly whenever single episodes cannot span the task: stitched goals (23.3\% across-wall starts, \cref{fig:companatomy}d), PointMaze-Large (18\% and 9\% versus our 76\% and 45\%), mid-episode kidnapping (22\% vs.\ 65\%), random-play data (56\% vs.\ 94\%), and Reacher (34\% vs.\ 68\%).

\paragraph{NGC: undirected manifold cost under $O(N^2)$ subsampling.}
Replacing Euclidean distance with shortest-path length over 5{,}000 undirected snapshots solves medium mazes (\Cc{ngc-pmm-200}\% on PointMaze-Medium), but violates two axes. First, discarding directed temporal edges creates a clear \emph{geometry-vs.-executability gap} on Push-T (\cref{fig:compquant}b, \cref{tab:comp_mech}a): NGC scores higher offline object-pose ranking accuracy ($\rho_{\text{obj}}{=}0.715$) than Graph Anchors (0.649) because undirected snapshots hop across state space without obeying pushing physics, yet 65.9\% of its route hops teleport the block over 15\,px (up to 217\,px / 131$^\circ$ in \cref{fig:companatomy}e), collapsing closed-loop Push-T to \Cc{ngc-pusht-150}\% at $d{=}150$ and \Ed{ngc-200}\% at $d{=}200$. Second, storing an $N{\times}N$ distance table and re-attaching every CEM candidate hits an $O(N^2)$ wall (\cref{fig:compquant}c): on an RTX 3090 Ti, $N{=}5{,}000$ already takes 95.4\,MB and 1{,}729.9\,ms per decision, while full datasets would require 148\,GB on PointMaze-Medium and 789\,GB on Push-T. Capping $N{=}5{,}000$ on 200k-frame PointMaze-Large forces \textbf{9.51\%} of edges into false long-jump shortcuts up to 22.8 maze units, collapsing NGC to 24\% / 20\%. Graph Anchors indexes all 199k--200k nodes in 72.9\,MB and 64.6\,ms per decision, running $26.8\times$ faster.

\paragraph{Metro-WM: underpriced cross-episode bridges and terminal-$h$ planning.}
Metro-WM builds a sparse frame graph like ours, directly isolating why our three design choices matter. First, checking 80{,}000 Push-T bridges shows that \textbf{87.6\%} are priced below a real 5-frame temporal edge (median 1.32 frames; 100\% on PointMaze-Large). Dijkstra therefore abandons recorded pushes immediately (\cref{fig:compquant}b, \cref{fig:companatomy}f), splitting Push-T into 7.03 switches per 6 moves and stalling success at \Cc{metro-pusht-150}\% ($d{=}150$) and \Ed{metro-200}\% ($d{=}200$). Our switch cost $\lambda{=}3$ cuts switches $4.2\times$ to 1.68, stretches unbroken pushes $2.7\times$ to 5.40 moves (\cref{fig:companatomy}g), and lifts Push-T $d{=}150$ to \Fx{pusht-k2sw}\%. Second, scaling bridge radius with budget $H$ admits sparse-data shortcuts: on 1\% PointMaze-Medium data (\cref{fig:companatomy}b), 1.88\% of bridges exceed a real 5-step move up to $3.0\times$ (dropping success from 98\% to 76--78\%), versus 0.00\% shortcuts for our fixed 1-block $\tau$. Third, executing fixed-$h$ terminal-cost plans prevents any $(h,H)$ pair from transferring across tasks (\cref{fig:compquant}d): Two-Room's best ($1{:}10$, 100\%) drops to 8\% on Push-T $d{=}150$, while PointMaze-Medium's best ($3{:}20$, 98\%) drops to 72\% on Two-Room, 33\% on Push-T, and 6\% on PointMaze-Large $d{=}400$. Full reproduction checks, Push-T distance curves through $d{=}200$, and complete numbers appear in \cref{app:frozenplanners}.

\subsection{When Does It Help?}
\label{sec:where}\label{sec:negative}\label{sec:beyond}

Our diagnosis also predicts where Graph Anchors help before running any evaluation. First, aiming at a nearby target restores positive closed-loop drift. Counting progress in hops on a true-state reference graph, fewer than half of LeWM's decisions end closer to the goal on all three traced benchmarks. Adding Graph Anchors raises improving decisions to \Pf{dr-pmm-ours}\% on PointMaze-Medium, \Pf{dr-tworoom-ours}\% on Two-Room, and \Pf{dr-pusht-ours}\% on Push-T (\cref{tab:where} and \cref{app:theory}). As \cref{prop:drift} proves, positive net drift changes expected hitting time from quadratic $\Omega(D^2)$ random wandering to linear $O(D)$ progress.

\begin{table}[t]
\centering
\caption{\textbf{Diagnosis and closed-loop outcome per environment.} $\rho(1)$ and $\rho(6)$ report directional ranking one and six hops from the goal. Hops gives median goal graph distance. LeWM and Ours report success (\%) at the longest distance ($^c$contact setting). Closer gives the percentage of decisions ending closer to the goal (LeWM $\to$ Ours).}
\label{tab:where}
\small
\renewcommand{\arraystretch}{1.08}
\setlength{\tabcolsep}{7pt}
\begin{tabular}{@{}lccclccc@{}}
\toprule
Environment & $\rho(1)$ & $\rho(6)$ & Hops & Failure & LeWM & \textbf{Ours} & Closer (\%) \\
\midrule
PM-M 200 & 0.99 & 0.59 & 34 & range & 13 & \textbf{85} & 48$\to$72 \\
Two-Room 75 & 0.99 & 0.50 & 11 & range & 31 & \textbf{93} & 33$\to$74 \\
Reacher 150 & 1.00 & 0.56 & 10 & -- & 60 & \textbf{68} & -- \\
Push-T 150 & 0.91 & 0.74 & 18 & focus & 5 & \textbf{71}$^c$ & 31$\to$53 \\
Cube 150 & 0.99 & 0.77 & 13 & focus & 19 & \textbf{39}$^c$ & -- \\
 
\bottomrule
\end{tabular}
\end{table}

Second, offline goal graph distance strongly predicts both failure and anchor gain across all environments and distances (\cref{fig:predict}). It correlates with LeWM success at Spearman \Pf{pred-lewm} across \Pf{pred-n} settings, and with the gain of Graph Anchors over LeWM + min-over-plan at Spearman \Pf{pred-gain} across \Pf{pred-ngain} settings. \Cref{tab:where} pairs each offline diagnosis directly with its closed-loop outcome: where range dominates, navigation anchors solve the task; where focus dominates, contact anchors solve Push-T and improve Cube; and on compact Reacher (goals staying within 10 hops at every distance), LeWM already reaches \Pf{bey-reacher-lewm}\% and intermediate anchors only add small detours (removable via optional gated goal handoff in \cref{app:handoff}).

% ==============================================================================
\section{Conclusion}
\label{sec:conclusion}

The visual goal cost of a latent world model ranks local progress accurately only when its target is nearby and lies on a recorded route. Measuring this directly separates two root failures, \emph{range} and \emph{focus}, explains why distant goals and contact tasks collapse, and predicts where retrieved subgoals help. Graph Anchors put this finding to work without modifying or retraining the world model, using a sparse full-dataset graph, cross-episode switch penalties, and closest-approach scoring to remove range failure across mazes and focus failure on Push-T while avoiding the step-horizon fragility, quadratic scaling wall, and contact fragmentation of alternative training-free planners.

\paragraph{Limitations.}
Graph Anchors rely on recorded trajectories covering the route and use two task-level settings for navigation versus contact; the contact setting trades off efficiency on long stitched mazes and short-range Reacher. Because a recorded route cannot synthesize a delicate grasp when the local world model misses initial contact, Cube remains below AP, and our Meta-World demonstration test serves as an exact-dynamics upper bound rather than a deployable manipulation controller. Extending training-free graph guidance to dexterous grasping and open-world video world models remains an exciting direction for future work.

% ==============================================================================
\bibliographystyle{plainnat}
\bibliography{references}

% ==============================================================================
\clearpage
\appendix
\renewcommand{\thefigure}{A\arabic{figure}}
\renewcommand{\thetable}{A\arabic{table}}
\renewcommand{\theequation}{A\arabic{equation}}
\renewcommand{\thealgorithm}{A\arabic{algorithm}}
\setcounter{figure}{0}
\setcounter{table}{0}
\setcounter{equation}{0}
\setcounter{algorithm}{0}

\section*{Appendix Table of Contents}
\vspace{0.3em}
\noindent\rule{\textwidth}{0.6pt}
\vspace{0.4em}

\noindent\textbf{A}\quad Extended Goal-Cost Diagnosis \dotfill \pageref{app:diagnosis}\\[3pt]
\hspace*{2em}A.1\quad Range Failure Across Encoders and Epochs \dotfill \pageref{app:range}\\[2pt]
\hspace*{2em}A.2\quad Focus Failure Across Encoders and Pushing \dotfill \pageref{app:focus}\\[2pt]
\hspace*{2em}A.3\quad Formal Analysis and Closed-Loop Drift \dotfill \pageref{app:theory}\\[5pt]
\noindent\textbf{B}\quad Graph Anchors: Algorithm and Implementation \dotfill \pageref{app:method}\\[3pt]
\hspace*{2em}B.1\quad Algorithm and Walkthrough \dotfill \pageref{app:walkthrough}\\[2pt]
\hspace*{2em}B.2\quad Implementation Details and Graph Quality \dotfill \pageref{app:impl}\\[5pt]
\noindent\textbf{C}\quad Extended Results and Ablations \dotfill \pageref{app:results}\\[3pt]
\hspace*{2em}C.1\quad All Distances, Retraining Sensitivity, and Paired Tests \dotfill \pageref{app:seeds}\\[2pt]
\hspace*{2em}C.2\quad Contact Setting, Switch Cost, and Goal Handoff \dotfill \pageref{app:handoff}\\[2pt]
\hspace*{2em}C.3\quad Ablations, Scaling, and Robustness Stress Tests \dotfill \pageref{app:ablations}\\[5pt]
\noindent\textbf{D}\quad Additional Baselines and Planner Details \dotfill \pageref{app:planners}\\[3pt]
\hspace*{2em}D.1\quad World-Model Baselines and Graph-Guided Policies (GAS, TTGS) \dotfill \pageref{app:gasttgs}\\[2pt]
\hspace*{2em}D.2\quad Implementation Fidelity, Push-T Scaling, and Full Metrics for AP, NGC, Metro-WM \dotfill \pageref{app:frozenplanners}\\[5pt]
\noindent\textbf{E}\quad Full Per-Evaluation Results, Qualitative Rollouts, and Reproducibility \dotfill \pageref{app:limits}\\[4pt]

\noindent\rule{\textwidth}{0.6pt}
\vspace{0.5em}

% ==============================================================================
\section{Extended Goal-Cost Diagnosis}
\label{app:diagnosis}\label{app:phenomenon}\label{app:graphs}\label{app:focusmain}

On the standard 25-step LeWM evaluation setup summarized in \cref{tab:published}, both visual world-model planners and offline policies solve most goals easily. Once the goal is placed further away, however, those same planners break down sharply. As shown in \cref{fig:results} and \cref{tab:compare}, LeWM drops to \R{pmm-released-200}\% at $d{=}200$ on PointMaze-Medium, \R{tworoom-released-75}\% at $d{=}75$ on Two-Room, \R{pusht-released-150}\% at $d{=}150$ on Push-T, and \R{pml-released-400}\% at $d{=}400$ on PointMaze-Large.

Simply throwing more search compute at the distant goal does not rescue the planner. Doubling the planning horizon and more than tripling the number of CEM samples lifts LeWM to only \R{x22-pmm-best}\% on PointMaze-Medium and \R{x22-tworoom-best}\% on Two-Room in \cref{tab:search}b. Even switching to our closest-approach scoring while still aiming at the distant goal stays at \Q{fa-pmm-relmin}\% on PointMaze-Medium in \cref{tab:factorial}a. Tracing closed-loop rollouts in true-state graph hops reveals what goes wrong at runtime. Fewer than half of LeWM's decisions actually end closer to the goal, with only \Pf{dr-pmm-released}\% improving on PointMaze-Medium, \Pf{dr-tworoom-released}\% on Two-Room, and \Pf{dr-pusht-released}\% on Push-T, as documented in \cref{tab:where} and \cref{tab:driftt8}.

In this appendix we expand our diagnosis of the two root causes inside the latent goal cost. \Cref{app:range} tests \emph{range} across six image representations and ten training epochs, \cref{app:focus} tests \emph{focus} across five encoders and nine Meta-World tasks, and \cref{app:theory} formalizes how local ranking translates into closed-loop drift.

\begin{table}[t]
\centering
\small
\renewcommand{\arraystretch}{1.08}
\setlength{\tabcolsep}{7pt}
\caption{\textbf{Published success (\%) on the standard LeWM benchmark.} Results are taken from Fig.~6 of~\cite{maes2026leworldmodel}, where goals lie 25 steps ahead with a 50-step budget. These numbers provide short-horizon context only, since none of these methods was evaluated at long range in that work.}
\label{tab:published}
\begin{tabularx}{0.86\textwidth}{@{}Xcccccc@{}}
\toprule
Environment & LeWM & PLDM & DINO-WM & GCBC & GCIQL & GCIVL \\
\midrule
Push-T & 96 & 78 & 74 & 75 & 20 & 33 \\
Two-Room & 87 & 97 & 100 & 100 & 100 & 100 \\
OGBench-Cube & 74 & 65 & 86 & 84 & 64 & 56 \\
 
\bottomrule
\end{tabularx}
\end{table}

% ------------------------------------------------------------------------------
\subsection{Range Failure Across Encoders and Epochs}
\label{app:ranktest}\label{app:range}\label{app:epochs}\label{app:checks}

\paragraph{Measuring directional ranking.}
To test whether a latent distance actually guides an agent toward a distant goal, we ask a simple local question. Suppose the agent is at a recorded frame $v$ that lies $h$ graph hops from a goal frame $g$, and consider two neighbours $u$ and $u'$ of $v$ where one is genuinely closer to $g$ than the other. We count a latent distance as correct on this pair if it assigns a smaller cost to whichever neighbour is truly closer to $g$. The ranking accuracy $\rho(h)$ is the share of pairs ordered correctly. A value of $\rho{=}1$ means the distance always points downhill toward the goal, whereas $\rho{=}\tfrac12$ carries no directional information and is no better than flipping a coin.

To avoid any circularity with a model's own latent graph, all cross-encoder and training-epoch checks below measure hops on a fixed \emph{reference graph} built strictly from ground-truth positions. This reference graph connects consecutive recorded frames along each episode and adds edges between frames closer in physical position than the median five-step move. Every encoder is tested on the exact same five-step frames drawn from a seeded fifth of the episodes, over 200 goals with up to four frames per goal and hop. Shaded bands in the plots mark 95\% bootstrap confidence intervals over goals.

\begin{figure}[t]
\centering
\begin{minipage}[t]{0.475\textwidth}
\centering
\includegraphics[width=\linewidth]{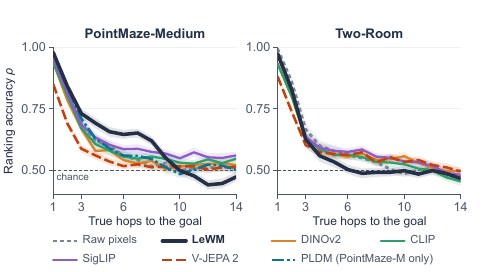}
\caption{\textbf{Every image encoder loses its direction a few hops from the goal.} Ranking accuracy $\rho(h)$ against true-position reference hops for six representations on identical frames and 200 goals per environment, together with one-epoch PLDM on PointMaze-Medium. Bands show 95\% bootstrap intervals and dashed line marks chance.}
\label{fig:e1range}
\end{minipage}\hfill
\begin{minipage}[t]{0.475\textwidth}
\centering
\includegraphics[width=\linewidth]{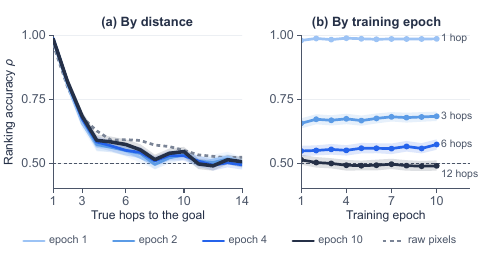}
\caption{\textbf{Training longer does not extend range.} LeWM checkpoints from epochs 1 to \Ec{neps} of one PointMaze-Large training run on identical frames and 200 goals, alongside raw pixels as a control. Panel~(a) plots $\rho(h)$ against true-position hops, and panel~(b) tracks $\rho$ at 1, 3, 6, and 12 hops across epochs.}
\label{fig:e2epochs}
\end{minipage}
\end{figure}

\paragraph{Six encoders, including raw pixels.}
\Cref{fig:diagnosis}b evaluates LeWM, PLDM, and frozen DINOv2. Here we expand that comparison to six image representations on identical frames and goals: raw pixels average-pooled to $32{\times}32$ with no learned network at all, alongside frozen DINOv2-S, CLIP ViT-B/16, SigLIP-B/16, video-trained V-JEPA~2 ViT-L, and LeWM. For each vision backbone, we compute squared distance over patch features pooled to $4{\times}4$, matching how the planners score images.

As shown in \cref{fig:e1range}, all six representations follow the exact same curve. One hop from the goal, every encoder ranks local moves reliably, with $\rho(1)$ spanning \Ea{pmm-r1lo} to \Ea{pmm-r1hi} on PointMaze-Medium and \Ea{tworoom-r1lo} to \Ea{tworoom-r1hi} on Two-Room. Within just a few hops, however, that directional signal fades toward chance. Six hops from the goal, $\rho(6)$ falls to \Ea{pmm-r6lo}--\Ea{pmm-r6hi} on PointMaze-Medium and \Ea{tworoom-r6lo}--\Ea{tworoom-r6hi} on Two-Room, crossing below 0.55 between hops \Ea{pmm-hlo} and \Ea{pmm-hhi}. At twelve hops on PointMaze-Medium, LeWM drops to $\rho{=}\Ea{pmm-lewm-12}$ and even leans slightly away from the goal. PLDM behaves identically on the same frames, dropping from $\rho(1){=}\Ea{pmm-pldm-1}$ to $\rho(6){=}\Ea{pmm-pldm-6}$ after one epoch and $\Ea{pmm-pldm5-6}$ after five epochs.

The raw-pixel control highlights why this happens. Because raw pixel distance decays just like every learned encoder, including video-trained V-JEPA~2, short range is not a flaw of a specific architecture or training loss. It is a basic property of judging progress by direct distance between two whole images when walls and corridors lie between them. \Cref{fig:e1maps} and \cref{fig:e1hops} show this geometry visually. Far from the goal, every encoder treats frames in distant corridors or behind solid walls as just as close to the goal as frames sitting on the true path.

\paragraph{Training longer does not extend range.}
One might wonder whether short range simply reflects an undertrained world model. To check this, we run the exact same ranking test across every epoch from 1 to \Ec{neps} of our second PointMaze-Large LeWM training run from \cref{app:seeds}, keeping raw pixels as a control. As shown in \cref{fig:e2epochs}, extended training does not fix the decay. One hop from the goal, every checkpoint is already highly accurate, moving from \Ec{ep1-1} at epoch 1 to \Ec{ep10-1} at epoch \Ec{neps}. Six hops away, $\rho(6)$ shifts only from \Ec{ep1-6} to \Ec{ep10-6}, staying slightly below raw pixels at \Ec{pix-6}. Twelve hops away, accuracy remains right at chance from \Ec{ep1-12} to \Ec{ep10-12}. Over all ten epochs, the hop where ranking drops below 0.55 moves only from hop \Ec{hlo} to hop \Ec{hhi}. Combined with the five-epoch PLDM check above, this confirms that longer training does not extend visual range.

\begin{table}[t]
\centering
\small
\renewcommand{\arraystretch}{1.08}
\setlength{\tabcolsep}{7pt}
\caption{\textbf{Graph-free ground-truth checks.} For a state $k$ action blocks before a goal in the same recorded episode, $\rho_T(k)$ reports how often the latent cost prefers one block forward at $t{+}1$ over one block backward at $t{-}1$ across 4{,}000 states per $k$, where 0.5 is chance. The Spearman columns report rank correlation between latent cost and true steps to the goal locally within 5 blocks and globally over all $k{\le}30$.}
\label{tab:truediag}
\begin{tabularx}{0.82\textwidth}{@{}Xccccc@{}}
\toprule
 & \multicolumn{3}{c}{$\rho_T(k)$} & \multicolumn{2}{c}{Spearman} \\
\cmidrule(lr){2-4}\cmidrule(l){5-6}
Environment & $k{=}3$ & 6 & 10 & local & all \\
\midrule
Push-T & 0.96 & 0.83 & 0.60 & 0.69 & 0.48 \\
Two-Room & 0.62 & 0.56 & 0.56 & 0.26 & 0.27 \\
PointMaze-M & 0.98 & 0.66 & 0.32 & 0.82 & 0.27 \\
Reacher & 0.74 & 0.59 & 0.52 & 0.43 & 0.33 \\
OGBench-Cube & 0.84 & 0.68 & 0.61 & 0.54 & 0.44 \\
 
\bottomrule
\end{tabularx}
\end{table}

\begin{figure}[t]
\centering
\includegraphics[width=0.68\textwidth]{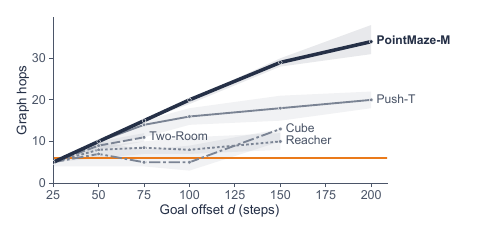}
\caption{\textbf{Goal distance in the graph.} Curves show the median and interquartile range of graph hops between a start and a goal $d$ steps later in the same episode across 200 pairs, with the orange line marking our 6-hop anchor span. On PointMaze-Medium and Push-T, goals move steadily farther in graph hops as $d$ grows, whereas on Reacher they saturate near ten hops.}
\label{fig:hops}
\end{figure}

\paragraph{Graph-free checks on recorded time.}
Because \cref{eq:rho} measures distance using graph hops, it is important to confirm that the range collapse is not an artifact of building a graph. We therefore run a completely graph-free test directly along recorded episodes and report the results in \cref{tab:truediag}. Take a frame at time $t$ and a future goal $k$ blocks ahead at $t{+}k$ within the same episode, testing $k{\ge}2$ across 4{,}000 states per separation. We simply check whether stepping one block forward to $t{+}1$ yields a lower latent distance to the goal than stepping one block backward to $t{-}1$.

The exact same collapse appears without any graph. Close to the goal, $\rho_T$ stays near 1, so the latent cost reliably distinguishes forward from backward. By 10 blocks away, however, $\rho_T$ drops to chance or below on every environment except Cube, falling to \Q{td-pmm-rho10} on PointMaze-Medium. On Cube, $\rho_T$ rises back to 0.77 at 20 blocks because the arm often swings back near an earlier pose during longer trajectories. Spearman correlation with true steps to the goal confirms the same drop, falling on PointMaze-Medium from \Q{td-pmm-sploc} locally within 5 blocks to \Q{td-pmm-spall} globally across $k{\le}30$. Finally, \cref{fig:hops} shows why Reacher behaves differently from the other benchmarks. Even when a Reacher goal lies 150 steps ahead in time, its graph distance plateaus around ten hops inside the compact workspace, whereas PointMaze-Medium and Push-T stretch far past the reliable local range.

\begin{figure}[p]
\centering
\includegraphics[width=0.96\textwidth]{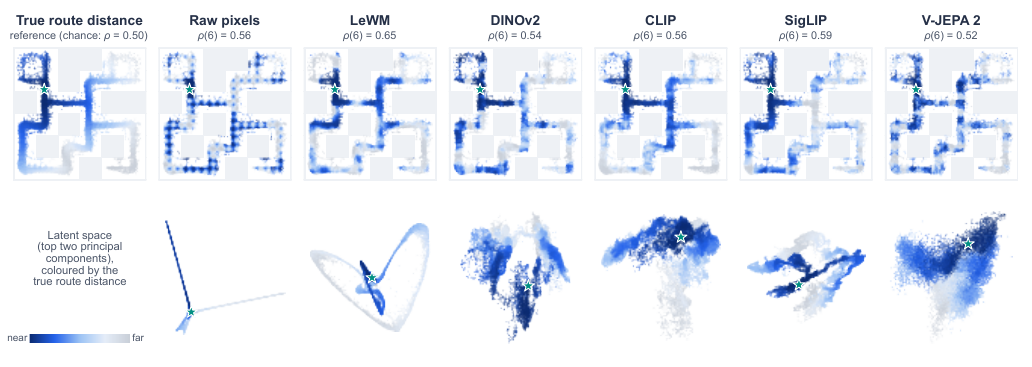}\\[4pt]
\includegraphics[width=0.96\textwidth]{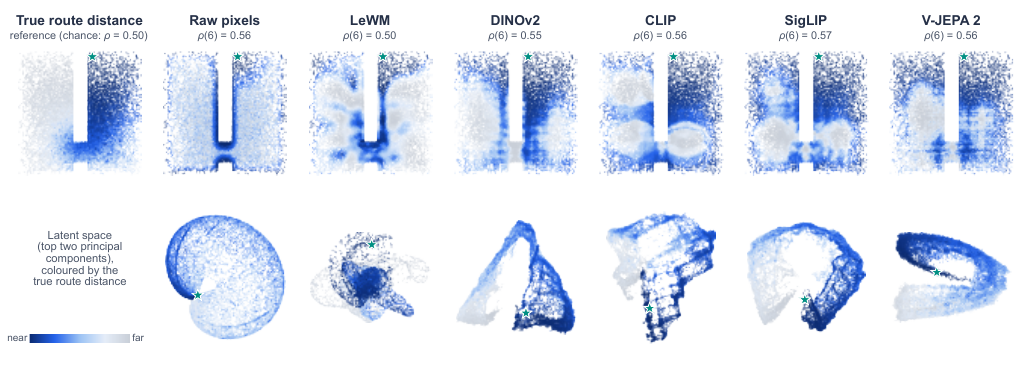}
\caption{\textbf{Where each encoder thinks the goal is near.} The top two rows show PointMaze-Medium and the bottom two rows show Two-Room for one goal marked by a star. In each block, the first row shows true route distance followed by each encoder's latent distance on the same 20k recorded frames plotted as ranks, where darker shades are closer. The second row projects each latent space onto its top two principal components coloured by true route distance, and each title reports $\rho(6)$ over all 200 goals. On Two-Room the goal sits in the right room with only a bottom doorway, yet every encoder treats much of the left room as close to the goal.}
\label{fig:e1maps}
\end{figure}

\begin{figure}[p]
\centering
\includegraphics[width=0.92\textwidth]{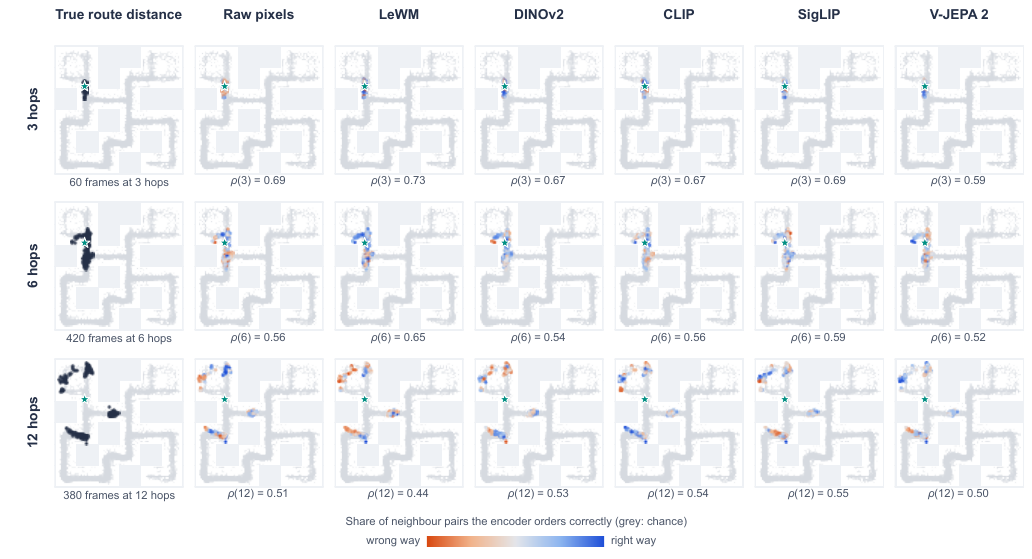}\\[6pt]
\includegraphics[width=0.92\textwidth]{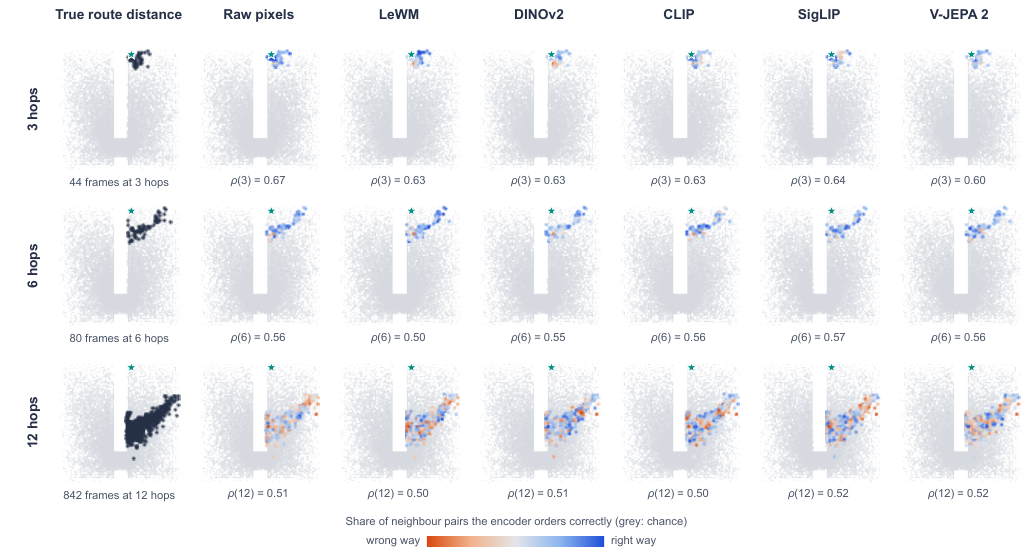}
\caption{\textbf{Frame-by-frame collapse of directional ranking} on the goals of \cref{fig:e1maps}, with PointMaze-Medium on top and Two-Room on the bottom. Each row displays recorded frames located exactly 3, 6, or 12 true hops from the goal, coloured by the fraction of reference-graph neighbour pairs ordered correctly by each encoder. Blue points toward the goal, orange points away, grey marks chance, and each panel lists $\rho(h)$ over all 200 goals. At 3 hops every encoder is mostly blue, but by 12 hops blue and orange mix together randomly so decreasing latent distance is no better than chance.}
\label{fig:e1hops}
\end{figure}

\FloatBarrier
% ------------------------------------------------------------------------------
\subsection{Focus Failure Across Encoders and Pushing}
\label{app:focus}\label{app:pushbias}\label{app:remedyproof}

Short range explains why planners fail on mazes, but Push-T exposes a second obstacle. On Push-T, the latent cost still ranks local moves well six hops from the goal with $\rho(6){=}\Dg{pusht-r6}$, yet closed-loop planning still falls to \R{pusht-released-150}\% at $d{=}150$. This second failure concerns \emph{what} whole-image distance actually weighs.

\begin{figure}[t]
\centering
\includegraphics[width=0.72\textwidth]{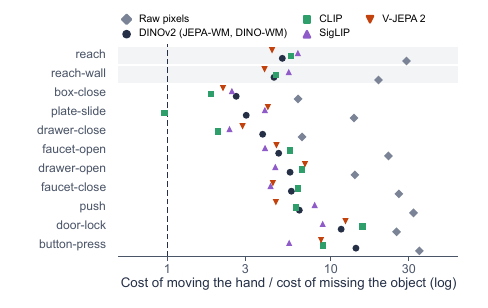}
\caption{\textbf{Every image distance weighs the robot arm over the object.} Bars show the median exchange rate over 30 inverse-kinematics placements per task across five visual representations. This ratio divides the feature distance of shifting only the robot hand by 5\,cm by the distance of leaving only the object off target by its success threshold. Shaded rows mark reach tasks where positioning the arm is the entire task. Anywhere to the right of the dashed line, the cost prefers fixing a small hand error over completing the task.}
\label{fig:exchange}
\end{figure}

\begin{table}[t]
\centering
\footnotesize
\renewcommand{\arraystretch}{1.12}
\setlength{\tabcolsep}{3.5pt}
\caption{\textbf{Meta-World closed-loop success (\%) under exact dynamics.} Each entry averages 12 identical test episodes per task. The two controls verify baseline behavior when no focus failure exists, since \task{drawer-close} is already solved by the released cost and \task{reach-wall} depends only on the arm. Aiming the unchanged released cost at a demonstration frame 20 steps ahead serves as a privileged upper bound and solves \Q{mw-demo-solved} of \Q{mw-demo-n} object episodes, averaging \Q{mw-demo-mean}\% with 50\% on \task{push}. Controls were not re-run for this upper bound.}
\label{tab:mw}
\resizebox{\textwidth}{!}{%
\begin{tabular}{@{}lccccccccc@{}}
\toprule
 & \multicolumn{6}{c}{Object tasks} & & \multicolumn{2}{c}{Controls} \\
\cmidrule(lr){2-7}\cmidrule(l){9-10}
Goal cost & \task{button} & \task{push} & \task{drawer-open} & \task{plate-slide} & \task{faucet-open} & \task{window-open} & Mean & \task{drawer-close} & \task{reach-wall} \\
\midrule
Released latent cost~\cite{terver2025drives} & 0 & 0 & 0 & 0 & 0 & 0 & 0.0 & 100 & 96 \\
TEMPO temporal distance~\cite{moukheiber2026long} & 0 & 0 & 0 & 0 & 0 & 0 & 0.0 & 100 & 42 \\
State head (privileged labels) & 0 & 0 & 0 & 67 & 0 & 0 & 11.1 & 100 & 58 \\
Image-only reach-then-change & 25 & 0 & 8 & 0 & 0 & 0 & 5.6 & 83 & 67 \\
\midrule
Cost trained on planner states, one round & 25 & 58 & \textbf{83} & \textbf{92} & 17 & 58 & 55.6 & 100 & 92 \\
\rowcolor{pactrow}\textbf{Cost trained on planner states, two rounds} & \textbf{83} & \textbf{83} & 25 & \textbf{92} & \textbf{33} & \textbf{92} & \textbf{68.1} & 100 & 83 \\
\midrule
 
\bottomrule
\end{tabular}}
\end{table}

\paragraph{Setup and exchange rate across five encoders.}
To isolate focus failure cleanly on manipulation, we evaluate the released JEPA-WM Meta-World model of~\cite{terver2025drives}, which encodes images into 256 patch tokens using frozen DINOv2 ViT-S/14. Its CEM planner optimizes 6 action chunks with 300 samples, 10 elites, and 15 iterations across 100-step episodes. To remove dynamics prediction error completely, every candidate sequence is rolled out inside a copy of the true simulator and only its final frame is encoded and scored. We evaluate six object tasks and two controls over 12 identical test episodes per task.

We next measure how strongly an image distance favors the robot arm over the object. Using inverse kinematics, we place the arm and the object independently across 30 configurations in each of eleven Meta-World tasks. We then compute the \emph{exchange rate}, defined as the feature distance caused by shifting only the hand by 5\,cm divided by the distance caused by leaving only the object off target by its success threshold. Any value above 1 means the visual cost prefers fixing a small hand error over finishing the task.

As shown in \cref{fig:exchange}, every image distance sits far above 1 across all \Eb{ntask} object tasks. Shifting the hand by 5\,cm costs \Eb{pixel-lo} to \Eb{pixel-hi}$\times$ more than missing the object in raw pixels, and roughly 2 to 14$\times$ more across DINOv2, SigLIP, and V-JEPA~2, with CLIP neutral at \Eb{clip-lo} on \task{plate-slide} and \Eb{clip-lo2} to \Eb{clip-hi}$\times$ on the other eight tasks. Inspecting real start and goal frames explains why. Robot patches fill roughly half the image yet carry 72 to 79\% of the squared feature distance, even though a linear readout on those exact same frozen DINOv2 features locates the object within 0.14 to 0.77 of the success threshold.

\paragraph{The same bias stalls pushing.}
Push-T displays the exact same imbalance in a 2D workspace. Even though overall local ranking looks healthy at $\rho(1){=}\Dg{pusht-r1}$ and $\rho(6){=}\Dg{pusht-r6}$, the latent cost orders the T-block's true object progress correctly only \Dg{pt-obj-lat} of the time because the moving agent circle dominates image changes. By contrast, graph distance along recorded routes orders true object progress correctly \Dg{pt-obj-graph} of the time, as plotted in \cref{fig:theory}c. Without keeping routes on continuous recorded pushes, merely placing an anchor three moves ahead leaves Push-T success stuck at \Fx{pusht-k3}\% in \cref{tab:contactcfg}.

\paragraph{Meta-World remedy: learned costs vs.\ near recorded targets.}
\Cref{tab:mw} compares two possible fixes under exact dynamics. Because the focus failure lives in the cost function, we first test alternative costs built on the same frozen DINOv2 features. Both the released visual distance and TEMPO~\cite{moukheiber2026long}, a learned temporal-distance cost that remains dominated by arm travel time, solve 0\% of the six object tasks. A learned progress head trained on planner-visited states, which scores hand-to-object distance before contact and object-to-goal distance after contact, lifts mean success to 68\% while keeping both controls intact. However, once exact simulator rollouts are replaced with the learned predictor, success falls back near zero because predicted latent futures fail to resolve fine contact.

Second, we leave the released cost, the CEM search, and exact dynamics completely unchanged, and replace only the final goal image with the encoded frame 20 steps ahead on the test episode's own demonstration. That single change solves \Q{mw-demo-solved} of \Q{mw-demo-n} object episodes, averaging \Q{mw-demo-mean}\% with 50\% on \task{push}. Because this check reads from the test demonstration, it acts as a privileged upper bound rather than a deployable method. Even so, it establishes a unifying principle. \emph{Even the unchanged whole-image distance succeeds at manipulation when its target is nearby and lies on a recorded route}, because matching the arm pose along a recorded trajectory automatically moves the object as recorded. Graph Anchors turns this principle into a deployable, training-free planner (\cref{sec:method} and \cref{app:method}).

% ------------------------------------------------------------------------------
\subsection{Formal Analysis and Closed-Loop Drift}
\label{app:theory}\label{app:predicts}

We now formalize why aiming at a nearby graph anchor moves the agent closer to a distant goal, how approximation error in graph distance affects that guarantee, and why losing directional pull forces slow quadratic hitting times. Let $d(\cdot,\cdot) \in [0,\infty]$ be the directed shortest-path quasimetric on the graph $G$ with edge weights $w(i,j)\ge 1$. Even when temporal edges are directed and cross-episode edges carry a higher weight $\lambda > 1$, the directed triangle inequality $d(y,g) \le d(y,a)+d(a,g)$ still holds. Write $x_t$ for the state at decision $t$ and $h_t = d(x_t, g) \ge 0$ for the remaining graph distance to $g$. In plain terms, \cref{lem:transfer} says that moving closer to any stepping stone along a shortest path guarantees at least as much progress toward the final goal.

\begin{lemmabox}
\begin{lemma}[Progress transfer]\label{lem:transfer}
Suppose $d(x,g) < \infty$ and let $a$ lie on a shortest directed path from $x$ to $g$. Then, for every state $y$,
\[
d(x,g) - d(y,g) \;\ge\; d(x,a) - d(y,a).
\]
In particular, $d(y,a) < d(x,a) \implies d(y,g) < d(x,g)$.
\end{lemma}
\end{lemmabox}
\noindent\emph{Proof.} Since $a$ lies on a shortest path from $x$ to $g$, $d(x,g) = d(x,a) + d(a,g)$. By the directed triangle inequality, $d(y,g) \le d(y,a) + d(a,g)$. Subtracting the two gives $d(x,g) - d(y,g) \ge d(x,a) - d(y,a)$.\hfill$\square$

Because our latent graph distance $d$ approximates the true physical route distance $d^*$, \cref{cor:true} checks how much distortion a single planning step can absorb while still guaranteeing true physical progress.

\begin{corbox}
\begin{corollary}[Transfer under metric distortion]\label{cor:true}
Let $d^*$ be a true distance and suppose there exists $\epsilon \in [0, 1)$ such that
\[
(1-\epsilon)\,d^*(u,g) \;\le\; d(u,g) \;\le\; (1+\epsilon)\,d^*(u,g)
\]
for $u \in \{x, y\}$. Let $m = d(x,a) - d(y,a)$ be the progress toward the anchor in \cref{lem:transfer}. If $m > 2\epsilon\, d^*(x,g)$, then $d^*(y,g) < d^*(x,g)$.
\end{corollary}
\end{corbox}
\noindent\emph{Proof.} By \cref{lem:transfer}, $d(y,g) \le d(x,g) - m \le (1+\epsilon)\,d^*(x,g) - m$. Since $1-\epsilon > 0$, dividing by $1-\epsilon$ yields
\[
d^*(y,g) \;\le\; \frac{d(y,g)}{1-\epsilon} \;\le\; \frac{(1+\epsilon)\,d^*(x,g) - m}{1-\epsilon},
\]
which is strictly less than $d^*(x,g)$ if and only if $m > 2\epsilon\,d^*(x,g)$.\hfill$\square$

\noindent\emph{Empirical stretch and multi-step behavior.} On PointMaze-Medium, using hops on a true-position reference graph as $d^*$, the ratio $d/d^*$ stays tightly concentrated near 1. From the 5th to the 95th percentile, it lies between \Th{stretchmidlo} and \Th{stretchmidhi} at 10--19 hops and between \Th{stretchfarlo} and \Th{stretchfarhi} beyond 30 hops. That corresponds to roughly $\epsilon \approx 0.03$, so even 34 hops from the goal a single-step margin $m > 2\epsilon d^* \approx 2$ hops suffices for most pairs.

More importantly, that linear-in-$d^*$ margin is only a worst-case single-step artifact caused by distortion pushing in opposite directions at $x$ and $y$. Distortion does \emph{not} compound over an episode. Whenever $(1-\epsilon)d^* \le d \le (1+\epsilon)d^*$ holds across $T$ decisions, we have $h_0 = d(x_0,g) \le (1+\epsilon)\,d^*(x_0,g)$ and $d^*(x_t,g) \le h_t/(1-\epsilon)$ at every step. Any sustained negative drift $-\mu$ in graph distance $d$ from \cref{prop:drift}b therefore drives $d^*(x_t,g)$ to zero with no $2\epsilon D$ step barrier, inflating the total horizon by at most the constant factor $1+\epsilon$.

\Cref{prop:drift} connects per-decision progress to total episode length. Intuitively, an agent without downhill pull wanders like a random walk and needs quadratically many decisions to reach a distant goal, whereas even a modest steady drift toward the goal cuts the reaching time down to linear.

\begin{propbox}
\begin{proposition}[No drift vs.\ negative drift]\label{prop:drift}
Let $(h_t)_{t=0}^T$ be a non-negative process ($h_t \ge 0$) with $h_0 = D$ and bounded increments $|h_{t+1} - h_t| \le c$ for all $t < T$.

\smallskip\noindent(a) \emph{No drift.} Let $0 \le r < D$. If $\mathbb{E}[h_{t+1}-h_t \mid x_{0:t}] \ge 0$ whenever $h_t > r$, then
\[
\Pr\!\Big(\min_{0 \le t \le T} h_t \le r\Big) \;\le\; \exp\!\Big(-\frac{(D-r)^2}{2Tc^2}\Big).
\]
(b) \emph{Drift.} If $\mu > 0$, $\mu T > D$, and $\mathbb{E}[h_{t+1}-h_t \mid x_{0:t}] \le -\mu$ whenever $h_t > 0$, then
\[
\Pr\!\Big(\min_{0 \le t \le T} h_t > 0\Big) \;\le\; \exp\!\Big(-\frac{(\mu T - D)^2}{2Tc^2}\Big).
\]
\end{proposition}
\end{propbox}
\noindent\emph{Proof.} For part~(a), define the stopping time $\tau = \inf\{t \ge 0 : h_t \le r\}$ and the stopped process $M_t = D - h_{t \wedge \tau}$. Then $M_0 = 0$, the conditional increment satisfies $\mathbb{E}[M_{t+1}-M_t \mid x_{0:t}] \le 0$ for every $t$ and equals zero after $\tau$, and $|M_{t+1}-M_t| \le c$ gives a predictable interval of width $2c$. Because the event $\{\min_{t\le T} h_t \le r\}$ coincides with $\{M_T \ge D-r\}$, applying Azuma--Hoeffding at time $T$ yields the bound. For part~(b), let $\tau = \inf\{t \ge 0 : h_t \le 0\}$ and $Y_t = h_{t \wedge \tau} + \mu (t \wedge \tau)$. Conditioned on $x_{0:t}$, the increment $Y_{t+1}-Y_t$ has non-positive expectation and lies inside $[-c+\mu, c+\mu]$ of width $2c$ before $\tau$ and at $0$ after $\tau$, with $\mu \le c$ following automatically from $|h_{t+1}-h_t|\le c$. On the failure event $\{\min_{t\le T} h_t > 0\}$ we have $\tau > T$ and $h_T > 0$, so $Y_T - Y_0 = h_T + \mu T - D > \mu T - D > 0$. Applying Azuma--Hoeffding at time $T$ completes the proof.\hfill$\square$

\noindent\emph{Scaling intuition.} To picture what \cref{prop:drift} means in closed loop, suppose each five-block decision travels $s$ hops and ends closer to the goal with probability $p$. Here $p$ measures closed-loop decision accuracy, which is separate from the offline one-hop score $\rho(h)$. The expected hop change per decision is $\mathbb{E}[h_{t+1}-h_t] = (1-2p)\,s$. When the planner merely moves at chance with $p=\tfrac12$, case~(a) applies and reaching a goal at distance $D$ requires $\Omega(D^2)$ decisions. Entering the linear $O(D)$ regime of case~(b) requires $p > \tfrac12 + \frac{D}{2sT}$, which equals roughly \Th{need} on PointMaze-Medium at $d{=}200$ where $D{=}34$, $T{=}16$, and $s{=}5$. These bounds explain the shift from quadratic wandering to linear progress rather than yielding tight constants, and they do not model contact execution such as initiating a grasp on Cube.

\begin{figure}[t]
\centering
\includegraphics[width=\textwidth]{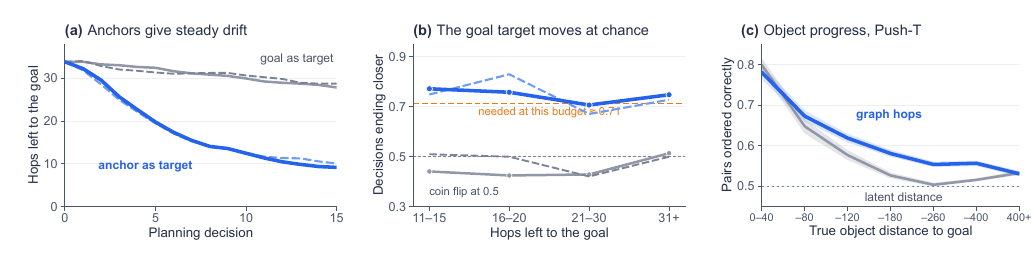}
\caption{\textbf{Closed-loop drift and object progress.} Panel~(a) plots mean remaining true-position graph hops per decision on PointMaze-Medium at $d{=}200$, where finished episodes retain their final value and dashed curves show the alternate search horizon per target. Panel~(b) tracks the share of decisions ending closer to the goal against remaining distance. Panel~(c) evaluates Push-T, showing how often state pairs are ordered consistently with true object distance defined as block position plus 100\,px/rad, binned by object distance with 95\% intervals over \Th{goals} goals and ties scored as $\tfrac12$.}
\label{fig:theory}
\end{figure}

\paragraph{Verification on real rollouts.}
\Cref{fig:theory}a and \cref{fig:theory}b check these conditions directly on traced PointMaze-Medium rollouts at $d{=}200$ across three seeds of 50 pairs each, measured in true-position reference hops. Whenever the agent sits more than 10 hops from the goal, three sharp differences stand out:
\begin{itemize}[leftmargin=1.4em,itemsep=2pt,topsep=2pt]
\item \emph{Closer decisions.} Goal-aimed LeWM planners end only \Th{plainclose} of decisions closer to the goal, whereas Graph Anchors end \Th{anchorclose} closer.
\item \emph{Mean drift.} Graph Anchors hold a steady negative drift between $-2.2$ and $-3.7$ hops per decision across all distance bins. LeWM sits at zero drift through the maze interior, averaging \Th{plainmid} hops at 21--30 hops and $+0.19$ at 16--20 hops. Its slight $-$\Th{plainfar} mean beyond 30 hops comes from bouncing off outer maze corners rather than directed navigation.
\item \emph{Stalls and recovery.} Immediately after a non-improving decision, Graph Anchors still drift at $-$\Th{afterstallanchor} hops while LeWM stays flat at \Th{afterstallplain}. Overall, anchors stall far less often, recording \Th{stallsanchor} stalls compared to \Th{stallsplain} for LeWM, and recover far more frequently at \Th{recanchor}\% versus \Th{recplain}\%.
\end{itemize}
\Cref{tab:driftt8} repeats this drift measurement on Two-Room at $d{=}75$ across three seeds and on Push-T at $d{=}150$ across two seeds, building reference graphs over true agent and block states. Once again, goal-aimed planning ends closer only \Th{trplain} of the time on Two-Room and \Th{ptplain} on Push-T, showing almost zero net drift. Graph Anchors raise those shares to \Th{tranchor} and \Th{ptanchor} with strong negative drift. On Push-T, measuring graph-free block-pose error confirms the exact same ordering. The share of decisions that reduce true block error is 28\% for LeWM and 31\% for LeWM with min-over-plan, jumping to 46\% for Graph Anchors and 47\% with $\lambda{=}3$. Across episodes, the corresponding 95\% confidence intervals are 24--31\%, 28--35\%, 41--50\%, and 43--51\%.

\begin{table}[t]
\centering
\small
\renewcommand{\arraystretch}{1.08}
\setlength{\tabcolsep}{7pt}
\caption{\textbf{Closed-loop drift on Two-Room and Push-T.} Each decision runs for 25 steps. Closer reports the percentage of decisions ending closer to the goal in true-state reference hops with 95\% confidence intervals over episodes, and drift reports the mean hop change per decision.}
\label{tab:driftt8}
\begin{tabularx}{0.82\textwidth}{@{}Xcccc@{}}
\toprule
Planner & Succ. (\%) & Decisions & Closer (\%) & Drift \\
\midrule
\multicolumn{5}{@{}l}{\emph{Two-Room}, $d{=}75$}\\
LeWM & 32 & 705 & 33 [29, 37] & $-$0.08 \\
LeWM + min-over-plan & 61 & 533 & 46 [41, 51] & $-$0.60 \\
\rowcolor{oursrow}\textbf{Graph Anchors} & \textbf{94} & 474 & \textbf{74} [71, 78] & \textbf{$-$1.54} \\
\midrule
\multicolumn{5}{@{}l}{\emph{Push-T}, $d{=}150$}\\
LeWM & 4 & 1106 & 31 [28, 34] & $-$0.03 \\
LeWM + min-over-plan & 6 & 1050 & 37 [34, 40] & $-$0.14 \\
\rowcolor{oursrow}Graph Anchors & 31 & 979 & 53 [50, 56] & $-$0.67 \\
\rowcolor{oursrow}\textbf{Graph Anchors ($\lambda{=}3$)} & \textbf{46} & 897 & \textbf{54} [51, 58] & \textbf{$-$0.77} \\
\bottomrule
\end{tabularx}
\end{table}

It is worth noting that \cref{lem:transfer} is a sufficient condition rather than a literal picture of every decision. In practice, an executed move often stops partway before reaching its active anchor. Measured in straight-line node steps of one median five-step move, PointMaze-Medium anchors begin \Th{pmlinestart} node steps away and come within \Th{pmlineclosest} at closest approach. Across decisions, the agent reaches within 1 and 2 node steps on \Th{pmlineone}\% and \Th{pmlinetwo}\% of moves on PointMaze-Medium, \Th{trlineone}\% and \Th{trlinetwo}\% on Two-Room, and \Th{ptlineone}\% and \Th{ptlinetwo}\% on Push-T. Reference-graph hops make this approach look slightly smaller because fine-grained reference edges are shorter than a full recorded move. For example, on Two-Room the median closest approach is \Th{trlineclosest} node steps compared to \Th{trhopclosest} reference hops, and on PointMaze-Medium the agent ends closer to its anchor on \Th{endline}\% of decisions in straight-line distance versus \Th{endnear}\% in reference hops. What \cref{prop:drift} actually requires, however, is steady net drift toward the final goal, which \cref{fig:theory}a, \cref{fig:theory}b, and \cref{tab:driftt8} verify directly.

\paragraph{Object progress and failure prediction.}
Even though graph nodes store whole-frame latents with no object factorization, the latent space still represents the manipulated object cleanly at short range. Linear probes on held-out Push-T episodes recover block position with $R^2{=}\Th{r2block}$ and block angle with $R^2{=}\Th{r2angle}$. Likewise, cross-episode neighbour edges shift the block by only \Th{nnblock}\,px at the median, compared to \Th{randblock}\,px for random frame pairs. Over \Th{goals} held-out Push-T goals shown in \cref{fig:theory}c, shortest-path graph hops order state pairs by true object distance accurately \Th{pairgraph} of the time, compared to \Th{pairlatent} for direct latent distance. Graph hops stay ahead across every distance bin, scoring \Th{neargraph} near the goal, \Th{midgraph} at mid-range, and \Th{fargraph} far away, against \Th{nearlat}, \Th{midlat}, and \Th{farlat} for latent distance.

Finally, because the graph-distance test runs completely offline before planning begins, it predicts closed-loop difficulty across environments in advance. As shown in \cref{fig:predict} and \cref{tab:where}, goal graph distance correlates strongly with LeWM success at Spearman \Pf{pred-lewm} across \Pf{pred-n} settings, and it also predicts where anchors gain the most over min-over-plan at Spearman \Pf{pred-gain} across \Pf{pred-ngain} settings.

\FloatBarrier
% ==============================================================================
\section{Graph Anchors: Algorithm and Implementation}
\label{app:remedy}\label{app:method}

Graph Anchors turns the diagnosis of \cref{sec:problem} and \cref{app:diagnosis} into a simple planning method without changing or retraining the world model. As detailed in \cref{alg:sga}, the method builds a sparse latent graph once from training trajectories, selects an anchor a few moves ahead along the shortest path to the goal at each replanning step, and scores candidate CEM rollouts by their closest approach to that nearby anchor.

% ------------------------------------------------------------------------------
\subsection{Algorithm and Walkthrough}
\label{app:walkthrough}

\begin{algorithm}[t]
\caption{Graph Anchors (complete specification). Part 1 runs once per environment; Parts 2--3 run at every replanning step.}
\label{alg:sga}\label{alg:graph}\label{alg:anchor}\label{alg:plan}
\small
\begin{algorithmic}[1]
\Require training episodes $\mathcal{D}$, frozen encoder $f$ and predictor $p$, goal latent $z_g$; $k{=}10$ neighbours, span $K$ (nav: 6; contact: 2), switch cost $\lambda$ (nav: 1; contact: 3), horizon $H$ (nav: 10; contact: 2)
\Statex \textbf{Part 1: Build sparse graph (once)}
\State $\mathcal{V}\gets$ every fifth frame of every episode; $z_i\gets f(o_i)$ \Comment{nodes}
\State $\mathcal{E}_{\text{T}}\gets\{(i,j): e(j){=}e(i),\ t(j){=}t(i){+}5\}$ \Comment{temporal}
\State $\tau\gets\operatorname{median}_{(i,j)\in\mathcal{E}_{\text{T}}}\Vert z_i-z_j\Vert^2$ \Comment{1-block median}
\State $\mathcal{E}_{\text{N}}\gets\{(i,j),(j,i): j\in\mathrm{NN}_k(i),\ \Vert z_i-z_j\Vert^2<\tau\}$ \Comment{k-NN}
\State remove test episodes; $\mathcal{G}\gets(\mathcal{V},\mathcal{E}_{\text{T}}\cup\mathcal{E}_{\text{N}})$
\Statex \textbf{Part 2: Select anchor (current latent $z_t$)}
\State $s\gets\arg\min_i\Vert z_i-z_t\Vert$;\ $g\gets\arg\min_i\Vert z_i-z_g\Vert$
\State Dijkstra from $g$ on reversed $\mathcal{G}$, cached per goal (temporal cost 1, neighbour cost $\lambda$)
\If{$g$ is reachable from $s$ and $>K$ moves away}
  \State $z_a\gets$ node $K$ moves along $\pi(s,g)$ \Comment{$\lambda{>}1$: temporal only}
\Else
  \State $z_a\gets z_g$
\EndIf
\Statex \textbf{Part 3: Plan via unchanged CEM}
\For{30 iterations}
  \State sample 300 sequences of $H$ action blocks; roll out $\hat z^{(n)}_{1:H}$ with $p$
  \State score $d_a^{(n)}=\min_h\Vert\hat z^{(n)}_h-z_a\Vert^2$ \Comment{closest approach, \cref{eq:cost}}
  \State refit Gaussian to the 30 lowest-scoring sequences
\EndFor
\State execute best plan (nav: 5 blocks; contact: to closest approach), replan
\end{algorithmic}
\end{algorithm}

\begin{figure}[t]
\centering
\includegraphics[width=\textwidth]{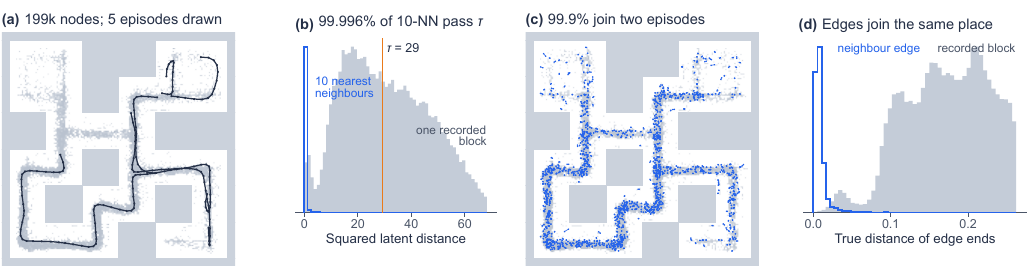}
\caption{\textbf{Building the latent graph} on real PointMaze-Medium data, where grey dots mark node positions. Panel~(a) encodes every fifth training frame as a node and links consecutive frames within an episode by directed temporal edges. Panel~(b) sets the threshold $\tau$ to the median squared latent distance of one recorded block in grey, so each node's 10 nearest latent neighbours in blue fall far below $\tau$. Panel~(c) draws 1{,}500 neighbour edges at their true positions, showing that 99.9\% bridge distinct episodes and enable route stitching. Panel~(d) confirms that neighbour edges connect states physically much closer than one five-step move.}
\label{fig:suppgraph}
\end{figure}

\begin{figure}[t]
\centering
\includegraphics[width=\textwidth]{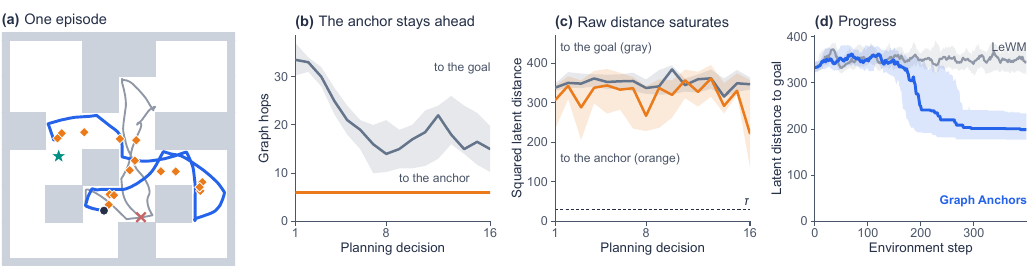}
\caption{\textbf{Anchors in closed loop} on PointMaze-Medium at $d{=}200$ over 50 identical start and goal pairs, showing medians and interquartile ranges. Panel~(a) plots the longest-path episode solved by Graph Anchors in blue with anchor diamonds and missed by LeWM in grey with its final cross. Panel~(b) compares remaining graph hops to the goal in grey against hops to the active anchor in orange, illustrating that every planning step stays a 6-hop problem. Panel~(c) shows that raw squared latent distance flattens out beyond a few hops, confirming that local \emph{ranking} from \cref{fig:diagnosis}a and \cref{tab:truediag}, rather than raw magnitude, drives progress. Panel~(d) shows latent distance to the goal falling steadily with anchors while stagnating without them.}
\label{fig:suppanchor}
\end{figure}

\begin{figure}[t]
\centering
\includegraphics[width=\textwidth]{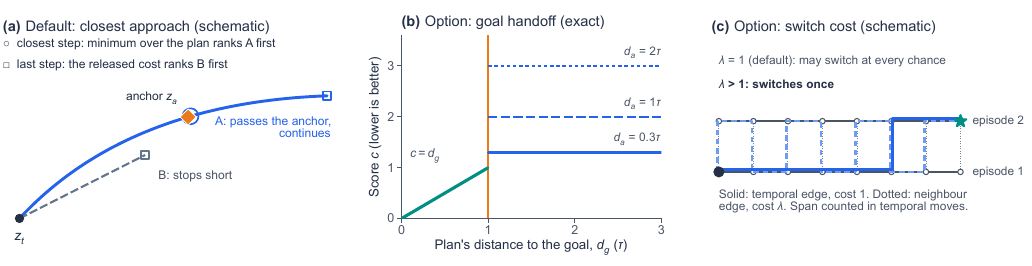}
\caption{\textbf{Scoring plans and preserving trajectory continuity.} Panel~(a) illustrates closest-approach scoring from \cref{eq:cost}, which favors a rollout passing through the anchor toward the goal in A over one stopping at the anchor in B. Panel~(b) shows optional gated goal handoff from \cref{app:handoff}, where any plan coming within $\tau$ of $z_g$ once $d_G(s,g)\le 12$ is scored by $d_g$ and beats every anchor-only plan scored by $d_a+\tau$. Panel~(c) shows how charging an episode-switch cost $\lambda>1$ on neighbour edges keeps shortest paths on continuous recorded transitions.}
\label{fig:suppcost}
\end{figure}

\paragraph{Building the latent graph.}
We encode every fifth frame, corresponding to one action block, of every training episode with the frozen encoder and link consecutive frames by directed temporal edges as shown in \cref{fig:suppgraph}a. To stitch routes across different trajectories, we add bidirectional neighbour edges from each node to its $k{=}10$ nearest latent neighbours whose squared latent distance falls below $\tau$. As illustrated in \cref{fig:suppgraph}b, $\tau$ is set automatically to the empirical median squared latent length of a single recorded action block. Because $\tau$ comes directly from the model's own typical one-step jump size, no environment-specific distance threshold needs to be tuned. On PointMaze-Medium this builds 199k nodes, 194k temporal edges, and 2.0\,M neighbour edges. \Cref{fig:suppgraph}c and \cref{fig:suppgraph}d show that 99.9\% of neighbour edges connect distinct episodes at physical separations far smaller than one five-step move. Test episodes are always removed before evaluation so no anchor can ever replay a test trajectory.

\paragraph{Selecting and following anchors.}
At each replanning step, the current latent $z_t$ and goal latent $z_g$ map to their nearest graph nodes $s$ and $g$ via a single GPU matrix product. One reversed Dijkstra search from $g$ is run once and cached for the entire episode. Each replanning decision therefore only walks $K$ successors forward along the shortest path $\pi(s,g)$, falling back directly to $z_g$ once the goal is within $K$ moves or unreachable, as written in Part~2 of \cref{alg:sga}. \Cref{fig:suppanchor} traces this closed loop on real PointMaze-Medium episodes at $d{=}200$ on seed 0. Anchors stay six hops ahead around walls, turning a difficult 34-hop navigation problem into a sequence of reliable 6-hop subproblems.

\paragraph{Closest-approach scoring and goal handoff.}
CEM keeps LeWM's exact optimizer with 300 samples, 30 elites, and 30 iterations, changing only the target latent, the scored step, and the horizon. Rather than scoring only the final step of a rollout, \cref{fig:suppcost}a scores each rollout by its minimum distance to $z_a$ across steps $h \in \{2,\dots,H{+}1\}$. Intuitively, this encourages the agent to pass smoothly \emph{through} the stepping stone instead of braking to a stop at it. In navigation we plan over $H{=}10$ blocks and execute 5 blocks before replanning. In the contact setting we use tighter anchors $K{=}2$, switch cost $\lambda{=}3$, and horizon $H{=}2$, executing up to the block of closest approach. \Cref{fig:suppcost}b also illustrates an optional gated goal handoff evaluated in \cref{app:handoff}. Whenever $d_G(s,g)\le 12$, any plan entering within $\tau$ of $z_g$ is scored directly by $d_g$, while all other plans are scored by $d_a+\tau$.

% ------------------------------------------------------------------------------
\subsection{Implementation Details and Graph Quality}
\label{app:impl}

\begin{table}[t]
\centering
\small
\renewcommand{\arraystretch}{1.08}
\setlength{\tabcolsep}{7pt}
\caption{\textbf{Latent graphs across environments.} Node counts, temporal edges, neighbour edges, the automatic 1-block median threshold $\tau$, and fp16 node storage in MB for 192-D latents.}
\label{tab:graphs}
\begin{tabularx}{0.84\textwidth}{@{}Xccccc@{}}
\toprule
Environment & Nodes & Temporal & Neighbour & $\tau$ & MB \\
\midrule
Push-T & 460k & 0.44M & 6.21M & 25.1 & 169 \\
Two-Room & 181k & 0.17M & 2.10M & 295.7 & 66 \\
PointMaze-M & 199k & 0.19M & 2.32M & 29.2 & 73 \\
OGBench-Cube & 400k & 0.39M & 4.81M & 26.9 & 146 \\
Reacher & 400k & 0.39M & 4.57M & 76.2 & 146 \\
 
\bottomrule
\end{tabularx}
\end{table}

\paragraph{Evaluation setup and checkpoints.}
All evaluations use the released \texttt{stable-worldmodel} framework~\cite{maes2026leworldmodel} with $224{\times}224$ images, 5-step action blocks, a step budget of $2d$, and each environment's native success criteria. Every random seed fixes the exact same 50 start and goal pairs and identical CEM noise across methods. Push-T, Two-Room, Cube, and Reacher use the released LeWM checkpoints and datasets, while PointMaze-Medium uses the public D4RL LeWM checkpoint of~\cite{jang2026latent} verified for safe class loading. Because PointMaze-Large has no released checkpoint, we train LeWM with its default script on OGBench \texttt{pointmaze-large-navigate} for 10 epochs of 3{,}000 batches on one GPU over 1{,}000 episodes of 1{,}000 steps. As summarized in \cref{tab:graphs}, this yields \Q{pml-nodes-k}\,k nodes, \Q{pml-tedges-k}\,k temporal edges, and \Q{pml-nedges-m}\,M neighbour edges. For the unseen-dataset graph in \cref{fig:robustbars}, we render OGBench \texttt{pointmaze-large-stitch} with the exact same environment class, verify pixel and encoding agreement against stored training frames, and encode those frames with the frozen model.

\paragraph{Hyperparameters, timing, and offline diagnostics.}
All environments share identical graph construction rules with $k{=}10$ neighbours, 5-step nodes, and the empirical 1-block median threshold $\tau$. On PointMaze-Medium, building the graph takes \Q{time-build-s}\,s once, and one CPU Dijkstra search takes 84 to 93\,ms across three timed runs. In interleaved timed runs at $d{=}100$ across 10 pairs and 3 repeats, one planning decision averages 0.8 to 1.5\,s for LeWM, 2.6\,s for LeWM with min-over-plan, and 1.8\,s for Graph Anchors. Because solved episodes finish early, wall-clock episode time mixes solve rate and compute cost, so exact per-step GPU overhead is isolated separately in \cref{tab:comp_mech}b. For offline diagnostics, \cref{fig:diagnosis}a and \cref{fig:hops} rebuild identical CPU graphs from stored node latents, testing 300 goals with up to 4 states per hop up to $h{=}40$ and 200 pairs per distance in \cref{fig:hops}.

\paragraph{Edge fidelity and Push-T route fragmentation.}
To verify that neighbour edges do not create illegal shortcuts across walls or teleports, we count how many neighbour edges span a larger physical distance than the 95th percentile of a real five-step move. These false-shortcut rates are tiny on every benchmark, measuring \Q{td-pmm-worm}\% on PointMaze-Medium, \Q{td-tworoom-worm}\% on Two-Room, and \Q{td-pusht-worm}\% on Push-T, while more than 94\% of neighbour edges bridge distinct episodes. Checking all 994{,}834 unique neighbour edges on PointMaze-Medium finds zero wall crossings, as reported in \cref{tab:comp_mech}b.

On Push-T, however, accurate local edges alone are not enough. When neighbour edges carry the same weight $\lambda{=}1$ as temporal edges, unweighted shortest paths jump repeatedly between different recorded episodes. Across full $d{=}150$ routes in \cref{tab:comp_mech}a, 53.4\% of edges are cross-episode switches rather than recorded temporal transitions, and that share rises to 64\% inside the first six hops where anchors are selected, averaging 8.2 switches per 6-hop anchor over 400 paths. Jumping between demos mid-push breaks continuous block contact. Raising the neighbour edge weight to $\lambda{=}3$, illustrated in \cref{fig:suppcost}c, cuts switches per anchor to 4.6, and down to 3.1 at $\lambda{=}10$, while keeping every node reachable. This observation directly motivates the contact setting evaluated in \cref{app:handoff}.

\FloatBarrier
% ==============================================================================
\section{Extended Results and Ablations}
\label{app:results}

% ------------------------------------------------------------------------------
\subsection{All Distances, Retraining Sensitivity, and Paired Tests}
\label{app:seeds}\label{app:retrain}\label{app:stats}

\Cref{tab:compare} reports every environment and goal distance for all world-model planners and offline policies, with every individual seed run recorded in \cref{tab:seedsx}.

\begin{table}[t]
\centering
\caption{\textbf{Success (\%) at every goal distance} for world-model planners and offline policies on identical pairs and $2d$ budgets, with stitched budgets of 200 on Two-Room and 400 on the mazes. Entries report mean $\pm$ std over 50-pair evaluations, where $^\dagger$ marks a single evaluation and bold marks the best score per column. Exact paired McNemar tests appear in \cref{app:stats} and the full per-evaluation results appear in \cref{tab:seedsx}.}
\label{tab:compare}
\footnotesize
\renewcommand{\arraystretch}{1.06}
\setlength{\tabcolsep}{5pt}

\begin{tabularx}{0.92\textwidth}{@{}p{4.1cm}YYYYY@{}}
\toprule
\textbf{Push-T} & $d{=}25$ & $d{=}50$ & $d{=}75$ & $d{=}100$ & $d{=}150$ \\
\midrule
LeWM~\cite{maes2026leworldmodel} & 89\pmstd{1} & 41\pmstd{13} & 21\pmstd{4} & 17\pmstd{1} & 5\pmstd{4} \\
LeWM + min-over-plan & \textbf{91}\pmstd{3} & 55\pmstd{10} & 33\pmstd{7} & 17\pmstd{7} & 9\pmstd{1} \\
\midrule
GCIVL~\cite{park2025ogbench} & 2\pmstd{3} & 2\pmstd{0} & 0\pmstd{0} & 0\pmstd{0} & 0\pmstd{0} \\
GCIQL~\cite{park2025ogbench} & 0\pmstd{0} & 0\pmstd{0} & 0\pmstd{0} & 0\pmstd{0} & 0\pmstd{0} \\
HILP~\cite{park2024foundation} & 4\pmstd{3} & 0\pmstd{0} & 0\pmstd{0} & 0\pmstd{0} & 0\pmstd{0} \\
\midrule
\rowcolor{oursrow}\textbf{Graph Anchors (ours)} & 82\pmstd{6} & \textbf{65}\pmstd{1} & \textbf{50}\pmstd{0} & \textbf{40}\pmstd{3} & \textbf{30}\pmstd{3} \\
\bottomrule
\end{tabularx}

\vspace{5pt}
\begin{tabularx}{0.92\textwidth}{@{}p{4.1cm}YYYY@{}}
\toprule
\textbf{Two-Room} & $d{=}25$ & $d{=}50$ & $d{=}75$ & stitched \\
\midrule
LeWM~\cite{maes2026leworldmodel} & 88\pmstd{4} & 54\pmstd{5} & 31\pmstd{3} & 61\pmstd{6} \\
LeWM + min-over-plan & \textbf{99}\pmstd{1} & 81\pmstd{5} & 62\pmstd{10} & 82\pmstd{3} \\
\midrule
GCIVL~\cite{park2025ogbench} & \textbf{99}\pmstd{2} & 90\pmstd{0} & 81\pmstd{2} & 86\pmstd{6} \\
GCIQL~\cite{park2025ogbench} & 61\pmstd{1} & 51\pmstd{11} & 37\pmstd{4} & 46\pmstd{0} \\
HILP~\cite{park2024foundation} & 98\pmstd{0} & \textbf{95}\pmstd{1} & 87\pmstd{3} & 89\pmstd{1} \\
\midrule
\rowcolor{oursrow}\textbf{Graph Anchors (ours)} & 96\pmstd{4} & 93\pmstd{1} & \textbf{93}\pmstd{1} & \textbf{99}\pmstd{1} \\
\bottomrule
\end{tabularx}

\vspace{5pt}
\begin{tabularx}{0.92\textwidth}{@{}p{4.1cm}YYYYY@{}}
\toprule
\textbf{PointMaze-Medium} & $d{=}50$ & $d{=}100$ & $d{=}150$ & $d{=}200$ & stitched \\
\midrule
LeWM~\cite{maes2026leworldmodel} & 50\pmstd{12} & 25\pmstd{4} & 13\pmstd{3} & 13\pmstd{2} & 39\pmstd{10} \\
LeWM + min-over-plan & 70\pmstd{5} & 33\pmstd{6} & 17\pmstd{4} & 12\pmstd{5} & 41\pmstd{4} \\
\midrule
GCIVL~\cite{park2025ogbench} & 30$^\dagger$ & 4$^\dagger$ & 2$^\dagger$ & 0$^\dagger$ & 16\pmstd{0} \\
GCIQL~\cite{park2025ogbench} & 34\pmstd{5} & 4\pmstd{2} & 0\pmstd{0} & 1\pmstd{1} & 12\pmstd{3} \\
HILP~\cite{park2024foundation} & 32\pmstd{4} & 3\pmstd{2} & 1\pmstd{1} & 1\pmstd{1} & 18\pmstd{0} \\
\midrule
\rowcolor{oursrow}\textbf{Graph Anchors (ours)} & \textbf{91}\pmstd{2} & \textbf{89}\pmstd{4} & \textbf{88}\pmstd{6} & \textbf{85}\pmstd{6} & \textbf{95}\pmstd{4} \\
\bottomrule
\end{tabularx}

\vspace{5pt}
\begin{tabularx}{0.92\textwidth}{@{}p{4.1cm}YYYYY@{}}
\toprule
\textbf{PointMaze-Large} & $d{=}50$ & $d{=}100$ & $d{=}200$ & $d{=}400$ & stitched \\
\midrule
LeWM~\cite{maes2026leworldmodel} & 14\pmstd{3} & 15\pmstd{7} & 13\pmstd{4} & 8\pmstd{0} & 5\pmstd{1} \\
LeWM + min-over-plan & 44\pmstd{8} & 28\pmstd{3} & 23\pmstd{4} & 26\pmstd{6} & 17\pmstd{1} \\
\midrule
GCIVL~\cite{park2025ogbench} & 9\pmstd{1} & 8\pmstd{3} & 9\pmstd{1} & 1\pmstd{1} & 3\pmstd{1} \\
GCIQL~\cite{park2025ogbench} & 9\pmstd{1} & 8\pmstd{3} & 9\pmstd{1} & 3\pmstd{1} & 3\pmstd{1} \\
HILP~\cite{park2024foundation} & 12\pmstd{0} & 13\pmstd{4} & 16\pmstd{3} & 9\pmstd{1} & 10\pmstd{6} \\
\midrule
\rowcolor{oursrow}\textbf{Graph Anchors (ours)} & \textbf{62}\pmstd{6} & \textbf{63}\pmstd{7} & \textbf{73}\pmstd{4} & \textbf{87}\pmstd{1} & \textbf{66}\pmstd{6} \\
\bottomrule
\end{tabularx}
\end{table}

\paragraph{Retraining sensitivity on PointMaze-Large.}
Because our original PointMaze-Large LeWM checkpoint from \cref{tab:main} was no longer stored on disk, we trained a second model from scratch using the exact same code, data, and hyperparameters. As shown in \cref{tab:retrain}, absolute success shifts modestly between training runs. For Graph Anchors in the navigation setting, $d{=}400$ success moves from 87\% to 76\% and stitched success moves from 66\% to 45\%. Importantly, the relative separation stays wide, with Graph Anchors remaining far ahead of LeWM, LeWM with min-over-plan, and every training-free alternative from \cref{sec:frozenplanners} at every distance. The PointMaze-Large contact setting in \cref{tab:main} ran on this second checkpoint. As expected on a 400-step maze, taking small 2-move anchors exhausts the step budget before reaching the goal, scoring 13\% at $d{=}400$ and 9\% on stitched pairs.

\begin{table}[t]
\centering
\small
\renewcommand{\arraystretch}{1.08}
\setlength{\tabcolsep}{6pt}
\caption{\textbf{Two training runs of PointMaze-Large LeWM.} Success (\%) averaged over seeds 0 and 42 with 50 pairs per seed, and over seeds 0, 42, and 1 for the contact setting, using identical code, data, and test pairs.}
\label{tab:retrain}
\begin{tabularx}{0.84\textwidth}{@{}Xccccc@{}}
\toprule
Method & $d{=}50$ & 100 & 200 & 400 & stitched \\
\midrule
\multicolumn{6}{@{}l}{\emph{Original model (Table 1)}} \\
\quad LeWM & 14 & 15 & 13 & 8 & 5 \\
\quad LeWM + min-over-plan & 44 & 28 & 23 & 26 & 17 \\
\quad AP (our re-implementation) & 17 & 11 & 16 & 12 & 6 \\
\rowcolor{oursrow}\quad Graph Anchors, navigation setting & 62 & 63 & 73 & 87 & 66 \\
\multicolumn{6}{@{}l}{\emph{Second training run}} \\
\quad LeWM & 14 & 11 & 10 & 4 & 8 \\
\quad LeWM + min-over-plan & 38 & 25 & 24 & 20 & 21 \\
\quad AP (our re-implementation) & 31 & 16 & 17 & 18 & 9 \\
\rowcolor{oursrow}\quad Graph Anchors, navigation setting & 47 & 43 & 58 & 76 & 45 \\
 
\bottomrule
\end{tabularx}
\end{table}

\paragraph{Exact paired statistical tests.}
Because every seed fixes the exact same 50 start and goal pairs across methods, we can test gains cleanly on paired episodes. For each pair of methods, we pool all shared seeds and apply an exact two-sided McNemar test to the episodes where one method succeeds and the other fails. Against both LeWM and LeWM with min-over-plan, every improvement of Graph Anchors at $d{\ge}50$ and on stitched goals is statistically significant at $p<10^{-2}$, and most reach $p<10^{-6}$. The sole non-significant cell is Two-Room at $d{=}25$ versus min-over-plan, where both methods are already near ceiling. Against GCBC, gains at $d{=}200$ and on stitched goals are significant while $d{=}100$ is not. Likewise, building the graph from random play does not differ significantly from the expert graph, and feeding held graph waypoints to GCBC makes it significantly worse at $d{=}200$.

% ------------------------------------------------------------------------------
\subsection{Contact Setting, Switch Cost, and Goal Handoff}
\label{app:handoff}\label{app:contactcfg}

\begin{figure}[t]
\centering
\includegraphics[width=\textwidth]{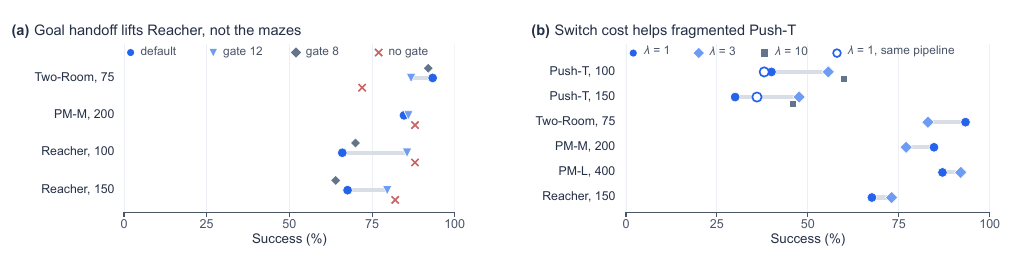}
\caption{\textbf{Options of Graph Anchors} measured by mean success (\%) across seeds. Panel~(a) tests goal handoff from \cref{fig:suppcost}b gated at 8 or 12 graph hops or left ungated. Gated handoff removes anchor detours on short-range Reacher, whereas ungated handoff misfires through the wall on Two-Room. Panel~(b) compares episode-switch cost $\lambda{=}3$ against the default at the longest distance. Open rings show the Push-T default re-run in the switch-cost pipeline at \ptsameA{} and \ptsameB{} for an exact like-for-like check against the older-pipeline default of \R{pusht-ours-100} and \R{pusht-ours-150}.}
\label{fig:suppoptions}
\end{figure}

\begin{figure}[t]
\centering
\includegraphics[width=0.72\textwidth]{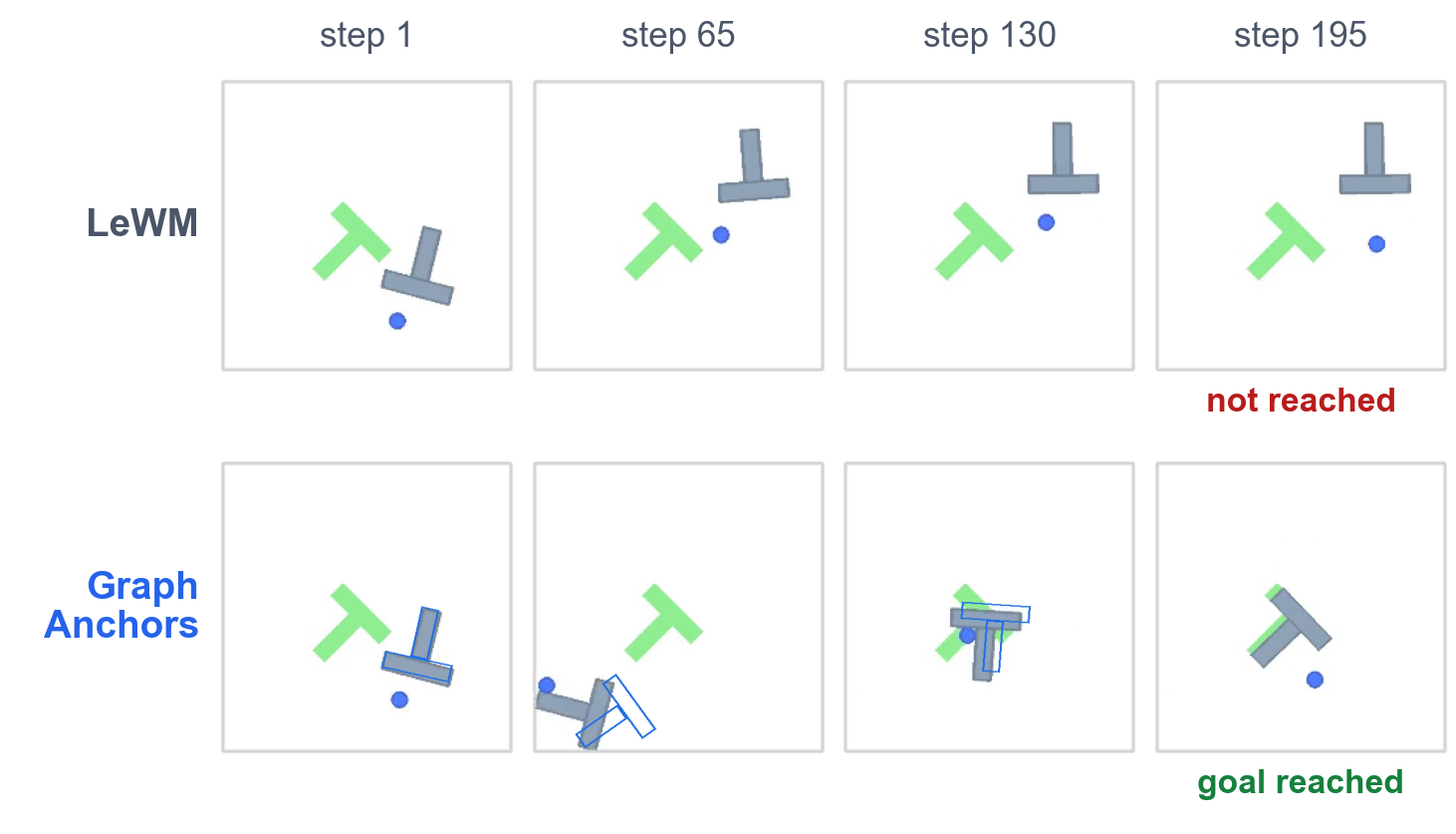}
\caption{\textbf{Push-T in the contact setting} on episode 21 at $d{=}150$ and seed 0 with identical start and goal. The green T marks the goal, and the blue outline marks the active anchor two moves ahead on a recorded push. LeWM leaves the block off target, while Graph Anchors guide the block smoothly onto the goal by step 195 of 300, solving 29 of 50 episodes compared to 1 of 50 for LeWM.}
\label{fig:plannerviewpusht}
\end{figure}

\paragraph{Goal handoff.}
When a task's goals already lie within the reliable range of the visual cost, stopping at intermediate anchors adds unnecessary detours. On Reacher, \cref{fig:hops} showed that graph distances stay at or below 10 hops even at $d{=}150$. Enabling our goal handoff from \cref{fig:suppcost}b, which scores any rollout coming within $\tau$ of $z_g$ directly by $d_g$ and all others by $d_a+\tau$, eliminates those detours. As shown in \cref{fig:suppoptions}a, Reacher success rises from \R{ho-reacher-ours-100}\% to \R{ho12-reacher-100}\% at $d{=}100$ and from \R{ho-reacher-ours-150}\% to \R{ho12-reacher-150}\% at $d{=}150$, outperforming min-over-plan by \R{ho12-gain-reacher-100} and \R{ho12-gain-reacher-150} points across five runs per setting. On Cube at $d{=}150$ across two cluster seeds, handoff changes little, shifting from \R{cube-ho-ours-150}\% to \R{cube-ho-150}\% compared to \R{cube-ho-released-150}\% for LeWM and \R{cube-ho-relmin-150}\% for min-over-plan, because failing to initiate a grasp dominates every planner.

Importantly, handoff must be gated by graph distance so it cannot fire prematurely through walls. On Two-Room at $d{=}75$ on seed 0, leaving handoff ungated causes the visual cost to latch onto goals behind the dividing wall and drops success to \R{ho0-tworoom-75}\%. Gating handoff at 12 graph hops restores Two-Room to \R{ho12-tworoom-75}\% and preserves \R{ho12-pmm-200}\% on PointMaze-Medium at $d{=}200$. Tightening the gate to 8 hops still preserves Two-Room at \R{ho8-tworoom-75}\%, but lowers Reacher to \R{ho8-reacher-100}\% at $d{=}100$ and \R{ho8-reacher-150}\% at $d{=}150$. Because handoff fires on nearly every step once the gate opens, we present it as an optional short-range switch rather than altering the default method.

\paragraph{Episode-switch cost and contact configurations.}
On Push-T, charging $\lambda{=}3$ on cross-episode neighbour edges and counting span strictly in temporal moves keeps routes on continuous recorded pushes, as visualized in \cref{fig:plannerviewpusht}. Even with the navigation horizon $H{=}10$, adding $\lambda{=}3$ lifts $d{=}150$ success from \R{pusht-ours-150}\% to \Q{sw-3-150}\% in \cref{fig:suppoptions}b. Combining $\lambda{=}3$ with tighter anchors $K{=}2$, horizon $H{=}2$, and closest-approach execution in our full contact setting lifts Push-T $d{=}150$ all the way to \Fx{pusht-k2sw}\% in \cref{tab:contactcfg}. Both ingredients matter together. Placing the anchor closer at $K{=}3$ without $\lambda{=}3$ stays stuck at 30\%, whereas adding $\lambda{=}3$ jumps to 72\% across three seeds on the exact same machine and pairs used for AP. On Cube, gains remain modest from \Fx{cube-k6}\% to \Fx{cube-k2sw}\%, staying below AP's \Fx{cube-ap}\%, because Cube episodes require initiating a grasp rather than maintaining a continuous push.

\begin{table}[t]
\centering
\small
\renewcommand{\arraystretch}{1.08}
\setlength{\tabcolsep}{8pt}
\caption{\textbf{Contact-setting configurations at $d{=}150$.} Success (\%) averaged across seeds 0, 42, and 1 on the same machine and identical pairs, where $^\dagger$ marks seed 0 only.}
\label{tab:contactcfg}
\begin{tabularx}{0.72\textwidth}{@{}Xcc@{}}
\toprule
Configuration & Push-T & Cube \\
\midrule
Graph Anchors, $K{=}6$ (default) & 30 & 26 \\
$K{=}3$, closest-approach execution & 30 & 38$^\dagger$ \\
$K{=}3$ + switch cost & \textbf{72} & 26$^\dagger$ \\
\rowcolor{oursrow}$K{=}2$ + switch cost (contact setting) & 71 & \textbf{39} \\
\midrule
AP~\cite{liu2026aim} (same machine) & 65 & 53 \\
 
\bottomrule
\end{tabularx}
\end{table}

% ------------------------------------------------------------------------------
\subsection{Ablations, Scaling, and Robustness Stress Tests}
\label{app:ablations}\label{app:beyond}

\begin{figure}[t]
\centering
\includegraphics[width=0.72\textwidth]{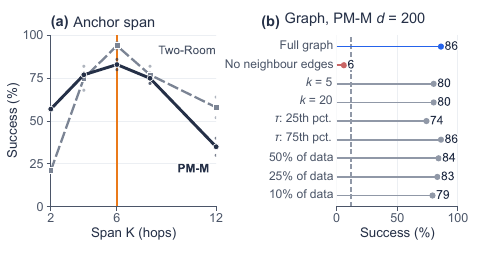}
\caption{\textbf{Anchor span and graph construction ablations.} Panel~(a) sweeps anchor span $K$ across environments, averaged over seeds 0 and 42. Panel~(b) ablates graph construction on PointMaze-Medium at $d{=}200$ on seed 0, with the no-neighbour-edges and data-fraction points averaged over seeds 0 and 42. The dashed line marks LeWM with $3\times$ CEM samples.}
\label{fig:ablate}
\end{figure}

\begin{table}[t]
\centering
\caption{\textbf{Separating target choice from search strength.} Entries report success (\%). Panel~(a) runs a 2$\times$2 factorial crossing target and search rule as in \cref{fig:ablations}a over seeds 0, 42, and 1 on a separate machine. Panel~(b) scales planning horizon $H$ and CEM samples $N$ while aiming at the distant goal, compared against default Graph Anchors in the same runs over seeds 0 and 42.}
\label{tab:factorial}\label{tab:search}
\footnotesize
\begin{minipage}[t]{0.47\textwidth}\centering
\textbf{(a)} Target crossed with search\\[3pt]
\setlength{\tabcolsep}{4pt}
\resizebox{\linewidth}{!}{%
\begin{tabular}{@{}llcc@{}}
\toprule
Target & Search & PM-M, 200 & Two-Room, 75 \\
\midrule
goal & $H{=}5$, last step & 12 & 32 \\
goal & $H{=}10$, min over plan & 9 & 61 \\
anchor & $H{=}5$, last step & 85 & 66 \\
\rowcolor{pactrow}anchor & $H{=}10$, min over plan & 87 & 94 \\
 
\bottomrule
\end{tabular}}
\end{minipage}\hfill
\begin{minipage}[t]{0.50\textwidth}\centering
\textbf{(b)} Stronger search aimed at the goal\\[3pt]
\setlength{\tabcolsep}{4pt}
\resizebox{\linewidth}{!}{%
\begin{tabular}{@{}lcccc@{}}
\toprule
Planner & $H$ & $N$ & PM-M, 200 & Two-Room, 75 \\
\midrule
LeWM + min-over-plan & 10 & 300 & 6 & 58\\
LeWM + min-over-plan & 10 & 1000 & 14 & 56\\
LeWM + min-over-plan & 20 & 300 & 22 & 76\\
LeWM + min-over-plan & 20 & 1000 & 28 & 77\\
\midrule
Graph Anchors & 10 & 300 & \textbf{89} & \textbf{94}\\
 
\bottomrule
\end{tabular}}
\end{minipage}
\end{table}

\begin{table}[t]
\centering
\caption{\textbf{Span--horizon coupling and execution length for Graph Anchors.} Entries report success (\%). Panel~(a) crosses anchor span $K$ with planning horizon $H$ averaged over seeds 0 and 42, with bold marking the best span per row. Panel~(b) varies plan length and execution rule across span $K$ on seed 0, while the default plan-10 execute-5 row appears in panel~(a). The closest-approach rows for $K{\in}\{1,2,6\}$ and the 10-block $K{=}1$ run were evaluated on a second machine.}
\label{tab:spanh}\label{tab:exec}
\footnotesize
\begin{minipage}[t]{0.42\textwidth}\centering
\textbf{(a)} Span $K$ crossed with horizon $H$\\[3pt]
\setlength{\tabcolsep}{4.5pt}
\resizebox{\linewidth}{!}{%
\begin{tabular}{@{}lccccc@{}}
\toprule
& $K{=}2$ & 4 & 6 & 8 & 12 \\
\midrule
\multicolumn{6}{@{}l}{\emph{PointMaze-Medium, $d{=}200$}}\\
$H{=}5$ & 66 & \textbf{92} & 88 & 60 & 11\\
$H{=}10$ & 57 & 77 & \textbf{89} & 75 & 35\\
$H{=}15$ & 45 & 82 & \textbf{87} & 71 & 65\\
\multicolumn{6}{@{}l}{\emph{Two-Room, $d{=}75$}}\\
$H{=}5$ & 24 & \textbf{86} & 80 & 52 & 39\\
$H{=}10$ & 21 & 75 & \textbf{94} & 77 & 58\\
$H{=}15$ & 11 & 78 & \textbf{95} & 85 & 66\\
 
\bottomrule
\end{tabular}}
\end{minipage}\hfill
\begin{minipage}[t]{0.55\textwidth}\centering
\textbf{(b)} Plan length and execution rule\\[3pt]
\setlength{\tabcolsep}{3.5pt}
\resizebox{\linewidth}{!}{%
\begin{tabular}{@{}lccccc@{}}
\toprule
& $K{=}1$ & 2 & 3 & 4 & 6 \\
\midrule
\multicolumn{6}{@{}l}{\emph{PointMaze-Medium, $d{=}200$}}\\
plan $K$, execute $K$ & 48 & 92 & 92 & 88 & 86 \\
plan $K$, execute to the closest approach & 48 & 92 & 92 & 88 & 94 \\
plan 10, execute to the closest approach & 22 & 60 & 72 & 82 & 80 \\
\multicolumn{6}{@{}l}{\emph{Two-Room, $d{=}75$}}\\
plan $K$, execute $K$ & 4 & 76 & 82 & 96 & 86 \\
plan $K$, execute to the closest approach & 4 & 92 & 100 & 100 & 94 \\
plan 10, execute to the closest approach & 10 & 92 & 100 & 100 & 92 \\
 
\bottomrule
\end{tabular}}
\end{minipage}
\end{table}

\paragraph{Target vs.\ search and stronger CEM.}
\Cref{tab:factorial}a separates the benefit of changing the target from the benefit of changing the search rule. On PointMaze-Medium, changing the target accounts for nearly the entire gain. Aiming LeWM's unchanged search at our anchors lifts success from \Q{fa-pmm-released}\% to \Q{fa-pmm-gah5}\%, whereas aiming our closest-approach search at the distant goal stays at \Q{fa-pmm-relmin}\%. On Two-Room, both changes contribute, lifting success from \Q{fa-tworoom-gah5}\% to \Q{fa-tworoom-ours}\%. Scaling up search toward the goal cannot substitute for anchors either. In \cref{tab:search}b, doubling the horizon to $H{=}20$ and raising CEM samples from 300 to 1{,}000 leaves goal-aimed planning at \R{x22-pmm-best}\% on PointMaze-Medium and \R{x22-tworoom-best}\% on Two-Room, far behind default Graph Anchors in the same runs at \R{x22-pmm-ours}\% and \R{x22-tworoom-ours}\%.

\paragraph{Span, horizon, and execution length.}
\Cref{fig:ablate}a and \cref{tab:spanh}a confirm that anchor span $K$ and planning horizon $H$ couple just as the range diagnosis predicts. Longer imagined rollouts can chase farther stepping stones, so the optimal span shifts from $K{=}4$ at $H{=}5$ to $K{=}6$ at $H{=}10$ and $H{=}15$ on both mazes. Across the entire table, our default $K{=}6$ stays within \R{x21-gap6} points of the optimum. Placing anchors far out at $K{=}12$ overwhelms a short 5-block plan on PointMaze-Medium at \R{x21-pmm-k12-h5}\%, but recovers to \R{x21-pmm-k12-h15}\% once the horizon stretches to $H{=}15$. Conversely, overly close anchors at $K{=}2$ stall under fixed 5-block execution, dropping from \R{x21-pmm-k2-h5}\% at $H{=}5$ to \R{x21-pmm-k2-h15}\% at $H{=}15$ on PointMaze-Medium and to just 8\% on Push-T at $d{=}50$.

\Cref{tab:exec}b isolates how execution length solves this short-span stall. Matching plan length to $K$ blocks and stopping execution right at closest approach lets short anchors succeed cleanly, reaching 92 to 94\% on PointMaze-Medium for $K{\in}\{2,3,6\}$ and 100\% on Two-Room for $K{\in}\{3,4\}$. That finding directly motivates the short-horizon closest-approach execution in our contact setting with $K{=}2$ and $H{=}2$. By contrast, cutting off a 10-block plan early at closest approach hurts PointMaze-Medium because min-over-plan gives CEM no reason to reach the anchor early in a long rollout. Finally, the graph ablation in \cref{fig:ablate}b shows that neighbour edges are essential. Dropping neighbour edges collapses PointMaze-Medium from \R{abl-full}\% down to \R{abl-knn0}\%, while varying $k$ across 5, 10, and 20 changes success very little.

\begin{table}[t]
\centering
\small
\renewcommand{\arraystretch}{1.08}
\setlength{\tabcolsep}{7pt}
\caption{\textbf{Robustness stress tests for Graph Anchors.} This table supplements \cref{fig:robustbars} with the complete training-data subsampling curve, larger stitched budgets, mid-episode kidnapping, and Reacher results. Entries show means and individual runs for seeds 0 and 42, while Reacher lists $d{=}25, 50, 100,$ and $150$ on seed 0.}
\label{tab:stress}
\begin{tabularx}{0.84\textwidth}{@{}Xcc@{}}
\toprule
Stress test & Mean & Runs \\
\midrule
Kidnapped mid-episode, PointMaze-M $d{=}200$ & 65 & 66/64 \\
Kidnapped mid-episode, Two-Room $d{=}75$ & 74 & 76/72 \\
Stitched goals, PointMaze-M, budget 400 & 94 & 98/90 \\
Stitched goals, PointMaze-M, budget 600 & 99 & 100/98 \\
Stitched goals, PointMaze-M, budget 800 & 99 & 100/98 \\
Reacher, mean over $d{=}25..150$ & 62 & 60/60/64/64 \\
\midrule
Data fraction 1\%, $d{=}200$ / stitched & 63 / 86 & 74/52 \\
Data fraction 2\%, $d{=}200$ / stitched & 80 / 73 & 86/74 \\
Data fraction 5\%, $d{=}200$ / stitched & 78 / 86 & 82/74 \\
Data fraction 10\%, $d{=}200$ & 79 & 80/78 \\
Data fraction 25\%, $d{=}200$ & 83 & 76/90 \\
Data fraction 50\%, $d{=}200$ & 84 & 86/82 \\
Data fraction 100\%, $d{=}200$ / stitched & 86 / 94 & 92/80 \\
 
\bottomrule
\end{tabularx}
\end{table}

\paragraph{Stitching, limited data, and displacement.}
\Cref{tab:stress} fills in the full numbers behind \cref{fig:robustbars}, covering training-data subsampling, larger stitched budgets, mid-episode kidnapping, and Reacher. Because 99.9\% of neighbour edges connect distinct episodes as shown in \cref{fig:suppgraph}c, the graph assembles valid routes even under extreme data scarcity or shift. Keeping just 1\% of training episodes still solves \Q{frac-0.01-ours}\% of PointMaze-Medium $d{=}200$ goals, climbing steadily to 86\% with the full dataset. Pure random-play data reaches \R{rand-ours-rand-200}\% on PointMaze-Medium, and encoding an unseen dataset never used during world-model training still reaches \Pf{pml-sd-400}\% on PointMaze-Large at $d{=}400$. Mid-episode displacement needs no extra recovery mechanism either. Whenever the agent is abruptly moved, the next replanning step simply re-anchors from the new observation, shifting only from \Q{st-kidref-pmm-ours}\% to \Q{st-kid-pmm-ours}\% on PointMaze-Medium and from 93\% to 74\% on Two-Room.

\FloatBarrier
% ==============================================================================
\section{Additional Baselines and Planner Details}
\label{app:planners}\label{app:baselines}\label{app:contemp}

% ------------------------------------------------------------------------------
\subsection{World-Model Baselines and Graph-Guided Policies (GAS, TTGS)}
\label{app:gasttgs}

\paragraph{World-model and offline-policy baselines.}
We train PLDM~\cite{sobal2026learning}, DINO-WM~\cite{zhou2024dino}, GCBC, GCIVL, GCIQL~\cite{park2025ogbench}, and HILP~\cite{park2024foundation} inside the exact same \texttt{stable-worldmodel} codebase on identical datasets, test pairs, and budgets. All offline policies act in closed loop from frozen DINOv2 features plus proprioceptive goal state, trained for one epoch of up to 5{,}000 batches of size 128 per phase. PLDM trains for one epoch of 10{,}000 batches and plans with LeWM's CEM setup. DINO-WM trains for five epochs of 5{,}000 batches and plans directly over patch tokens. Because DINO-WM takes 37 to 68\,s per decision on an H200 GPU, more than $20\times$ slower than LeWM, we evaluate it on 20 seed-0 pairs at long distances.

Because one epoch is much shorter than typical 100-epoch policy defaults, these policy rows act as matched-budget references. Even so, extending training to four or five epochs on PointMaze-Medium in \cref{tab:budget}b leaves GCIVL unchanged, lifts GCBC only at $d{=}50$ to \Q{bud-gcbc4-50}\%, and leaves five-epoch PLDM weak at $d{=}200$ at 1\% alone and 7\% with anchors. When PLDM already has a reliable local model on Two-Room in \cref{tab:pldm}a, adding Graph Anchors lifts $d{=}75$ success from \R{pldm-tworoom-released-75}\% to \R{pldm-tworoom-ours-75}\% and stitched goals from 81\% to 99\%. On Reacher, whose dataset is uniform random play, offline policies have no goal-directed demonstrations to imitate and remain at 0 to 4\%, whereas dynamics models still plan effectively, with PLDM reaching \R{reacher-pldm-25}\% at $d{=}25$ and \R{reacher-pldm-150}\% at $d{=}150$.

Finally, \cref{tab:dwmcost} explains DINO-WM's strong PointMaze-Large result of \R{pml-dwm-400}\%. Its default cost includes ground-truth proprioceptive agent coordinates alongside the image. Removing position coordinates drops image-only DINO-WM down to \Pf{dwmc-pml-pix}\%, compared to \Pf{dwmc-pml-prop}\% using position alone and \R{pml-ours-400}\% for image-only Graph Anchors.

\begin{table}[t]
\centering
\small
\renewcommand{\arraystretch}{1.08}
\setlength{\tabcolsep}{7pt}
\caption{\textbf{DINO-WM on PointMaze-Large relies on goal coordinates.} Success (\%) at $d{=}400$ for one trained model evaluated with three goal costs on 20 pairs per run. Graph Anchors uses images only across 50 pairs per seed.}
\label{tab:dwmcost}
\begin{tabularx}{0.84\textwidth}{@{}lXc@{}}
\toprule
Planner & Goal cost & Success (seeds) \\
\midrule
DINO-WM & image + position (default) & \R{pml-dwm-400} / \Pf{dwmc-pml-full-42} / \Pf{dwmc-pml-full-1} \\
DINO-WM & position only & \Pf{dwmc-pml-prop} \\
DINO-WM & image only & \Pf{dwmc-pml-pix} \\
\midrule
\rowcolor{oursrow}\textbf{Graph Anchors (LeWM)} & image only & \textbf{\R{pml-ours-400}} (mean) \\
\bottomrule
\end{tabularx}
\end{table}

\begin{table}[t]
\centering
\caption{\textbf{PLDM anchors and longer baseline training.} Entries report success (\%). Panel~(a) adds Graph Anchors to one-epoch PLDM across seeds 0 and 42, averaging two trainings on PointMaze-Medium. Panel~(b) tests longer training on PointMaze-Medium, with four-epoch GCBC and GCIVL evaluated on seed 0 and on stitched seeds 0 and 42.}
\label{tab:pldm}\label{tab:budget}
\footnotesize
\setlength{\tabcolsep}{3.5pt}
\begin{minipage}[t]{0.52\textwidth}\centering
\textbf{(a)} Anchors on PLDM\\[3pt]
\begin{tabular}{@{}lccccccc@{}}
\toprule
 & \multicolumn{3}{c}{Two-Room} && \multicolumn{3}{c}{PointMaze-Medium} \\
\cmidrule(lr){2-4}\cmidrule(lr){6-8}
 & 25 & 50 & 75 && 100 & 200 & St. \\
\midrule
PLDM planner & 98 & 75 & 50 && 17 & 3 & 28 \\
+ min-over-plan & 100 & 85 & 62 && 12 & 4 & 27 \\
\rowcolor{pactrow}\textbf{+ graph anchors (ours)} & 95 & 98 & 97 && 28 & 4 & 28 \\
 
\bottomrule
\end{tabular}
\end{minipage}\hfill
\begin{minipage}[t]{0.45\textwidth}\centering
\textbf{(b)} Longer training, PointMaze-Medium\\[3pt]
\begin{tabular}{@{}lccccc@{}}
\toprule
 & 50 & 100 & 150 & 200 & St. \\
\midrule
GCBC, 1 epoch (3 trainings) & 93 & 83 & 54 & 51 & 51 \\
GCBC, 4 epochs & 98 & 88 & 68 & 58 & 59 \\
GCIVL, 1 epoch & 30 & 4 & 2 & 0 & 16 \\
GCIVL, 4 epochs & 30 & 2 & 0 & 0 & 12 \\
 
\bottomrule
\end{tabular}
\end{minipage}
\end{table}

\paragraph{Why graph-guided policies (GAS and TTGS) fail on visual mazes.}
A natural question is whether simply adding a graph on top of an offline policy already solves long-horizon visual planning. Prior methods like GAS~\cite{baek2025graph}, which builds a graph in HILP temporal-distance space $\phi$, and TTGS~\cite{opryshko2025test}, which builds a test-time graph from learned value distance $\log(V(1{-}\gamma){+}1)/\log\gamma$, take exactly this approach. Examining both their graph construction in \cref{tab:gasttgs_diag} and their closed-loop execution in \cref{tab:graphpol} reveals \emph{three coupled failure modes} that Graph Anchors directly resolves:
\begin{enumerate}[leftmargin=1.5em,itemsep=4pt,topsep=3pt]
\item \textbf{Metric coupling and the sharp $\tau$ threshold jump.} Both GAS and TTGS connect nodes by thresholding a \emph{learned} value or temporal-distance head at $d_V < \tau$, requiring per-task and per-agent $\tau$ tuning. Whenever that learned head saturates on visual mazes, it outputs nearly the same high distance for almost every pair of states, leaving no usable $\tau$ window at all. On simpler Two-Room, GCIVL predicts a healthy median distance of 2.9, so both TTGS and GAS form local edges and solve over 82\% of goals in \cref{tab:gasttgs_diag}. On PointMaze-Medium and PointMaze-Large, however, GCIVL distances cluster near 24 to 25. Any threshold below that plateau, such as $\tau{\le}16$, leaves TTGS with zero value edges out of more than 142k candidates, disconnecting episodes completely. Raising $\tau$ just past the plateau flips straight to \textbf{100\% of all candidates}, turning an empty graph directly into a dense graph full of false shortcuts that solves only 0\% on PointMaze-Medium and 5\% on PointMaze-Large. GAS hits a related collapse, shrinking PointMaze-Large from 2{,}837 nodes down to just \textbf{75 nodes} at $\tau{=}16$ and admitting every edge at $\tau{\ge}24$, yet solving at most 19\% on PointMaze-Medium and 2\% on PointMaze-Large.

\emph{Resolution.} Graph Anchors decouples the graph from any learned value head and sets $\tau$ automatically from the frozen encoder's empirical 1-block median, achieving 0.00\% wall crossings on PointMaze-Medium and 0.13\% long jumps on PointMaze-Large with zero tuning.
\item \textbf{Quadratic $O(N^2)$ neural distance matrices and undirected shortcuts.} TTGS evaluates $V(s_i,s_j)$ over all $N^2$ node pairs, forcing graphs to stay capped at 2k to 4k random frames. GAS similarly constructs a dense $N{\times}N$ pairwise matrix with bidirectional $\phi$-edges plus forced component-merging bridges, discarding temporal directionality and contact continuity. \emph{Resolution.} Graph Anchors indexes 100\% of training frames, from 199k up to 453k nodes, via a sparse 10-NN graph combined with directed temporal edges and switch cost $\lambda$.
\item \textbf{Double bottleneck: policy execution fails even on a verified graph.} Even when we discard GAS and TTGS's broken value graphs and hand GCBC, GCIVL, and HILP waypoints from our own verified shortcut-free graph, closed-loop execution still fails. As reported in \cref{tab:graphpol}, holding each waypoint until reached caps PointMaze-Medium $d{=}200$ at 21\% and PointMaze-Large $d{=}400$ at 7\%, dropping GCBC by 30 points below its no-graph score. Refreshing waypoints every five policy steps on PointMaze-Medium across three seeds hurts GCBC too, pulling $d{=}200$ down from 55\% alone to 43\% at $K{=}6$ and 25\% at $K{=}2$, since the policy reaches fewer than 1\% of $K{=}6$ waypoints before refresh. Offline RL policies also require goal-directed trajectories and fail on random play, scoring 0 to 4\% on Reacher.

\emph{Resolution.} Pairing our sparse graph with frozen world-model CEM executes nearby visual targets through imagined physics without policy training, solving PointMaze-Medium at \R{pmm-ours-200}\% and 94\% from random play, PointMaze-Large at \R{pml-ours-400}\%, and Reacher at 68\%. Although longer training beyond our 1-epoch budget would soften value saturation in GCIVL and HILP, training-free deployment, random-play robustness, and full-dataset sparse scaling remain structural advantages of Graph Anchors.
\end{enumerate}

\begin{table}[t]
\centering
\small
\renewcommand{\arraystretch}{1.08}
\setlength{\tabcolsep}{5pt}
\caption{\textbf{Value-graph saturation and collapse in TTGS and GAS}, averaged over all evaluation logs per environment and $\tau$. On visual mazes, learned value distances saturate near 24 to 25 so TTGS jumps abruptly from 0\% to 100\% edges with no usable $\tau$ window, while GAS collapses to 75 nodes on PointMaze-Large at $\tau{=}16$. Graph Anchors self-calibrates on 100\% of frames.}
\label{tab:gasttgs_diag}
\begin{tabular}{@{}llccccc@{}}
\toprule
Env. & Method ($\tau$) & Nodes & Med.\ $d_V$ & Value edges (\%) & Far (\%) & St.\ (\%) \\
\midrule
Two-Room & TTGS ($\tau{=}16$) & 2{,}981 & 2.9 & 99.8\% & 82 & 87 \\
($d{=}75$) & GAS ($\tau{=}24$) & 2{,}981 & 2.2 & 100.0\% & 86 & 86 \\
 & \cellcolor{oursrow}Ours (1-blk $\tau$) & \cellcolor{oursrow}181k & \cellcolor{oursrow}-- & \cellcolor{oursrow}0.0\% jumps & \cellcolor{oursrow}\textbf{93} (100$^c$) & \cellcolor{oursrow}\textbf{99} \\
\midrule
PM-M & TTGS ($\tau{=}16/24/32$) & 2{,}946 & 24.6 & \textbf{0.0 / 0.2 / 100\%} & 0 & 17 \\
($d{=}200$) & GAS ($\tau{=}24/32$) & 2{,}946 & 4.4 & 100.0\% & 19 & 36 \\
 & \cellcolor{oursrow}Ours (1-blk $\tau$) & \cellcolor{oursrow}199k & \cellcolor{oursrow}-- & \cellcolor{oursrow}0.0\% jumps & \cellcolor{oursrow}\textbf{85} (94$^c$) & \cellcolor{oursrow}\textbf{95} \\
\midrule
PM-L & TTGS ($\tau{=}16/24$) & 2{,}837 & 23.9 & \textbf{0.0 / 100\%} & 5 & 3 \\
($d{=}400$) & GAS ($\tau{=}16$) & \textbf{75} & 14.3 & 80.8\% & 2 & 3 \\
 & GAS ($\tau{=}24/32$) & 2{,}837 & 5.5 & 100.0\% & 2 & 3 \\
 & \cellcolor{oursrow}Ours (1-blk $\tau$) & \cellcolor{oursrow}200k & \cellcolor{oursrow}-- & \cellcolor{oursrow}0.13\% jumps & \cellcolor{oursrow}\textbf{87} [76] & \cellcolor{oursrow}\textbf{66} [45] \\
\bottomrule
\end{tabular}
\end{table}

\begin{table}[t]
\centering
\caption{\textbf{Offline policies paired with graphs on the mazes.} Success (\%) using held waypoints from our verified latent graph and GAS-style subgoals taking the best $\tau \in \{2,4,8\}$ per cell, averaged over seeds 0, 42, and 1, with stitched goals over seeds 0 and 42. Bold marks the best score per column. Even on our verified latent graph, offline policies remain far below Graph Anchors.}
\label{tab:graphpol}\label{tab:waypoint2}
\footnotesize
\renewcommand{\arraystretch}{1.06}
\setlength{\tabcolsep}{3.5pt}
\begin{tabularx}{0.96\textwidth}{@{}p{3.9cm}YYYYYYYYYY@{}}
\toprule
 & \multicolumn{5}{c}{PointMaze-Medium} & \multicolumn{5}{c}{PointMaze-Large} \\
\cmidrule(lr){2-6}\cmidrule(l){7-11}
Method & 50 & 100 & 150 & 200 & st. & 50 & 100 & 200 & 400 & st. \\
\midrule
GCBC + graph waypoints & 84 & 70 & 39 & 21 & 41 & 17 & 15 & 10 & 6 & 4 \\
GCIVL + graph waypoints & 34 & 8 & 3 & 0 & 18 & 8 & 7 & 9 & 2 & 1 \\
HILP + graph waypoints & 31 & 3 & 1 & 0 & 18 & 8 & 9 & 8 & 7 & 4 \\
GAS-style (HILP) & 31 & 3 & 1 & 0 & 18 & 9 & 10 & 10 & 10 & 8 \\
\midrule
GCBC, no graph & \textbf{93} & 83 & 54 & 51 & 51 & 33 & 25 & 18 & 18 & 17 \\
HILP, no graph & 32 & 3 & 1 & 1 & 18 & 12 & 13 & 16 & 9 & 10 \\
\midrule
\rowcolor{oursrow}\textbf{Graph Anchors (ours)} & 91 & \textbf{89} & \textbf{88} & \textbf{85} & \textbf{95} & \textbf{62} & \textbf{63} & \textbf{73} & \textbf{87} & \textbf{66} \\
\bottomrule
\end{tabularx}
\end{table}

\FloatBarrier
% ------------------------------------------------------------------------------
\subsection{Implementation Fidelity, Push-T Scaling, and Full Metrics for AP, NGC, Metro-WM}
\label{app:frozenplanners}

This section provides the full re-implementation details, reproduction check, Push-T distance scaling curve, and complete quantitative tables supporting the analysis in \cref{sec:frozenplanners}.

\paragraph{Shared evaluation setup and re-implementation details.}
We re-implemented AP~\cite{liu2026aim}, NGC~\cite{you2026latent}, and Metro-WM~\cite{lee2026metro} inside our shared evaluation framework on identical 5-step training nodes, test pairs, seeds, $2d$ budgets, and test-episode exclusions:
\begin{itemize}[leftmargin=1.4em,itemsep=2pt,topsep=2pt]
\item \emph{AP} retrieves the single-episode pair $(z_s, z_{s+h})$ whose standardized key $(z_s, z_{s+h}, z_{s+h}{-}z_s)$ best matches $(z_t, z_g, z_g{-}z_t)$. It then targets the next block $z_{s+1}$ using a 1-block terminal-cost CEM plan and decrements $h$ from the given true step distance $d$.
\item \emph{NGC} replaces Euclidean goal distance with shortest-path length over an undirected mutual 15-NN graph of 5{,}000 random training snapshots, attached via 4 nearest nodes on each side, while keeping LeWM's 5-block terminal-cost CEM.
\item \emph{Metro-WM-style} assigns temporal edges their 5-frame gap and links each node to its 4 nearest cross-episode nodes within a radius calibrated at the 0.25-quantile of intra-episode latent gaps. It selects the farthest reachable node within budget $H$ frames and executes an $h$-block terminal-cost plan in full, sweeping $(h,H)$ per task as in~\cite{lee2026metro}. Our implementation uses 5-step vertex strides and falls back to $z_g$ if disconnected.
\end{itemize}

\begin{figure}[t]
\centering
\begin{minipage}[c]{0.47\textwidth}
\centering
\includegraphics[width=\linewidth]{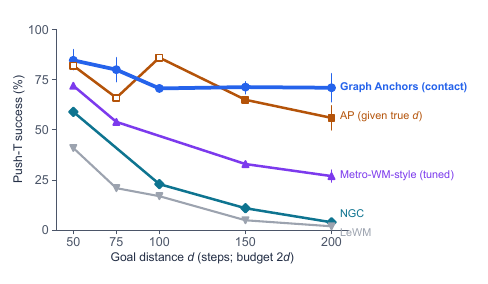}
\caption{\textbf{Push-T success across goal distance} for all four training-free planners on identical frozen LeWM checkpoints, graph nodes, and 50-pair seeds with budget $2d$. Graph Anchors uses its single unchanged contact setting ($K{=}2, \lambda{=}3$) at every distance, AP receives the true step count $d$, and Metro-WM uses its best Push-T setting ($3{:}20$). Filled markers show multi-seed means with range bars; hollow markers show one seed.}
\label{fig:pushtdist}
\end{minipage}\hfill
\begin{minipage}[c]{0.49\textwidth}
\centering
\captionof{table}{\textbf{Re-implementation reproduction check vs.\ published papers.} Across every overlapping setting reported in the original papers, all three re-implementations agree within 9 percentage points despite our tighter $2d$ step budget.}
\label{tab:fidelity}
\footnotesize
\renewcommand{\arraystretch}{1.10}
\setlength{\tabcolsep}{4pt}
\begin{tabularx}{\linewidth}{@{}Xccc@{}}
\toprule
Setting & Reported & Ours & Runs \\
\midrule
\input{tables/contemp_fid_rows.tex} 
\bottomrule
\end{tabularx}
\end{minipage}
\end{figure}

\Cref{tab:fidelity} verifies that all three re-implementations closely match their published results. \Cref{fig:pushtdist} extends the Push-T comparison all the way to $d{=}200$, where Graph Anchors' unchanged contact setting stays flat at \Ed{ours-200}\% while AP drops to \Ed{ap-200}\%, Metro-WM to \Ed{metro-200}\%, and NGC to \Ed{ngc-200}\%. \Cref{tab:comp_mech} lists the complete route-continuity statistics, RTX 3090 Ti memory and cost-latency benchmarks, maze shortcut rates, PointMaze-Large 4-way scores, and AP stress tests summarized in \cref{sec:frozenplanners}.

\begin{table}[t]
\centering
\caption{\textbf{Route quality, compute scaling, and stress tests of AP, NGC, and Metro-WM.} Panel~(a) measures Push-T $d{=}150$ route continuity across 200 held-out routes on 453k nodes, where the 95th percentile of a real 1-block move is 34.5\,px. Panel~(b) benchmarks GPU memory and per-decision CEM cost latency on one RTX 3090 Ti over 50 environments, 300 samples, and 30 iterations, alongside maze shortcut jumps exceeding the 95th percentile of a real 5-step move, equal to 0.252 on full PointMaze-Medium, 0.250 at 1\%, and 1.020 on PointMaze-Large. Panel~(c) reports the PointMaze-Large 4-way comparison on \texttt{lewm\_pml} seed 0 with original-checkpoint scores in brackets, together with AP stress tests where $^c$ marks the contact setting. On Reacher, NGC at 75\% and LeWM with min-over-plan at 77\% exceed both AP and Graph Anchors.}
\label{tab:comp_mech}\label{tab:metrosweep}\label{tab:apoffset}
\scriptsize
\renewcommand{\arraystretch}{1.04}
\textbf{(a)} Push-T $d{=}150$ route continuity (200 held-out routes, 453k training nodes)\\[3pt]
\setlength{\tabcolsep}{4pt}
\resizebox{\textwidth}{!}{%
\begin{tabular}{@{}lcccccc@{}}
\toprule
Method & Recorded temporal (\%) & Switches / 6 moves (med / mean) & Cross-jump ${>}15$\,px (\%) & Block jump med / p90 (px) & Angle jump p90 & Push-T $d{=}150$ (\%) \\
\midrule
NGC ($N{=}5{,}000$) & 0.0 & all cross-ep. & 65.9 & 23.4 / 60.3 & 87.5$^\circ$ & 11 \\
Metro-WM ($H{=}20,25$) & 45.9 & 6.0 / 7.03 & 12.1 & 2.8 / 17.5 & 10.9$^\circ$ & 33 \\
Graph Anchors ($\lambda{=}1$) & 46.6 & 5.0 / 6.12 & 21.9 & 6.7 / 22.3 & 15.1$^\circ$ & 30 \\
\rowcolor{oursrow}\textbf{Graph Anchors ($\lambda{=}3$, contact)} & \textbf{82.0} & \textbf{1.0 / 1.68} & 18.5 & 5.0 / 20.1 & 13.6$^\circ$ & \textcolor{bestred}{\textbf{71}} \\
\bottomrule
\end{tabular}}

\vspace{6pt}
\begin{minipage}[t]{0.50\textwidth}\centering
\textbf{(b)} Memory, in-CEM cost latency \& maze shortcut jumps\\[3pt]
\setlength{\tabcolsep}{2.2pt}
\resizebox{\linewidth}{!}{%
\begin{tabular}{@{}lccccc@{}}
\toprule
Method & Nodes & Mem. & Cost / dec. & PM-M full/1\% & PM-L ${>}$p95 (max) \\
\midrule
NGC ($5$k) & 5k (2.5\%) & 95.4\,MB & 1{,}729.9\,ms & 0.41 / 1.37\% & \textbf{9.51\%} (22.8) \\
Metro ($H{=}20$) & 199k (100\%) & CSR & plain L2 & 0.00 / 1.83\% & 0.07\% (5.05) \\
Metro ($H{=}25$) & 199k (100\%) & CSR & plain L2 & 0.00 / 1.88\% & -- \\
\rowcolor{oursrow}\textbf{Ours (full)} & \textbf{199k (100\%)} & \textbf{72.9\,MB} & \textbf{64.6\,ms} & \textbf{0.00 / 0.00\%} & \textbf{0.13\%} (p95 0.11) \\
\bottomrule
\end{tabular}}
\end{minipage}\hfill
\begin{minipage}[t]{0.48\textwidth}\centering
\textbf{(c)} PointMaze-Large 4-way \& AP stress tests (\%)\\[3pt]
\setlength{\tabcolsep}{2.4pt}
\resizebox{\linewidth}{!}{%
\begin{tabular}{@{}lccc@{}}
\toprule
Setting & Competitor(s) & Ours & $\Delta$ \\
\midrule
PM-L $d{=}400$ (4-way, \texttt{lewm\_pml}) & NGC 24 / AP 18 / Met.\ 6 & \textbf{76} [87] & +52/+70 \\
PM-L stitched (4-way, \texttt{lewm\_pml}) & NGC 20 / AP 9 / Met.\ 6 & \textbf{45} [66] & +25/+39 \\
Push-T $d{=}150$, AP wrong $d$ ($0.5$/$2$/$3\times$) & AP 16 / 2 / 0 & \textbf{71}$^c$ & +55/+71 \\
Kidnapped PM-M, $d{=}200$ & AP 22 & \textbf{65} & +43 \\
Random-data PM-M, $d{=}200$ & AP 56 & \textbf{94} & +38 \\
Reacher, $d{=}150$ & AP 34 & \textbf{68} & +34 \\
PM-M $d{=}200$, AP wrong $d$ ($0.5$/$3\times$) & AP 29 / 11 & \textbf{85} (94$^c$) & +56/+83 \\
Two-Room $d{=}75$, AP wrong $d$ ($0.25$/$3\times$) & AP 36 / 2 & \textbf{93} (100$^c$) & +57/+98 \\
\bottomrule
\end{tabular}}
\end{minipage}
\end{table}

\FloatBarrier
% ==============================================================================
\section{Full Per-Evaluation Results, Qualitative Rollouts, and Reproducibility}
\label{app:limits}\label{app:qual}\label{app:repro}

\begin{figure}[t]
\centering
\includegraphics[width=0.92\textwidth]{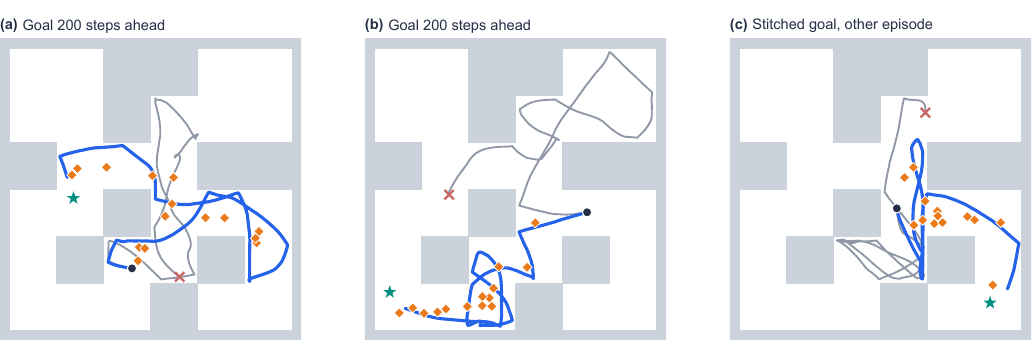}
\caption{\textbf{Rollouts on PointMaze-Medium} for seed 0 on identical start and goal pairs, comparing LeWM in grey with its final cross against Graph Anchors in blue with active anchor diamonds. Episodes follow a fixed selection rule choosing the longest successful paths solved by anchors and missed by LeWM whose starts lie at least 1.5 units apart.}
\label{fig:qual}
\end{figure}

\begin{figure}[t]
\centering
\includegraphics[width=0.88\textwidth]{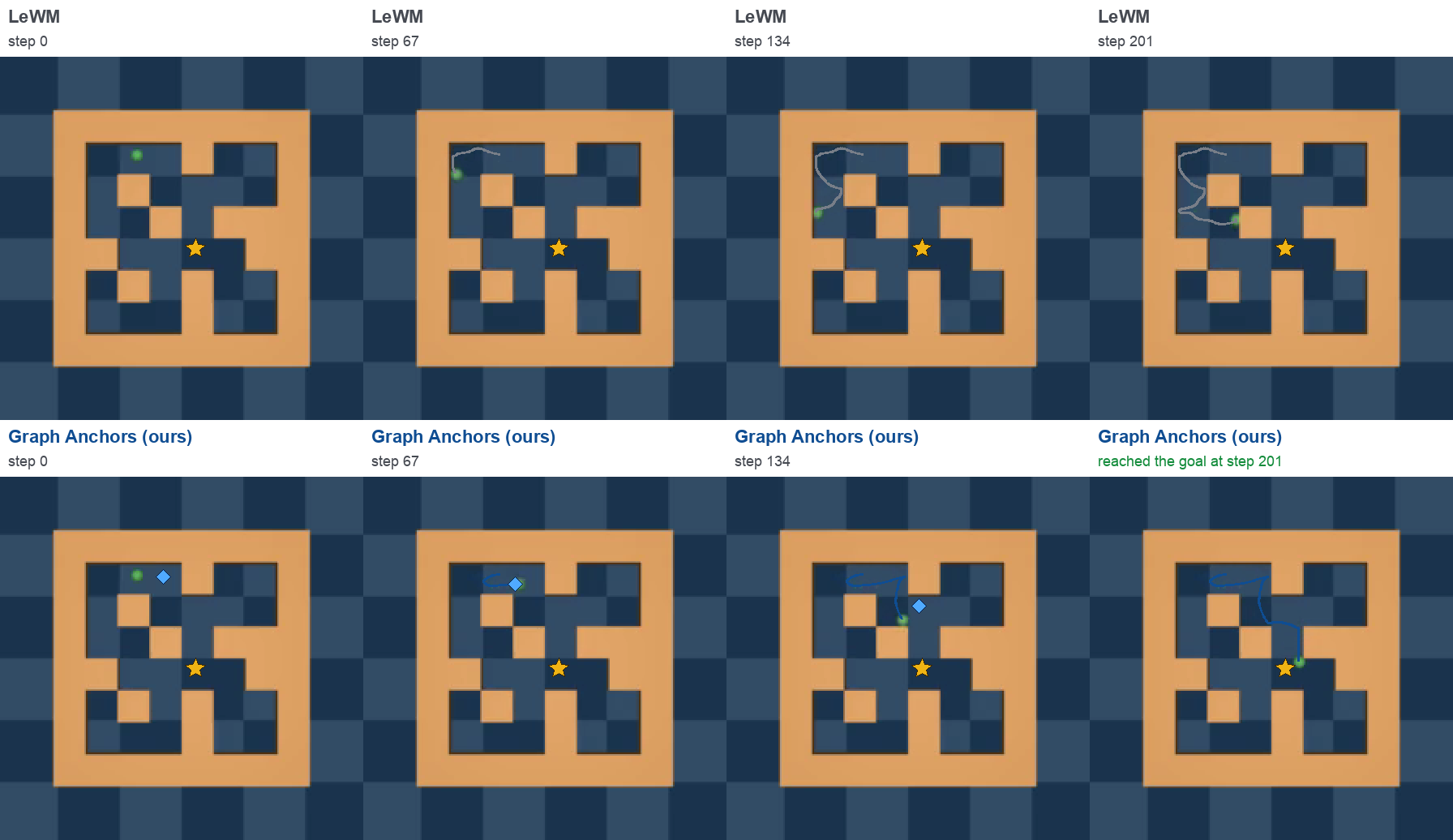}
\caption{\textbf{Planner camera view} on PointMaze-Medium at $d{=}200$ and seed 0 across four moments on an identical start and goal pair. The star marks the goal, the diamond marks the active anchor, and the curve traces the path taken so far.}
\label{fig:plannerview}
\end{figure}

\paragraph{Rollouts and failure analysis.}
\Cref{fig:qual} and \cref{fig:plannerview} show LeWM wandering near the start while Graph Anchors steer smoothly around maze walls. Inspecting every traced PointMaze-Medium failure on seed 0, covering \Q{fail-d200-failed} of \Q{fail-d200-n} episodes at $d{=}200$ and \Q{fail-x-failed} of \Q{fail-x-n} stitched episodes, confirms that anchors stayed active with zero stalls. Specifically, \Q{fail-d200-nearanchor} of \Q{fail-d200-failed} failed $d{=}200$ episodes came within 0.5 maze cells of an anchor and ended \Q{fail-d200-dmin} to \Q{fail-d200-dmax} units from the goal. Remaining failures therefore come from slow progress along valid routes rather than wrong paths or traps.

\paragraph{Reproducibility and video archive.}
Our project page hosts 29 rollout and diagnostic videos covering the range diagnosis, 3D latent PCA, side-by-side LeWM and anchor rollouts across environments, 4-way Push-T $d{=}200$ comparisons, and real-time active-anchor camera overlays (\cref{fig:plannerview,fig:plannerviewpusht}). Repeated runs on the same machine match identically, cross-machine runs stay within a few points so every main-table comparison pairs same-machine runs, and \cref{tab:seedsx} records every individual 50-pair evaluation behind \cref{tab:main,tab:compare,tab:pldm}.

\begin{table}[p]
\centering
\caption{\textbf{Complete per-evaluation results} behind \cref{tab:main,tab:compare,tab:pldm}, excluding DINO-WM which ran one 20-pair evaluation per cell of \cref{tab:main}. Each entry lists success (\%) for individual 50-pair runs separated by ``/'', while ``--'' means not run. LeWM, LeWM with min-over-plan, and Graph Anchors list seeds 0, 42, and 1, except Push-T $d{\ge}50$ and PointMaze-Large use seeds 0 and 42 and PointMaze-Medium stitched LeWM baselines use seeds 0 and 1. GCBC on PointMaze-Medium lists trainings A and B on seed 0 followed by training C on seeds 0, 42, and 1, with stitched runs adding seed 42 for training A. Other offline policies use seeds 0, 42, and 1 or 0 and 42. PLDM on PointMaze-Medium lists the first training on seeds 0 and 42 followed by the second training on seeds 0 and 42, with $d{=}50$, $d{=}150$, and stitched min-over-plan using the second training. Cube and Reacher use seed 0.}
\label{tab:seedsx}
\scriptsize
\renewcommand{\arraystretch}{1.12}
\setlength{\tabcolsep}{2.5pt}

\begin{minipage}[t]{0.485\textwidth}\centering
{\scriptsize\textbf{(a) PointMaze-Medium} (stitched: budget 400)}\\[3pt]
\resizebox{\linewidth}{!}{%
\begin{tabular}{@{}lccccc@{}}
\toprule
Method & $d{=}50$ & $d{=}100$ & $d{=}150$ & $d{=}200$ & stitched \\
\midrule
LeWM & 38/50/62 & 24/22/30 & 10/12/16 & 16/12/12 & 46/32 \\
LeWM + min & 66/68/76 & 30/40/30 & 12/20/18 & 14/6/16 & 44/38 \\
\rowcolor{oursrow}\textbf{Graph Anchors} & \textbf{90/90/94} & \textbf{84/92/90} & \textbf{88/94/82} & \textbf{92/80/82} & \textbf{98/90/96} \\
GCBC & 92/94 & 84/80/82/86/82 & 50/58 & 52/38/54/56/54 & 48/46/56/52/46/60/46 \\
GCIVL & 30 & 4 & 2 & 0 & 16/16 \\
GCIQL & 32/40/30 & 6/4/2 & 0/0/0 & 2/0/0 & 14/10 \\
HILP & 28/36/32 & 4/4/0 & 0/2/0 & 2/0/0 & 18/18 \\
PLDM & 26/34 & 20/14/16/18 & 4/4 & 6/0/6/0 & 30/24/32/26 \\
PLDM + min & 50/52 & 12/8/16/14 & 2/6 & 8/4/2/4 & 26/28 \\
\rowcolor{oursrow}PLDM + anchors & 52/66 & 30/24/30/28 & 8/10 & 4/2/4/4 & 30/30/30/24 \\
\bottomrule
\end{tabular}}
\end{minipage}\hfill
\begin{minipage}[t]{0.485\textwidth}\centering
{\scriptsize\textbf{(b) Two-Room} (stitched: budget 200)}\\[3pt]
\resizebox{\linewidth}{!}{%
\begin{tabular}{@{}lcccc@{}}
\toprule
Method & $d{=}25$ & $d{=}50$ & $d{=}75$ & stitched \\
\midrule
LeWM & 88/84/92 & 52/50/60 & 28/32/34 & 56/68/60 \\
LeWM + min & 98/98/100 & 76/86/82 & 50/68/68 & 84/78/84 \\
\rowcolor{oursrow}\textbf{Graph Anchors} & \textbf{92/96/100} & \textbf{92/94/94} & \textbf{94/92/94} & \textbf{100/98/98} \\
GCBC & 100 & 98 & 100 & 98/100 \\
GCIVL & 100/100/96 & 90/90/90 & 84/80/80 & 82/90 \\
GCIQL & 62/60/62 & 62/40/52 & 38/32/40 & 46/46 \\
HILP & 98/98/98 & 96/96/94 & 84/86/90 & 88/90 \\
PLDM & 100/96 & 76/74 & 46/54 & 80/82 \\
PLDM + min & 100/100 & 88/82 & 62/62 & -- \\
\rowcolor{oursrow}PLDM + anchors & 94/96 & 96/100 & 98/96 & 100/98 \\
\bottomrule
\end{tabular}}
\end{minipage}

\vspace{10pt}
\begin{minipage}[t]{0.485\textwidth}\centering
{\scriptsize\textbf{(c) PointMaze-Large} (stitched: budget 400)}\\[3pt]
\resizebox{\linewidth}{!}{%
\begin{tabular}{@{}lccccc@{}}
\toprule
Method & $d{=}50$ & $d{=}100$ & $d{=}200$ & $d{=}400$ & stitched \\
\midrule
LeWM & 16/12 & 20/10 & 10/16 & 8/8 & 4/6 \\
LeWM + min & 50/38 & 30/26 & 20/26 & 30/22 & 18/16 \\
\rowcolor{oursrow}\textbf{Graph Anchors} & \textbf{66/58} & \textbf{58/68} & \textbf{76/70} & \textbf{88/86} & \textbf{70/62} \\
GCBC & 24/42 & 22/28 & 16/20 & 20/16 & 16/18 \\
GCIVL & 8/10 & 10/6 & 10/8 & 2/0 & 2/4 \\
GCIQL & 8/10 & 10/6 & 10/8 & 2/4 & 2/4 \\
HILP & 12/12 & 16/10 & 14/18 & 10/8 & 6/14 \\
\bottomrule
\end{tabular}}
\end{minipage}\hfill
\begin{minipage}[t]{0.485\textwidth}\centering
{\scriptsize\textbf{(d) Push-T} (navigation setting, contact in \cref{tab:contactcfg})}\\[3pt]
\resizebox{\linewidth}{!}{%
\begin{tabular}{@{}lccccc@{}}
\toprule
Method & $d{=}25$ & $d{=}50$ & $d{=}75$ & $d{=}100$ & $d{=}150$ \\
\midrule
LeWM & 88/90/88 & 32/50 & 24/18 & 18/16 & 2/8 \\
LeWM + min & 94/90/88 & 48/62 & 38/28 & 22/12 & 8/10 \\
\rowcolor{oursrow}\textbf{Graph Anchors} & \textbf{88/82/76} & \textbf{64/66} & \textbf{50/50} & \textbf{38/42} & \textbf{28/32} \\
GCBC & 52/28 & 20/12 & 12/4 & 10/12 & 6/4 \\
GCIVL & 4/0 & 2/2 & 0/0 & 0/0 & 0/0 \\
GCIQL & 0/0 & 0/0 & 0/0 & 0/0 & 0/0 \\
HILP & 6/2 & 0/0 & 0/0 & 0/0 & 0/0 \\
\bottomrule
\end{tabular}}
\end{minipage}

\vspace{10pt}
\begin{minipage}[t]{0.485\textwidth}\centering
{\scriptsize\textbf{(e) OGBench-Cube} (seed 0)}\\[3pt]
\begin{tabular}{@{}lcccc@{}}
\toprule
Method & $d{=}25$ & $d{=}50$ & $d{=}100$ & $d{=}150$ \\
\midrule
LeWM & 70 & 64 & 50 & 20 \\
LeWM + min & 68 & 60 & 56 & 20 \\
\rowcolor{oursrow}\textbf{Graph Anchors} & \textbf{70} & \textbf{56} & \textbf{60} & \textbf{20} \\
\bottomrule
\end{tabular}
\end{minipage}\hfill
\begin{minipage}[t]{0.485\textwidth}\centering
{\scriptsize\textbf{(f) Reacher} (seed 0)}\\[3pt]
\begin{tabular}{@{}lcccc@{}}
\toprule
Method & $d{=}25$ & $d{=}50$ & $d{=}100$ & $d{=}150$ \\
\midrule
LeWM & 60 & 74 & 62 & 66 \\
LeWM + min & 82 & 80 & 78 & 74 \\
\rowcolor{oursrow}\textbf{Graph Anchors} & \textbf{60} & \textbf{60} & \textbf{64} & \textbf{64} \\
\bottomrule
\end{tabular}
\end{minipage}
\end{table}

\end{document}